\documentclass[11pt]{report}
\usepackage{upennstyle} 

\usepackage{textcomp}
\newcommand{\tildehack}{\raisebox{0.5ex}{\texttildelow}}

\usepackage{enumitem}
\setlist[itemize]{noitemsep, topsep=0pt, left=1.5em}
\setlist[enumerate]{noitemsep, topsep=0pt, left=1.5em}

\usepackage{tabularray}
\usepackage{booktabs}
\usepackage{threeparttable}
\usepackage{makecell}

\usepackage{array}  
\usepackage{multirow}  
\usepackage{arydshln}  

\usepackage{lmodern} 

\usepackage{tablefootnote} 

\usepackage{xcolor} 
\usepackage{tikz} 

\usepackage[square,numbers]{natbib}
\title{Elastomeric strain limitation for design of soft pneumatic actuators}
\author{Gregory M. Campbell}
\gradgroup{Mechanical Engineering and Applied Mechanics} 
\date{2025} 
\supervisor{Mark Yim}
\supervisortitle{Asa Whitney Professor of Mechanical Engineering and Applied Mechanics}
\gradchair{Jordan Raney, Associate Professor of Mechanical Engineering and Applied Mechanics}
\committee{Mark Yim, Asa Whitney Professor of Mechanical Engineering and Applied Mechanics}
\committee{James H. Pikul, Leon and Elizabeth Janssen Associate
Professor of Mechanical Engineering, University of Wisconsin-Madison}
\committee{Michael Posa, Associate Professor of Mechanical Engineering and Applied Mechanics}

\authorlegal{Gregory Matthew Campbell} 

\dedication{To my family, biological and chosen, thank you for shaping me into the person I am today. \\
Specifically to my daughter, Josephine, thank you for guiding me home when the world is dark.} 

\acknowledgement{

I am incredibly grateful to everyone that made this work possible. To any and everyone who ever advocated for me, shared their knowledge with me, helped me with my work, or just shared their time with me: thank you.

I would like to thank my advisor, Mark Yim for his priorities and perspective in facilitating my growth throughout this research. Also, for sharing his vast research expertise and providing me with the flexibility and freedom to explore. The atmosphere and access that Modlab provides make it a catalyst for learning and invention - a testament to Mark's leadership.

Thank you to James Pikul for his vision and enthusiasm for the topics in this dissertation, as well as many others within robotics. I appreciate him welcoming me into his research group and offering optimism and perspective when I was feeling negatively about my work.

Thank you to Michael Posa for taking on the role of committee chair and offering a dissenting voice for many of the paths taken. His role as an advocate for impact in this work helped to improve it dramatically.

I am grateful to all the current and past members of Modlab and Pikul Lab that I had the pleasure of interacting with and growing with during my time at Penn. I am especially thankful to Devin Carroll and Alexander Spinos for the thought and effort they put into maintaining Modlab. Thanks to Jang-ho Bae for his guidance and camaraderie. Thanks to Walker Gosrich for offering valuable frame of reference for the research we conduct, being a great co-TA, and for introducing me to proper coffee. Thanks to Jessica Yin for her constant support, collaboration, guidance, and intellectual drive. Her input made much of what I accomplished here possible. Finally, thanks to Spencer Folk for being a constant advocate, sounding board, and friend. I count myself lucky to have had him with me throughout this degree.

Thank you to all the students that contributed to my work or included me on their own endeavors. In particular, thanks to: Yi, Jason, Hannah, Charles, Brennan, Tina, Kayla, Ashna, Sharon, Tate, Eric, and Majd. I have learned more from them than they know.

Thanks to my funding sources from NSF and Toyota TEMA for allowing me to make this my full-time job. And thanks to all the colleagues and advocates that I met through these programs, especially Pamela Cacchione, Alp Aydinoglu, Kristin Field, and Dasha Peppard.

I am especially thankful to all my collaborators, without whom I never would have gotten any of this accomplished. Special thanks to Kevin Turner and Daelan Roosa for sharing their resources and expertise. Thanks to Yuyang Song and Umesh Gandhi for their contributions to this research. Thanks to Owen Field, Ryan Hayward, and David Levine for teaching me about polymers and clutches. Thanks to Gentian Muhaxeri and Chris Santangelo for implementing complex maths and having the patience to teach me along the way. Finally, thanks to L\'eo Guilhoto for sharing his brilliant skill-set, his work ethic, and above all else his kindness. I have learned so much working along-side him.

Thanks to everyone else in MEAM, GRASP, and Penn generally who has positively impacted my time. Thanks to Torrie Edwards, Wei-Hsi Chen, and Tim Greco for their interest in my research and well-being. Thanks to Jake Welde, Parker Lamascus, and Will Yang for their friendship and support. Thanks to all the MEAM and GRASP staff, especially Peter Litt and Charity Payne, for helping me navigate the world of a PhD student. Thanks to the staff at CETLI, especially Ian Petrie, for making me a better teacher and learner.

Finally, I would like to thank my family for their continued presence and support in every aspect of my life. Thanks to my mom, dad, and sister, Lilly, for embodying excellence and helping shape me towards that image. Thanks to the Fillman family for welcoming me into their ranks. Thanks to my wife, Julia, for helping me pursue my passion and for her patience and understanding during this degree program.

} 

\abstract{
Modern robots embody power and precision control. Yet, as robots undertake tasks that apply forces on humans, this power brings risk of injury. Soft robotic actuators use deformation to produce smooth, continuous motions and conform to delicate objects while imparting forces capable of safely pushing humans. This thesis presents strategies for the design, modeling, and strain-based control of human-safe elastomeric soft pneumatic actuators (SPA) for force generation, focusing on embodied mechanical response to simple pressure inputs.

We investigate electroadhesive (EA) strain limiters for variable shape generation, rapid force application, and targeted inflation trajectories. We attach EA clutches to a concentrically strain-limited elastomeric membrane to alter the inflation trajectory and rapidly reorient the inflated shape. We expand the capabilities of EA for soft robots by encasing them in elastomeric sheaths and varying their activation in real time, demonstrating applications in variable trajectory inflation under identical pressure sweeps.

We then address the problem of trajectory control in the presence of external forces by modeling the pressure-trajectory relationship for a concentrically strain-limited class of silicone actuators. We validate theoretical models based on material properties and energy minimization using active learning and automated testing. We apply our ensemble of neural networks for inverse membrane design, specifying quasi-static mass lift trajectories from a simple pressure sweep. Finally, we demonstrate the power of multiple pressure-linked actuators in a proof-of-concept mannequin leg lift. 

} 

\begin{document}
\maketitle 
\setcounter{page}{2}

\makecopyright 

\makededication 

\makeacknowledgement 

\makeabstract
\tableofcontents

\clearpage \phantomsection \addcontentsline{toc}{chapter}{LIST OF TABLES} \begin{singlespacing} \listoftables \end{singlespacing}

\clearpage \phantomsection \addcontentsline{toc}{chapter}{LIST OF ILLUSTRATIONS} \begin{singlespacing} \listoffigures \end{singlespacing}

\clearpage \phantomsection \addcontentsline{toc}{chapter}{LIST OF ACRONYMS} \begin{singlespacing}

\begin{center}
    LIST OF ACRONYMS
\end{center}

SPA - Soft pneumatic actuators\\
EA - Electroadhesive\\
DoF - Degree of freedom\\
PAM - Pneumatic artificial muscle\\
FREE - Fiber reinforced elastomeric enclosures\\
DEA - Dielectric elastomer actuator\\
BOPET - Biaxially-oriented polyethylene terephthalate\\
FE - Finite element\\
RSME  - Root-mean squared error\\
PWM - Pulse width modulation\\
MTS - Material testing system\\
AL - Active learning\\
BO - Bayesian optimization\\
DoE - Design of experiments\\
GPR - Gaussian process regression\\
EI - Expected improvement
\end{singlespacing}%

\begin{mainf} 

\chapter{INTRODUCTION}

Robotics has historically promised to improve human existence by automating tasks that are dull, dangerous, dirty, or difficult. Modern implementation of this automation has increased industrial productivity and safety, leading to a general increase in human quality of life. The next wave of automation is coming in robots that reach individuals on the personal level to enhance human ability or relieve daily hardship. To enable this transition, we aim to develop a system that safely manipulates human beings for applications in personal care and healthcare.  

The goal in this thesis is to explore open research topics of soft robotics, embodied intelligence, and pneumatic actuation to address the gap between industrial robots and robotic actuators in the medical setting. We will provide human scale forces with an actuator that also conforms to the human body, minimizing local pressure and improving comfort. Our final goal is a soft robotic platform that is \textit{helpful}, \textit{accessible}, and \textit{human safe}.

\section{Problem Statement}
In this thesis, we move towards controlling lift trajectories of soft pneumatic actuators (SPA) through the design of soft membrane mechanical properties. Given a simple form of pressure control and a known set of external bodies, how do we design a membrane that will facilitate the required force and position throughout inflation? How do we add versatility to these platforms through strain limitation? 

In particular, we investigate low-cost, hyperelastic elastomer SPA. We attempt active strain limitation with design and control of electroadhesive clutches. We model passive trajectories using active learning with data from automated experiments and use these models for inverse design of SPA performing quasi-static lifts.

\section{Background}

\subsection{Patient movement} 

For the elderly or those with significant medical ailments, standard activities of daily living can become unattainable due to injury or atrophy. These persons become reliant on the physical help of nurses or caregivers to achieve motion. Tasks such as repositioning in bed, standing from a sit, and transferring from seat to wheelchair all require assistance from one or more caregivers and the afflicted loses their autonomy. In turn, nurses and nurses’ aides are particularly susceptible to injury.



Though equipment exists for patient handling and transfer, the use of such equipment still doesn’t provide autonomy or fully eliminate the opportunity for injury for the patient or caregiver. Current solutions include ceiling-mounted lifts, portable base lifts, and inflatable wedges, yet they all require caregiver effort in the form of guidance or assistance \citep{nelson2003}. 



With the work in this dissertation, we aim to enable a soft robot that can transform a simple pneumatic input into reliable, comfortable human motion. We see this assisting with acts of daily living, potentially relieving reliance on constant physical support from caregivers. Using pneumatics and a primarily soft robot will allow us to achieve the same level of safety and force distribution as seen in consumer technologies with considerably higher precision. We want aging populations to find comfort and mobility with minimal labor on the parts of their caregivers and, eventually, to restore their own autonomy.


From discussions with nurses (special thanks to Pam Cacchione), we have identified a variety of potential uses for soft pneumatic actuation in the medical setting: 
turning (rolling) patients; propping patients on their side; elevating a limb; preventing hip abductor motion; preventing lateral slump in chair; sit-to-stand from chair; rolling patient up (toward pillow) in bed; preventing pressure ulcers at heels; keeping heels off mattress (holes/lift); keeping head elevated at 30~deg.; padding siderails; preventing contracture in extremities; peristaltic massage for DVT prevention; heating/cooling a patient; and preventing foot drop (dorsiflexion). 

\subsection{Soft Robotics}

Soft robotic actuators use deformation to produce smooth, continuous motions \citep{robinson1999} and conform to delicate objects \citep{pettersson2010} while remaining resilient to adverse environmental conditions or critical loading \citep{Tolley2014}. Soft materials are generally inexpensive, light, and easily compactable for storage or transportation. These properties make soft actuators useful across a variety of applications including locomotion, gripping, manipulation, and haptic response \citep{laschi2016,yin2020}. They also enable the use of soft actuators for direct physical human-robot interaction in a medical setting \citep{jiralerspong2025}.


Due to the lack of rigidity within their structures, the kinematics of soft robots is often more complex than in rigid robotics \citep{laschi2014}. For example, the position of a robotic arm, can be completely defined by a finite number of DoF. These DoF depend only on the non-rigid parts of the robot (i.e., the actuators), which allows the arm to precisely position its end-effector in a large workspace with only 6 DoF. If we introduce flexibility to each limb, we would need to add one or more additional DoF to assume the state of the end-effector. This is the challenge we face when working with soft actuators, and the state-space only grows with geometric complexity.

\begin{table}[!ht]

\renewcommand{\arraystretch}{1.5} 
\setlength{\tabcolsep}{4pt} 
\centering
\caption{Soft pneumatic actuators with embedded trajectory response to pneumatic input.}
\begin{tabular}{
    |>{\centering\arraybackslash}m{3cm} 
    |>{\centering\arraybackslash}m{2.6cm} 
    |>{\centering\arraybackslash}m{4.6cm} 
    |>{\centering\arraybackslash}m{2.3cm} 
    |>{\centering\arraybackslash}m{2.5cm}|
}
\hline
\textbf{Description} & \textbf{Examples} & \textbf{Degrees of Freedom} & \textbf{\shortstack{Number of\\Materials}} & \textbf{\shortstack{Extensible/\\Inextensible}}\tablefootnote{We will consider extension and extensible materials in the same way Baines describes shape change \citep{Baines2023a}: "there is a valid non-identity stretch tensor \textbf{U} in the polar decomposition of the deformation gradient \textbf{F}." Simply put, if the relative distance of two points on a body changes then there has been material stretch and that body has extended. The difference between an inextensible pressure chamber and an extensible pressure chamber is the difference between an air mattress and a balloon. An air mattress goes from an uninflated state to an inflated state: the membrane unfolds but does not undergo substantial stretch. The membrane of the balloon, on the other hand, stretches, extended from an unfolded shape to a new, larger shape.} \\
\hline
\multicolumn{5}{|l|}{\textbf{Single Response}} \\
\hline
Pneunet & \citep{Tolley2014,mosadegh2014} & Bend & 1 + & E \\
Limited Tube & \citep{hawkes2016,connolly2017, skorina2018} & Extend/Bend/Twist & 2 & E \\
FREE / PAM & \citep{greer2017,sedal2021,sholl2021,thomalla2022} & Contract/Extend/Twist & 1-2 & I \tablefootnote{Though these are often made with extensible materials, they are characterized by their enclosure in inextensible fibers. This fundamentally differentiates them from limited tube/surfaces, which have additional opportunity for unrestricted expansion.} \\
Limited Surface & \citep{tomholt2020, pikul2017, feizi2022, ceron2018} & Extend/Complex & 2 & E \\
Bellows & \citep{rajappan2022,rogatinsky2022} & Extend/Bend & 1 & I \\
\hline
\multicolumn{5}{|l|}{\textbf{Variable Response}} \\
\hline
Multi-chamber Pneunet & \citep{marchese2016,balak2020,huang2020} & Extend \& Bend & 1 + & E \\
Vine Robot [\tablefootnote{Vine robot considered with additional external actuation.}] & \citep{hawkes2017,greer2017,blumenschein2020} & Extend [\& Bend/Complex] & 1 [ + ] & I 
\\
Externally Limited Surface & \citep{stanley2015,rauf2023} & Complex & 2 + & E \\
Externally Limited Tube & \citep{lee2016,Arleo2023,firouzeh2015,yang2021} & Extend \& Bend/Twist & 1 + & E \\
\hline
\end{tabular}
\label{tb: Lit Review}
\end{table}




To quantify the high-dimensionality of soft actuators, common morphologies appear to provide a repeatable response to change in internal pressure. One example of this is the tube, beam, or trunk actuator \citep{thomalla2022,Sedal2018,marchese2016,ansari2017}. These trunks can be discretized into a finite number of segment-connected nodes based on fabrication geometry \citep{marchese2016} or desired output geometry \citep{Baines2023}. Beams have been controlled in extension, contraction, and bending with differential inflation \citep{ansari2017,marchese2016}, external pneumatic artificial muscles (PAMs) \citep{greer2017}, local stiffening \citep{Baines2023,do2020,Arleo2023}, and tendons \citep{lee2016}. Other common geometries include a surface, which generally responds to positive pressure by growing from a plane into a three-dimensional shape \citep{pikul2017, forte2022}, and inextensible bellows which move towards a target, unfolded, configuration. Any of these geometries can provide varied response if strain limitation or antagonistic force is varied prior to or during inflation. An investigation into common single response and variable response SPA can be found in Table \ref{tb: Lit Review}.


\subsection{Active Control of Soft Pneumatic Actuators}

 Researchers have been able to facilitate real-time dynamic control of SPA, relying on precise control of air flow and a well-characterized system \citep{marchese2016}. In this type of system, the input control (i.e., pneumatic syringe) needs to be just as precise as in rigid robotics, and the combination of air and material become a series-elastic element with assumed forward kinematics. This makes the control problem more difficult, but it has been solved with various assumptions (i.e., beam mechanics, constant curvature, or geometry restrictions) regarding the discretized segment forward kinematics \citep{marchese2016a}. The methods and results in these works are incredibly impressive, but will be difficult to apply with untethered or low-cost pneumatics. Our work relaxes the assumption of precise pneumatic control and attempt to operate while in contact with external bodies, particularly when the hyperelastic region of the material constitutive properties come into effect. We look to the passive response of the material elements (quasi-statically) and control this material response via passive or active strain limitation to modulate the inflation trajectory.


\subsection{Soft Pneumatic Lifting}

When beam-shaped soft pneumatic actuators are used for heavy lifting, it is primarily with loading in the radial direction, and primarily with inextensible membrane materials. This is in part because compressive force application in the axial orientation is limited for this geometry by its slenderness and buckling becomes a critical issue \citep{thomalla2022}. The other reason to use this orientation is that the geometry of inextensible membranes is difficult to characterize under partial inflation (termed the `geometric regime' \citep{rogatinsky2022}). Using relatively small radii, these devices can limit their time spent in these transient positions, while the length of the beam is used to increase contact area rather than object displacement. Though effective, this common beam geometry and radial  orientation limits the workspace of the actuators to just the radius of the beams, preventing larger-displacements and generally limiting directional lifting. 

A more versatile soft actuator geometry for pneumatic lifting is an elastic membrane spanning an aperture - creating a surface. These actuators grow from a plane into shapes defined by embedded, (relatively) inextensible strain limiters. The shape of the membrane can be fully defined by the change in position of the inextensible elements and the strain energy of the extensible areas. The design complexity of these membranes comes with added shape control, and they have been used for visual displays \citep{stanley2015}, camouflage \citep{pikul2017}, and infrared light deflection \citep{tomholt2020}. When the stiffening elements appear in high enough density that the motion is controlled by their position as opposed to the extension of the elastomer, this forms a Fiber Reinforced Elastomeric Enclosure (FREE). FREE have been effectively used for force generation \citep{thomalla2022,sholl2021,sedal2021} in specific orientations\footnote{ It should be noted that among the FREE explored for force generation, the more slender geometries \citep{thomalla2022,sedal2021} were more completely characterized but also showed higher propensity for catastrophic (buckling) failure.}, but lack the adaptability potential of general elastic membranes.

There has been significant success in shape planning for elastic membranes \citep{pikul2017,forte2022}, but lifting remains elusive. We will pursue this goal both theoretically and experimentally, tying in strategies from the autonomous experimentation materials science community \citep{snapp2023} to increase testing throughput. We aim for an output model that can relate pressure, force, and displacement for a known contact area in the same way that we can for a pneumatic piston.



\section{Contributions}

This thesis offers two main contributions.
\begin{itemize}
    \item A novel means of soft actuator active strain limitation via electroadhesive (EA) clutching. These strain limiters can be applied on any silicone pneumatic actuator and (de)activated in real-time via electrical signal. This signal is substantially faster than thermal approaches \citep{firouzeh2015} and the system competes directly with other active strain limiters \citep{stanley2015,yang2021}. Further advances in the design and control of these clutch systems are presented, achieving  a variety of use-cases.
    \item A pipeline for creating lightweight models of interactions between pneumatic elastomeric membranes and external forces, applied in 1 DoF. This model begins by confirming a fundamental understanding of the material undergoing expansion, and grows to an empirical (learning-based) model that directly relates the  state variables relevant in our lifting problem. 
\end{itemize}

\section{Overview}

This thesis begins to inform passive, embodied responses as soft robots interact via external contact, towards physical interactions with human beings.

Chapter 1, the introduction, will motivate soft pneumatic robots and mechanical response and give an overview of some recent advances in this space. 

Chapters 2 and 3 will detail our attempts to leverage  active strain limitation for control of elastomeric soft pneumatic actuators (SPA). We show progress in trajectory control through embedded strain limitation but fall short of trajectory control in the presence of external forces. 

Chapter 4 will dive into the inclusion of external force, leveraging modern tools in active learning and automated testing to characterize design choices for a class of SPA. 

Chapter 5, the conclusion, will provide our outlook for the future of soft pneumatic actuation and potential future directions in modeling and lifting.

In the addendum, we include ongoing work in which we question the use of low-data learning for mechanical design as we apply Bayesian Optimization (BO) for the design optimization of soft pneumatic valves and 3D-printed gear boxes. We also include supplementary information and a detailed procedure for membrane fabrication.



\chapter{PROGRAMMABLE SHAPING OF ELASTOMERIC MEMBRANES}
\label{ch: Shape Morphing}
This chapter is adapted from:

Gregory M Campbell, Jessica Yin, Yuyang Song, Umesh Gandhi, Mark Yim, and James Pikul. Electroadhesive clutches for programmable shape morphing of soft actuators. In 2022 IEEE/RSJ International Conference on Intelligent Robots and Systems (IROS), pages 11594–11599. IEEE, 2022.\\
\copyright 2022 IEEE

My contribution to this work involved leading the design and implementation of the research approach, the design and fabrication of the hardware platform and associated software, experimental data collection, Vicon data post-processing, and the majority of the writing.\\

\section{Introduction}


In many soft robotic applications, each robot must be compliant, but still retain local stiffness to restrict expansion towards desired deformations or to apply directed forces. The need to control where a soft robot is compliant or stiff has led to extensive research into understanding how local stiffness affects robot actuation and control \citep{manti2016,rus2015,marchese2016}. 

\begin{figure}[!ht]
      \centering
    \includegraphics[width=0.8\linewidth]{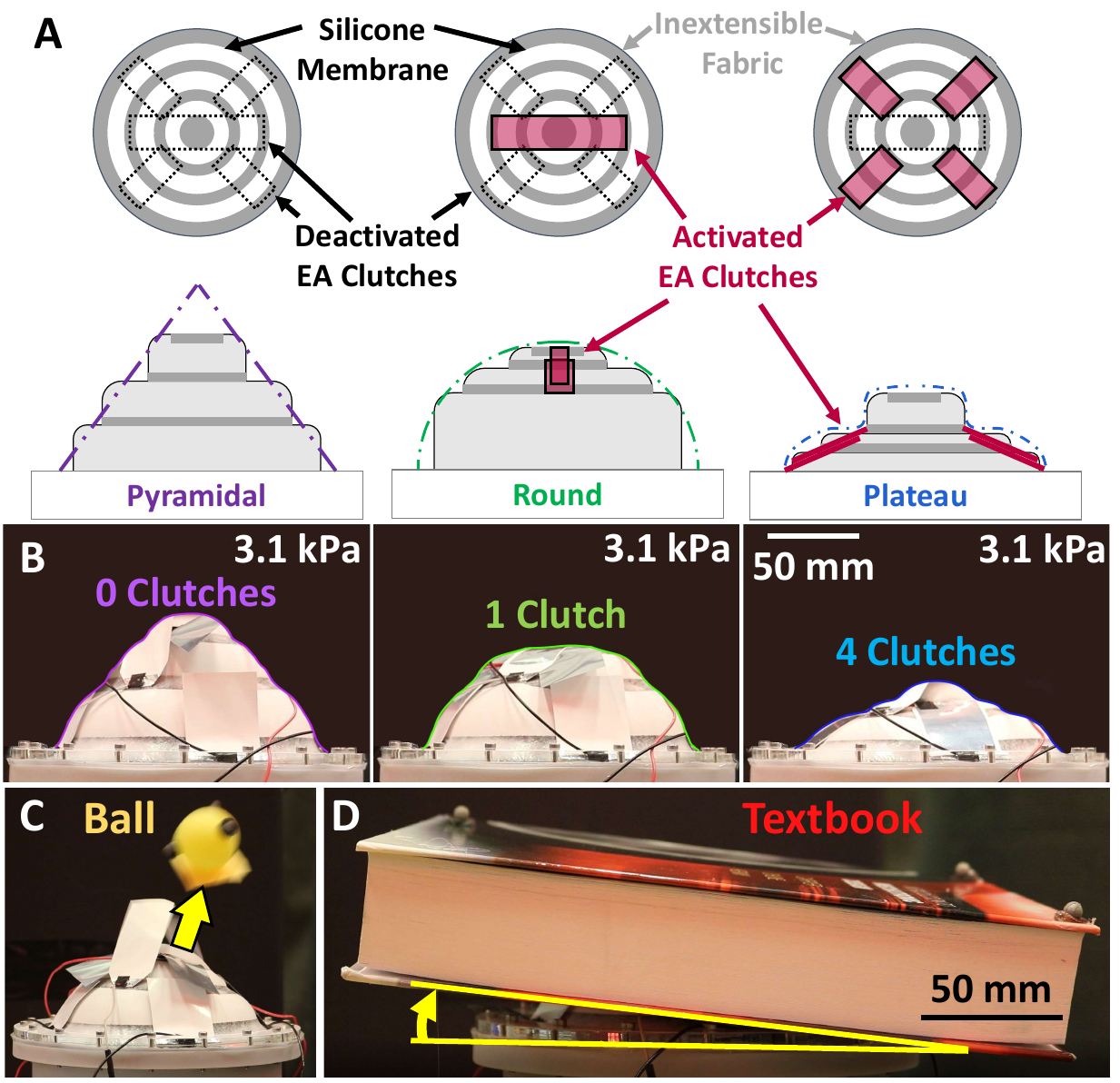}
      \caption{A. Actuator expansion formed into three different shapes; shape chosen by clutch activation. Top-view presented over side-view. B. Expansion of soft, elastomeric actuator to 3.1~kPa under three different clutch activation configurations. C. Soft actuator manipulation of 3.7~g ball. D. Manipulation of 820~g textbook.}
      \label{Figure1}
  \end{figure}
  
Most soft pneumatic actuators are constrained to deform along a single determined range of motion due to the static nature of their material composition. These single-configuration soft pneumatic actuators (SPA) are fundamentally limited in their ability to controllably adapt to new shapes, actuation modes, and forces. SPA are typically made from pressurized chambers or "pneu-nets" in elastomers, which influence the scale or speed of material expansion \citep{shepherd2011,mosadegh2014}. SPA generally use passive inextensible materials, such as fabric \citep{pikul2017} or fibers \citep{sholl2021},
to restrict expansion towards a single desired range of motion.  If an alternative range of motion is desired, an entirely new actuator must be manufactured. Achieving general 3-dimensional motion control with this type of SPA necessitates the use of multiple separate chambers working in tandem \citep{shepherd2011,balak2020}. These SPA therefore require multiple pressure sources and rely on rigid peripherals for electrical controllability across multiple degrees of freedom (DoF). An alternative approach is to actively program the stiffness of materials in soft actuators and, therefore, allow the range of motion to be modulated in real time with a single source of pressurized fluid.

Active stiffness modulation of soft materials has been explored with technologies such as granular jamming \citep{steltz2009}
and joule heating \citep{firouzeh2015}, 
both of which rely on long transition times (multiple seconds). Faster, electrostatic solutions have also been realized \citep{levine2021}, including dielectric elastomer actuators (DEA), which align opposite charges around a dielectric elastomer to reduce its thickness and increase its area. Electrostatic chucking \citep{imamura2017} stacks many layers of DEAs to increase stiffness, while DEA membranes \citep{Zou2013} 
use the DEA to decrease their already low stiffness. Electroadhesive (EA) clutches are an attractive alternative because they have been shown to modulate stiffness in tens of milliseconds while applying stresses over 8~N/$cm^2$ at 400~V \citep{Hinchet2019,hinchet2018,diller2018}. EA clutches apply opposite electric charge on two parallel pads, which are then adhered to each other and prevent relative motion. EA pads have been used to modulate stiffness for restricting human motion \citep{diller2016, Hinchet2019} and in soft robots to assist with grasping \citep{guo2018} 
or modulate physical connections \citep{Germann2012}. EA clutches present an opportunity for soft actuation, as they are lightweight, flexible yet inextensible, and require no off-board pumps or motors. 

This work demonstrates the potential of EA clutches to provide stiffness modulation and shape control for a single-chamber soft pneumatic actuator. By attaching clutches to a membrane and alternating clutch activation, we demonstrate the ability to vary stiffness and inflate into multiple shapes and curvatures. 
These curvatures are primitives for a broader set of complex shapes, as previously demonstrated for camouflage applications \citep{pikul2017}. We demonstrate the versatility of this electrical stiffness control in two modes of actuation along five DoF: expansion-driven manipulation of a \hbox{3.7 g} object and pneumatic manipulation of a \hbox{820 g} object. We use manipulation of a \hbox{3.7 g} object to characterize the actuator workspace, up to \hbox{12 mm} along the five DoF, and to assess the magnitude and consistency of force applied by local membrane expansion caused by deactivating a clutch, up to \hbox{3.2 N}. The actuator also tilts a \hbox{820 g} object by \hbox{5 degrees} via pneumatic inflation with clutch activation and rapidly repositions it via clutch deactivation. Using clutch deactivation instead of pressure increase to rapidly apply force allows this actuator to function untethered, with a small (4.5 V, 63 g) air pump supplying pressure and two batteries (5V, AA) as energy sources.

This work represents the first usage of EA clutches for soft actuator stiffness modulation and the first instance of real-time electrically-controllable soft membrane stiffness modulation for pneumatic force application. The actuator presented is mobile and versatile, lightweight and inexpensive, and provides both speed and strength. 
This work presents critical progress towards scalable, real-time variable, and broadly applicable soft actuators that achieve high strain and electrically-controllable actuation. 

\begin{figure}[th!]
\centering
\includegraphics[width=0.6\linewidth]{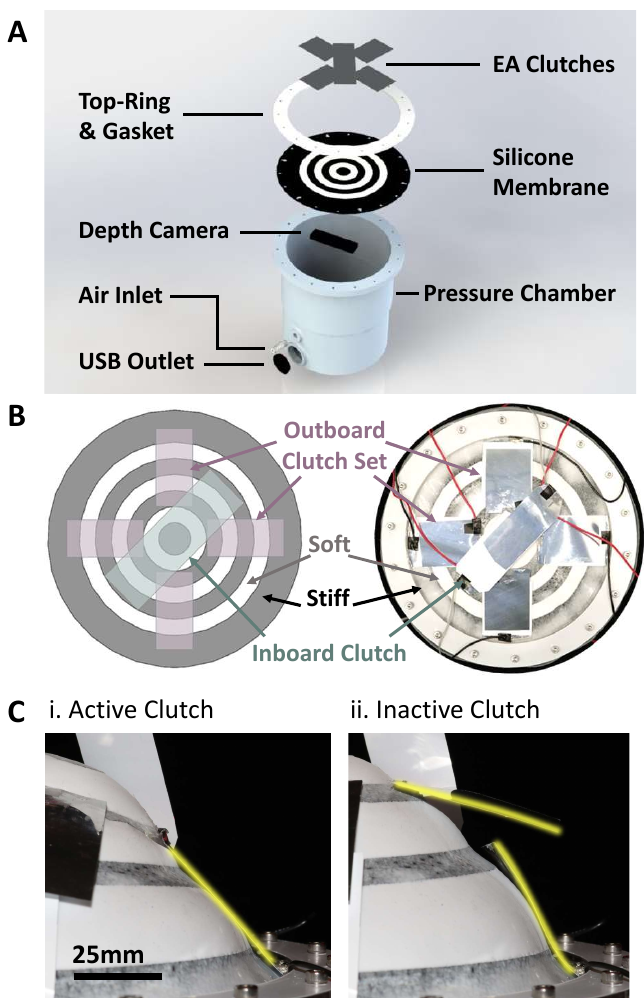}
\caption{System Overview. A. Overview of actuator test system with labels. B. Model of silicone membrane with clutch locations superimposed. Model includes stiff silicone reinforced with fabric, soft unsupported silicone, four outboard clutches, and one inboard clutch. C. Clutch is highlighted with yellow. i) Outboard clutch is activated and restricts membrane expansion. ii) Outboard clutch is deactivated and membrane is free to expand.}
\label{Membrane Layout}
\end{figure}

\section{System Development \& Fabrication}

Our soft membrane testing system (Figure \ref{Membrane Layout}A) inspired by \citep{alspach2019} consists of three main components: A) a pressurized chamber, B) the elastomeric membrane with EA clutches, and C) sensors and control electronics.

 
\subsection{Soft Membrane Test System - Chamber}
The pressurized chamber is assembled from two 3D-printed pieces. 
Both pieces are glued together using a resin epoxy adhesive. 
The chamber has two ports: an air inlet via a push-to-connect tube fitting and a dual USB 3.0 port for sensor data output. Both ports are sealed with silicone gaskets and resin epoxy adhesive. The elastomeric membrane is clamped down at the top of the chamber with a silicone gasket, laser-cut acrylic ring, and screws. The inner diameter of the top ring restricts the expanding area of the actuator to a diameter of \hbox{150 mm}. This diameter was governed by the size of the 3D-printer print-bed, but can be scaled up or down with the same clutch pattern and electrical controllability.


\subsection{Elastomeric Membrane and Electroadhesive (EA) clutches}

To make the membrane, we laser-cut Soft N' Shear fabric stabilizer and place it in Ecoflex 00-30 during curing to provide areas of high stiffness. We then connect EA clutches to these fabric-reinforced areas via Sil-Poxy silicone adhesive. 

We designed the Soft N' Shear stiff regions 
based on the zero Gaussian curvature layout from \citep{pikul2017}. The concentric rings create a pyramidal shape upon inflation
We space the rings such that there is a large enough stiffened region for adhering the EA clutch plates. We create three concentric regions of unstiffened silicone (Figure \ref{Membrane Layout}B) and control the activation of EA clutches to restrict the expansion of some subset of these three soft regions (Figure \ref{Membrane Layout}C). We can activate multiple clutches in parallel, which allows for pre-determined patterns to activate with a single signal. For the shape change configuration (Figure \ref{Membrane Layout}B) the red, outboard, clutches are activated by one signal, while the green, inboard, clutch is activated by another. Clutch and fabric positioning can be altered on subsequent membranes for separate types of applications.

The EA clutches use Dupont Luxprint 8153 as a dielectric. The parameters of Luxprint clutches are well characterized \citep{diller2018} and provide sufficiently high forces at hundreds of volts relative to other dielectric options \citep{Hinchet2019,Hinchet2019}. Following the fabrication process specified by Diller et al. \citep{diller2018}, we applied a nominally 50 $\mu m$ layer (actual thickness varies from 40-50 $\mu m$) of Luxprint to nominally 50 $\mu m$ thick aluminum-sputtered biaxially-oriented polyethylene terephthalate (BOPET).
 We cut the resulting sheets into clutch plates and attach wires with MG Silver Epoxy 8331. Electrical tape provides electrical insulation over wire leads and prevents delamination of the Epoxy. Clutches are activated at a DC voltage of approximately \hbox{420 V}. 

\begin{figure}[b!]
      \centering
      \includegraphics[width=0.3\linewidth]{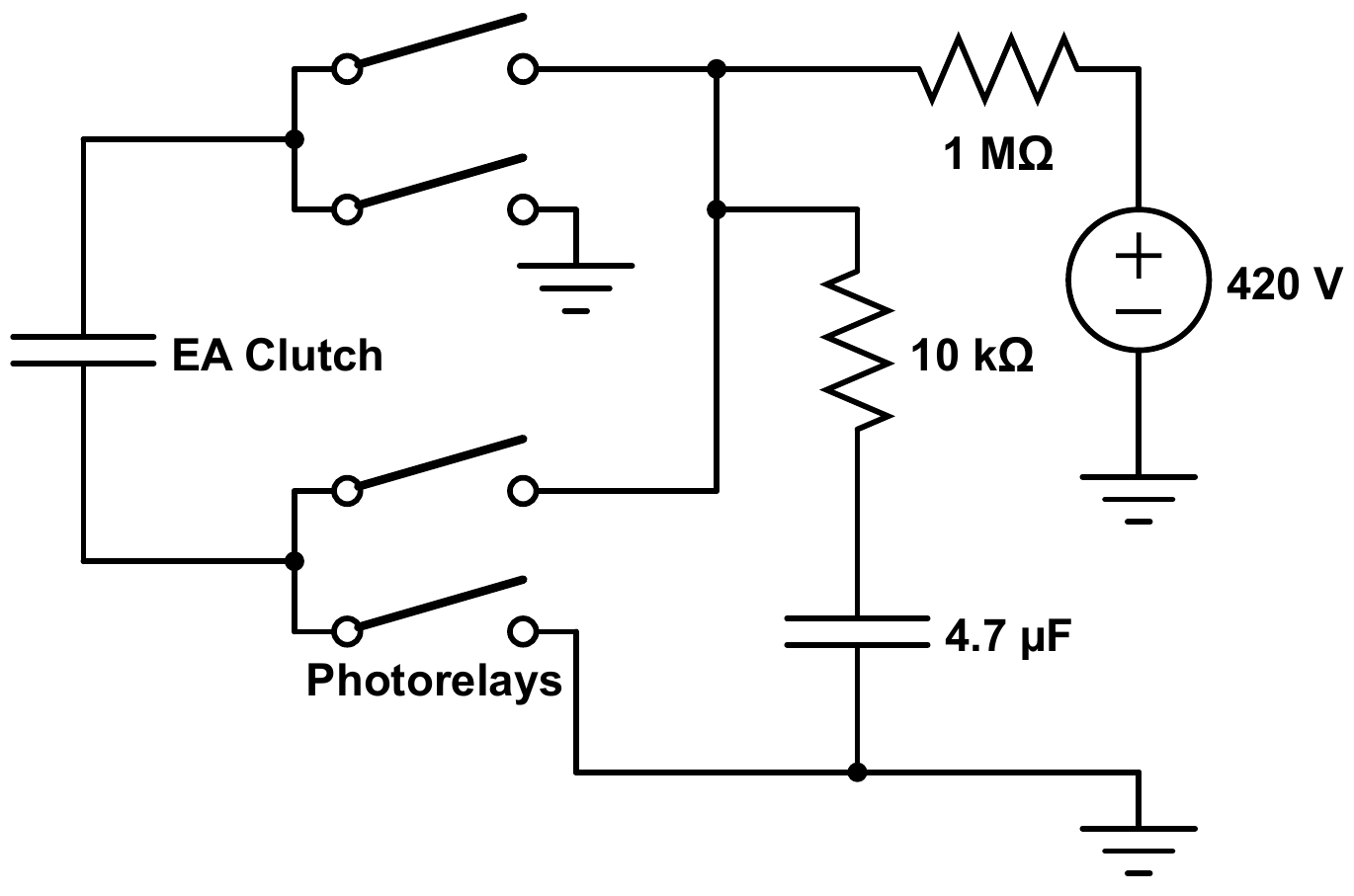}
      \caption{Circuit Diagram. Clutch circuit diagram, building upon Diller et al.'s \citep{diller2018}.}
      \label{Circuit}
\end{figure}

\subsection{Electronics}
The pressure chamber houses an air pressure sensor (Qwiic MicroPressure Sensor; Sparkfun), an ESP32 microcontroller, and a time-of-flight depth camera (Picoflexx; PMD). The air pressure sensor monitors the internal air pressure and connects to the microcontroller to output data through the USB port. The depth camera captures membrane deformation data by measuring the displacements of the membrane from inside the chamber and passes data through the USB port. 

External to the pressure chamber, a ZR370-02PM \hbox{4.5 V} DC air pump and vacuum (dimensions: 58x27x27 mm, mass: 63 g) powered by a 5~V battery applies air pressure. The electronic clutch control circuit (Figure \ref{Circuit}) includes TLP222G-2 photorelays which switch to bring one clutch plate to the high (\tildehack420 V) voltage \citep{diller2018}. $4.7 \mu F$, 400~V capacitors supply  additional current in parallel with the high voltage supply during activation transitions. Using two relays allows  each clutch to switch polarity. Switching polarity  between tests counteracts space-charge effects, enabling reliable deactivation. This circuitry was scaled up to allow for the simultaneous or subsequent activation of multiple clutches. An ESP32 microcontroller controls inflation and clutch (de)activation with inputs from a laptop computer. Similarly, the laptop records pressure data and depth images. An EMCO F101CT DC/DC converter connected to a AA battery supplies power for the clutches.



 \begin{figure}[tb!]
      \centering
      \includegraphics[width=0.7\linewidth]{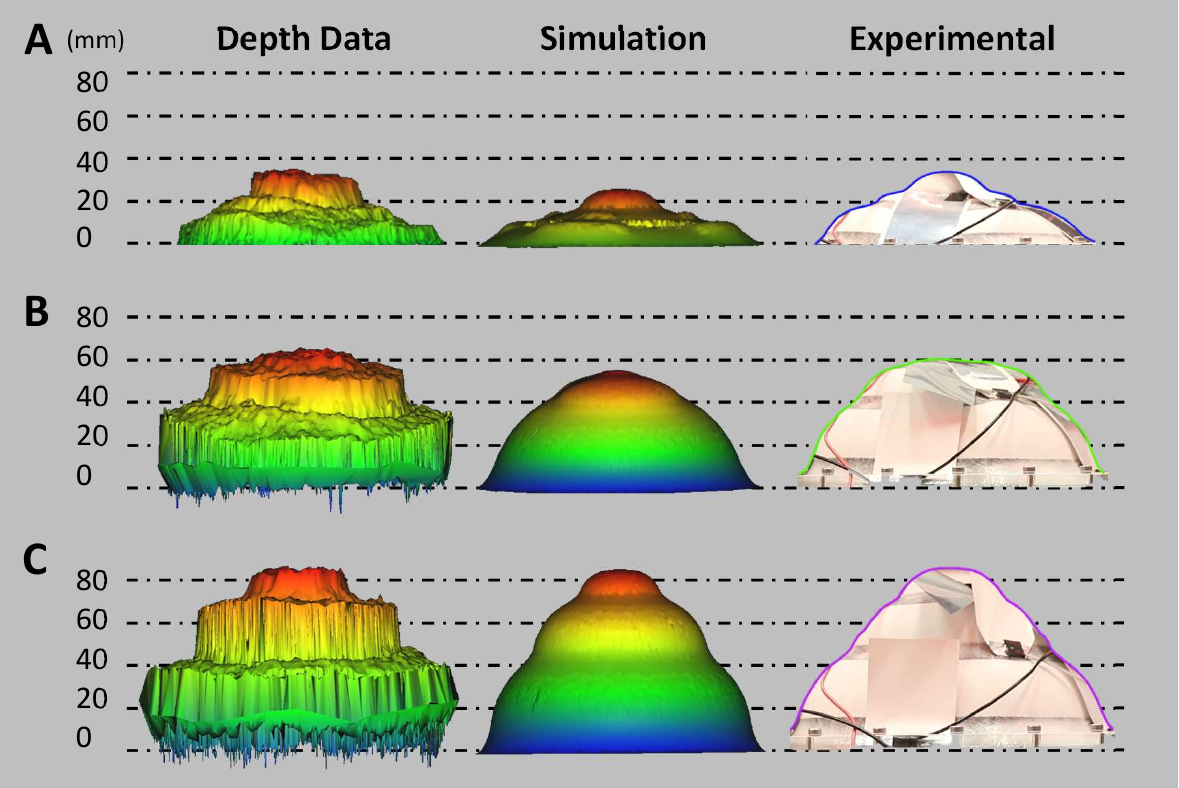}
      \caption{Depth camera data (left), simulation (center), and experimental system (right) for actuator shape comparison at 3.1~kPa. A. Plateau shape. B. Round shape. C. Pyramidal shape.}
      \label{FEA_Results}
      
\end{figure}
\section{Results \& Discussion}
  \begin{figure*}[t!]
      \centering
      \includegraphics[width=\textwidth]{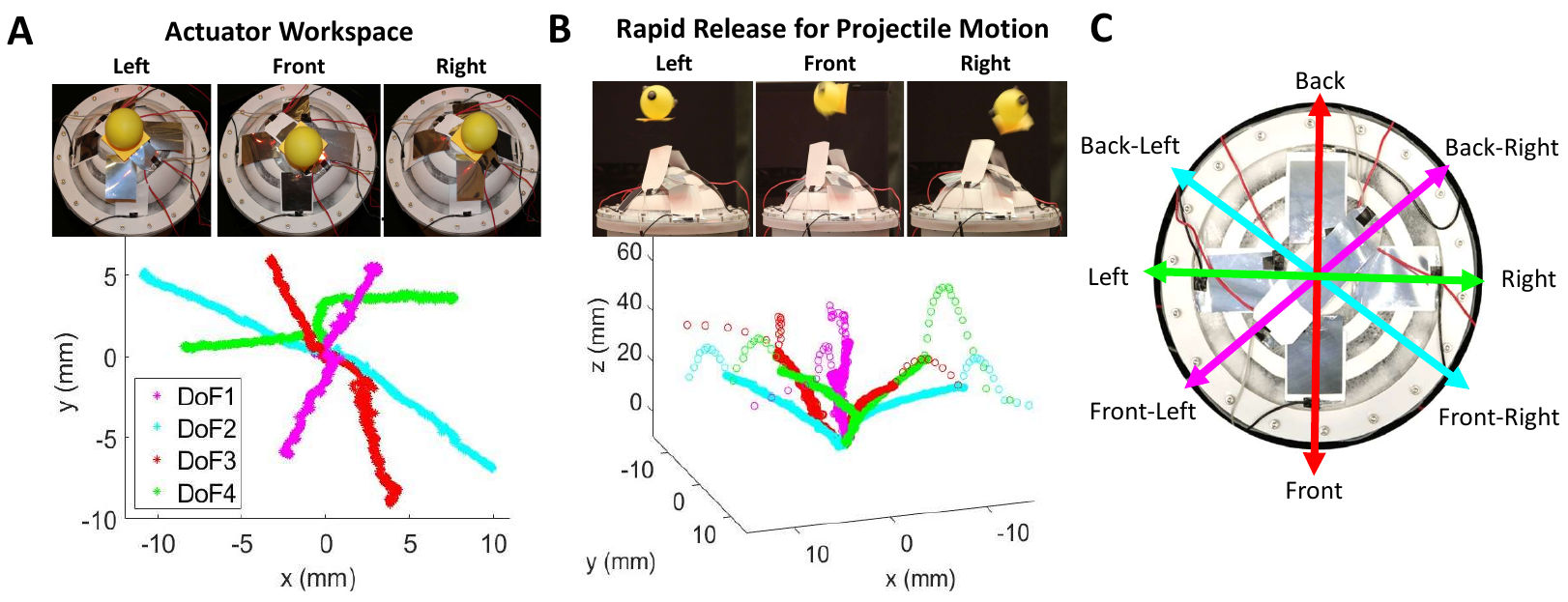}
      \caption{Mode 1 Actuator Characterization. A. Plot of actuator workspace along 4 DoF, looking at the actuator from a top-down view. Each DoF is accessed by activating a different set of clutches. Photos of corresponding positioning of 40 mm diameter ball. B. Plot that tracks position of ball once it has been placed on the actuator. The actuator lifts it to the specified direction and then releases the inboard clutch to rapidly apply force to the ball in the desired direction. Photos of corresponding projectile motion for 40 mm diameter ball. C. Labeled photo of actuator with directions and DoFs.}
      \label{Actuator-Characterization}
  \end{figure*}
\subsection{Finite element modeling}

We use the finite element (FE)  package ABAQUS 2020/Standard \citep{Systemes2020/Standard} to understand how the inflatable structure behaves under different clutch constraints. In this analysis, we model the silicone inflatable structure as a circular shell discretized into S4R element types. 
The thickness of the silicone shell and the attached clutches match the experimental setup at 1 mm and 0.2 mm, respectively. 
  
Pneumatic pressure is applied normal of the top surface of the membrane, while the edge of the structure is fixed with a zero-displacement constraint in all 6 DoF. Based on the membrane design, we assigned each portion of the ring to either Ecoflex 00-30 or Soft N' Shear fabric stabilizer (high stiffness). 

We model the Ecoflex's hyperelastic behavior using a 3-term Ogden model 
\citep{Toolkit}, Soft N' Shear areas using elastic material with a modulus of 8~MPa, and the clutch, including the two individual clutch plates and adhesive connections, using elastic material with a modulus of 100~MPa.
The clutch is attached with perfect bonding to the membrane over the entirety of the regions marked in Figure \ref{Membrane Layout}B. This differs from the experimental system, in which the clutches  bond to the membrane only at the ends.
The simulation runs dynamically, without mass scaling. The output displacement contour provides position values.

Our modeling predicts that thin clutches can provide significant change in  membrane inflation shape when inflated to 3.1~kPa (gauge pressure), as seen in the center column of Figure \ref{FEA_Results}. Four thin clutches are positioned in the outboard positions for Figure \ref{FEA_Results}A, one clutch is positioned at the inboard position for Figure \ref{FEA_Results}B, and the stabilized silicone is simulated without any clutches in Figure \ref{FEA_Results}C. The maximum height the membrane reaches for each simulation respectively is: \hbox{26.5 mm}, 57.1 mm, and 86.3 mm.

\begin{table}[t!]
\centering
\renewcommand{\arraystretch}{1.4} 
\caption{\normalsize Forces and Standard Deviations for Each DoF and Direction.}
\begin{tabular}{cccc}
\hline
\textbf{DoF}           & \textbf{Direction} & \textbf{Force (N)} & \textbf{Std (N)} \\ 
\hline
\multirow{2}{*}{DoF 1} & Back-Right          & 1.2   & 0.21  \\ 
                      & Front-Left          & 1.4    & 0.28  \\ 
                      \hline
\multirow{2}{*}{DoF 2} & Front-Right         & 1.1    & 0.41 \\ 
                      & Back-Left           & 1.4    & 0.19  \\ 
                      \hline
\multirow{2}{*}{DoF 3} & Front               & 2.4   & 0.35   \\ 
                      & Back                & 2.7    & 0.72     \\ 
                      \hline
\multirow{2}{*}{DoF 4} & Left                & 3.2      & 0.57      \\ 
                      & Right               & 2.2      & 0.71       \\ 
                      \hline
DoF 5                  & Up                  & 2.8      & 1.1       \\ 
\hline
\end{tabular}
\label{tab:force_table}
\end{table}

\subsection{Experimental Results}

The left column of Figure \ref{FEA_Results} shows the time-of-flight depth data of the membrane as we inflate it to \hbox{\tildehack3.1~kPa} with three different clutch configurations. Prior to and throughout inflation to the plateau shape (Figure \ref{FEA_Results}A), we apply the voltage difference of \tildehack420~V across the four outboard clutches (red rectangles in Figure \ref{Membrane Layout}B), restricting the expansion of the two outboard rings of unsupported silicone. The resulting expansion therefore occurs primarily at the innermost ring of silicone, leading to a height of 32.7 mm. Prior to and throughout inflating the membrane to the round shape (\hbox{Figure \ref{FEA_Results}B}), we activate only the interior clutch. When fully inflated, the membrane reaches a height of 66.7 mm. We activate no clutches for the pyramidal shape (Figure \ref{FEA_Results}C), leading to an inflation height of 86.3 mm.

To compare the experimental and simulation results, we align the point clouds using the Iterative Closest Point algorithm and calculated the average root-mean squared error (RMSE) for each shape with CloudCompare \citep{GirardeauMontaut2016}. We use standard point cloud pre-processing techniques on the experimental data, including statistical outlier removal and the k-Nearest Neighbor noise filter. Average RMSE for the plateau, round, and pyramidal shapes respectively were \hbox{4.4 mm}, 6.1 mm, and 6.3 mm.
  
\subsection{Actuation}

We leverage shape-changing characteristics of the membrane into two separate modes of actuation, exemplified in Figure \ref{Actuator-Characterization} and Figure \ref{Book} respectively. Mode 1 emphasizes the fast responsiveness of the actuator enabled by the electric signal control for EA clutches, while Mode 2 emphasizes the high force density of pneumatic actuation, while still providing active control via the EA clutches. Between expansions, the membrane is deflated by opening the pressure chamber to the ambient air and clutches are re-aligned by hand.

\subsubsection{Mode 1}

Mode 1 allows for the manipulation of light objects that have minimal effect on the shape of the membrane. We activate clutches relevant to the desired motion at the start of inflation, and after inflation is complete, we deactivate the inboard clutch to apply a rapid force in the desired direction.

We evaluate the actuator's workspace by altering clutch configurations in Mode 1. We induce motion along cardinal directions by activating a single outboard clutch and the inboard clutch, and ordinal directions by activating two adjacent outboard clutches and the inboard clutch. We position a light payload (ping pong ball, Vicon markers, and paper for stability - mass: \hbox{3.7 g}) along one of five different DoF.
We monitor ball position in three dimensions via Vicon cameras while inflating the membrane from 0 to \hbox{1.7 kPa} in separate trials for each cardinal and ordinal direction. The actuator displaces the ball between 6 and \hbox{12 mm} along each degree of freedom as displayed in Figure \ref{Actuator-Characterization}A. This represents up to \hbox{16 \%} of the 75 mm radius. The fifth and final degree of freedom, `Up', results from no clutches active and is in the \textit{z} plane, perpendicular to the plot in Figure \ref{Actuator-Characterization}A. 

After the membrane is inflated, we deactivate the inboard clutch, which causes the locally restrained membrane to rapidly "soften." This leads to a rapid expansion of the elastomer, which applies a force on the object in a direction corresponding to the actuator shape. Figure \ref{Actuator-Characterization}B shows subsequent projectile motions that appear as parabolas beginning at the edge of the actuator workspace. Positions are plotted at 0.01 second intervals. Sparser data indicates faster motion induced by clutch deactivation. 

We characterized force output separately from workspace. In this characterization, the membrane was inflated to \hbox{2.8 kPa} for cardinal directions (and `Up') and \hbox{1.7 kPa} for ordinal directions. Outboard clutches were subject to failure at pressures higher than this. Data from trials involving clutch failure is not included. We then deactivated the inboard clutch, causing the force response discussed for \hbox{Figure \ref{Actuator-Characterization}B}. We calculated force based on the motion of the ball as follows: we calculated the components of acceleration for the ball in each direction (\textit{x},\textit{y},\textit{z}) from the second derivative of its position with respect to time and identified the instantaneous acceleration most relevant to the actuator force. We adjusted the  z component of acceleration by summing it with the acceleration due to gravity. We calculated magnitude of force as the product of the mass of the ball and the $L^2$-norm of the three components of acceleration. We conducted five trials for each direction\footnote{Only three trials were conducted in the `Up' direction.}, and report the averages and standard deviations of force magnitudes in Table \ref{tab:force_table}. 


\begin{figure}[t!]
      \centering
      \includegraphics[width=.4\linewidth]{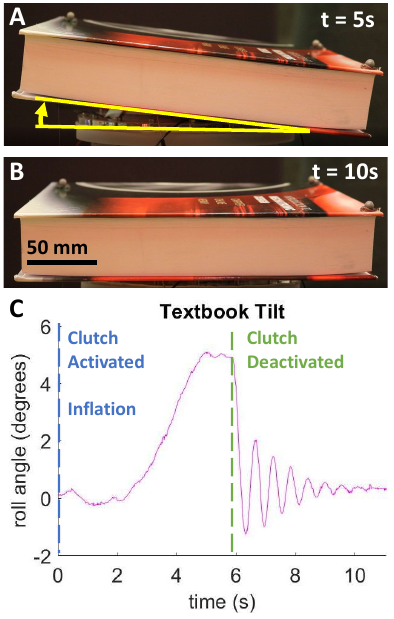}
      \caption{Mode 2 Manipulation. A. Textbook lifted and tilted by pneumatic inflation to 3.1~kPa during right clutch activation. B. Textbook returned from tilt by clutch deactivation. C. Vicon data for textbook roll angle response to inflation under clutch activation and to clutch deactivation.}
      \label{Book}
\end{figure}

\subsubsection{Mode 2}

Mode 2 allows for the manipulation of heavier objects that alter the shape profile of the membrane. 
These types of object cannot be instantaneously accelerated away from the membrane as in Mode 1. We activate the desired outboard clutch(es) prior to inflation, which causes a tilt during the object lift. Subsequent to inflation, deactivating outboard clutches provides a rapid actuator shape change. 

We inflate the actuator to 3.1~kPa with the right clutch active to bring the roll angle of the textbook (textbook with Vicon markers - mass: 820~g) to 5~degrees (Figure \ref{Book}A). We subsequently deactivate the EA clutch, allowing the mass of the textbook and material properies of the inflated membrane to return the roll angle to approximately zero (Figure \ref{Book}B). Vicon data displays the textbook's roll angle versus time in Figure \ref{Book}C, with inflation leading to a gradual increase in roll angle and clutch deactivation leading to a rapid decline in roll angle and an underdamped return to equilibrium.

\subsection{Discussion of Results}

We validate all three simulated shapes with high accuracy using the proposed experimental setup. The plateau shape represents local areas of negative Gaussian curvature and the round shape represents positive Gaussian curvature. The combinations of these different curvatures allow a range of different positions within the actuator workspace, which reaches up to \hbox{12 mm} along various degrees of freedom. Different combinations of membrane and clutch layouts could be fabricated to reach myriad different shapes and motions.

FEA results differ from experimental data by a RMSE ranging
from 7~\% of maximum height in the least constrained case to 13~\% of maximum height in the most constrained case. The maximum heights ranged in error from 0~\% to 19~\%, with the larger errors again occurring the in the most constrained case. These errors may be due, in part, to the fidelity of the depth camera data. But the maximum height error, at least, is more likely due to limitations of the FEA itself. We are forced to make a number of assumptions during the analysis, including the interaction (adhesion) between the clutch and the silicone and the non-slip properties of the clutches themselves. As SPA systems increase in complexity of material and external interactions, these types of assumptions will become harder to make accurately and we will likely need to focus more on experimental methods.

The negative Gaussian membrane condition proved the most difficult to capture due to clutch holding force limitations. The EA clutches were susceptible to slipping when faced with surface debris or out-of-plane deformations near their peak holding force. To combat surface debris, we wiped down clutch plates with rubber (HandPRO glove) prior to each usage. To combat out-of-plane deformation, we took particular care when attaching the clutch plates to the membrane, and cut slits on the edge of the top clutch plate near the wire attachment point to relieve tension from the wire. While these fabrication difficulties do not prevent the usage of clutches for stiffness modulation, a self-enclosed system that accounts for both surface cleanliness and plate lineup would greatly increase the ease of use of this system, and perhaps increase its fidelity relative to the FE model.

Discrepancies between experimental and simulation results occur primarily at clutch-restricted regions. These discrepancies could be due to the model's assumption that the clutch plates are fully overlapped at activation, and that no slipping occurs between activated clutch plates at any point during inflation. Both these assumptions can change in reality. Nonlinear behavior of the Soft N' Shear stiffening agent is ignored in the simulation, which could also contribute to discrepancies. The close correlation between the modeled and experimental results shows promise for the use of FEM with EA clutch placement on a soft membrane.

The ability to induce projectile motion of the 3.7 g ball during Mode 1 manipulation shows that this actuator is capable of rapid directional force application at the edges of its workspace. This type of explosive force could be useful in a lightweight mobile robot, particularly for a jumping motion. The inconsistent ranges of motion and force applications across different degrees of freedom are likely due to imprecisions in membrane manufacturing (we aligned clutches by hand). 

Without any reconfiguration, the actuator can then actuate in Mode 2 to position the 820 g textbook pneumatically and rapidly reposition it with clutch deactivation. Figure \ref{Book} represents a single trial, and Mode 2 was significantly less consistent between trials, as small offsets in the  position of the textbook's center of mass greatly altered the trajectory. Comparably sized Ecoflex pneumatic membranes have been shown to lift up to 5.84 kg \citep{sholl2021}, but we did not lift more than 820 g for this study. This lifting capacity could be increased with a larger actuator or by using more actuators in parallel. While this system won't reach the force output or precision of a rigid platform, its natural compliance can provide safety and comfort in the application of human contact forces. 

\section{Conclusions \& Future Work}

This work demonstrates EA clutches as a viable means of stiffness modulation for inflatable soft actuators. Furthermore, it shows that we can use clutches to alter the inflation of a single soft membrane to accurately recreate three target shapes and manipulate both light and heavy objects along five degrees of freedom while supplying air only from a small, low-voltage, DC pump. A light object manipulation workspace is defined for the actuator, and forces are characterized at the edges of this workspace. This actuator characterization and shape changing demonstration, along with the ability to predict and design for shapes with modeling, lays the groundwork for more advanced pneumatically actuated soft robots. This technology, applied locally on a larger membrane or on a series of membranes working in parallel, could provide a mechanism for powerful, soft manipulation.

There are a number of remaining challenges to be overcome in future work. The actuator needs to be made more robust for real-world use, as failure frequently occurs at clutch lead connections and membrane-clutch adhesive interfaces. EA clutch reliability and robustness need to be addressed to eliminate the need for careful alignment prior to clutch activation and regular cleaning of clutch plates between trials. If clutch reliability were addressed, this actuator could be controlled in more arbitrary degrees of freedom via time-dependent clutch activation during inflation. This actuator could also benefit from more precise control, specifically closed-loop clutch deactivation based on Vicon camera or internal depth camera data, which would allow for greater consistency during \hbox{Mode 2} manipulation. Refined \hbox{Mode 2} manipulation could provide direct human contact forces in a sit-to-stand assistance device,  bed-ridden patient manipulator, or other ergonomic mechanism.

\section{Lessons Learned}
While this project showed that clutches can be implemented on SPA with some interesting effect, it also demonstrated many of the limitations of EA clutches for soft systems. Though naturally flexible, EA clutches are strongest when they have rigid backing and limited motion out of plane \citep{levine2021}. In our use-case, the clutches have only the BOPET backing and are constantly being pushed out of plane by the actuator's inflation. Also, the clutch plates needed to be manually reset between trials, and contamination was a major issue. These issues are examined in detail in the next chapter to advance clutches for soft robotics. Significant effort was put into additional silicone sheathing to provide some stabilization, protection, and pretension to the clutches. Sheathing was ultimately abandoned for this system but is being used successfully in other applications (see Chapter \ref{chap:sheathed}). 

Directional control was limited, and further analysis revealed that we were only able to adjust direction due to the stiffened area and its fastening to the pressure chamber. The clutches acted as tendons connecting directly to the stiff area, restricting motion in a curve with a radius equal to the combined length of the clutch and stiff material (insert images). Though it wouldn't have increased the workspace in the same way, it would be interesting to see the effects of attaching similar clutches in a FREE pattern like the ones used by Sholl et al. \citep{sholl2021}. I imagine varied clutch activation could be applied to differ between pyramidal and circular shapes with underlying structures that differ more than those achieved in this work, which are more akin to turning different tiers of the structure on and off.

The energy build-up and release during Mode 1 manipulation was particularly interesting. Though relatively little energy was captured in this example, more energy could be captured in stiffness or more complex systems that constrain the clutches only to their direction of travel \citep{krimsky2024}.

In the spirit of applying these technologies for external force application, the greatest downfall of this work was its inability to alter shapes whilst supporting substantial loadings (beyond those shown in Fig. \ref{Book}). This is because increasing complexity of shapes provides diminishing benefit in increasing the range of object manipulation. We want to be able to simplify the transform between pneumatic input and body output, as is done for so many of the actuators referenced in Table~\ref{tb: Lit Review}. It is this realization that influences the direction of Chapters~\ref{chap:sheathed} and \ref{ch: Membrane_Design_Section}, in which we simplify the actuator shapes down to a single degree of freedom.


\chapter{SHEATHED CLUTCHING FOR SOFT ROBOTS}
\label{chap:sheathed}

This chapter is adapted from:

Gregory M Campbell, R Daelan Roosa, Kevin Turner, James Pikul, and Mark Yim. Control
of silicone-sheathed electrostatic clutches for soft pneumatic actuator position control. In
2024 IEEE 7th International Conference on Soft Robotics (RoboSoft), pages 299–304. IEEE,
2024.\\
\copyright 2024 IEEE

My contribution to this work involved leading the design and implementation of the research approach, the design and fabrication of the hardware platform and pressure-related software, theoretical model implementation, application data collection and post-processing, and the majority of the writing.\\

\section{Introduction}
Soft pneumatic actuators (SPA) often inflate along a single determined range of motion that is prescribed by the design of its material composition \citep{shepherd2011,mosadegh2014}. These single-configuration SPA are fundamentally limited in their ability to controllably adapt to new shapes and actuation modes when they rely on passive inextensible materials, such as fabric \citep{pikul2017} or fibers \citep{sholl2021},
to restrict expansion. If an alternative range of motion is desired, a new actuator must be manufactured or the structure must be manually altered \citep{kim2019}. 

\begin{figure}[!t]
      \centering
    \includegraphics[width=\linewidth]{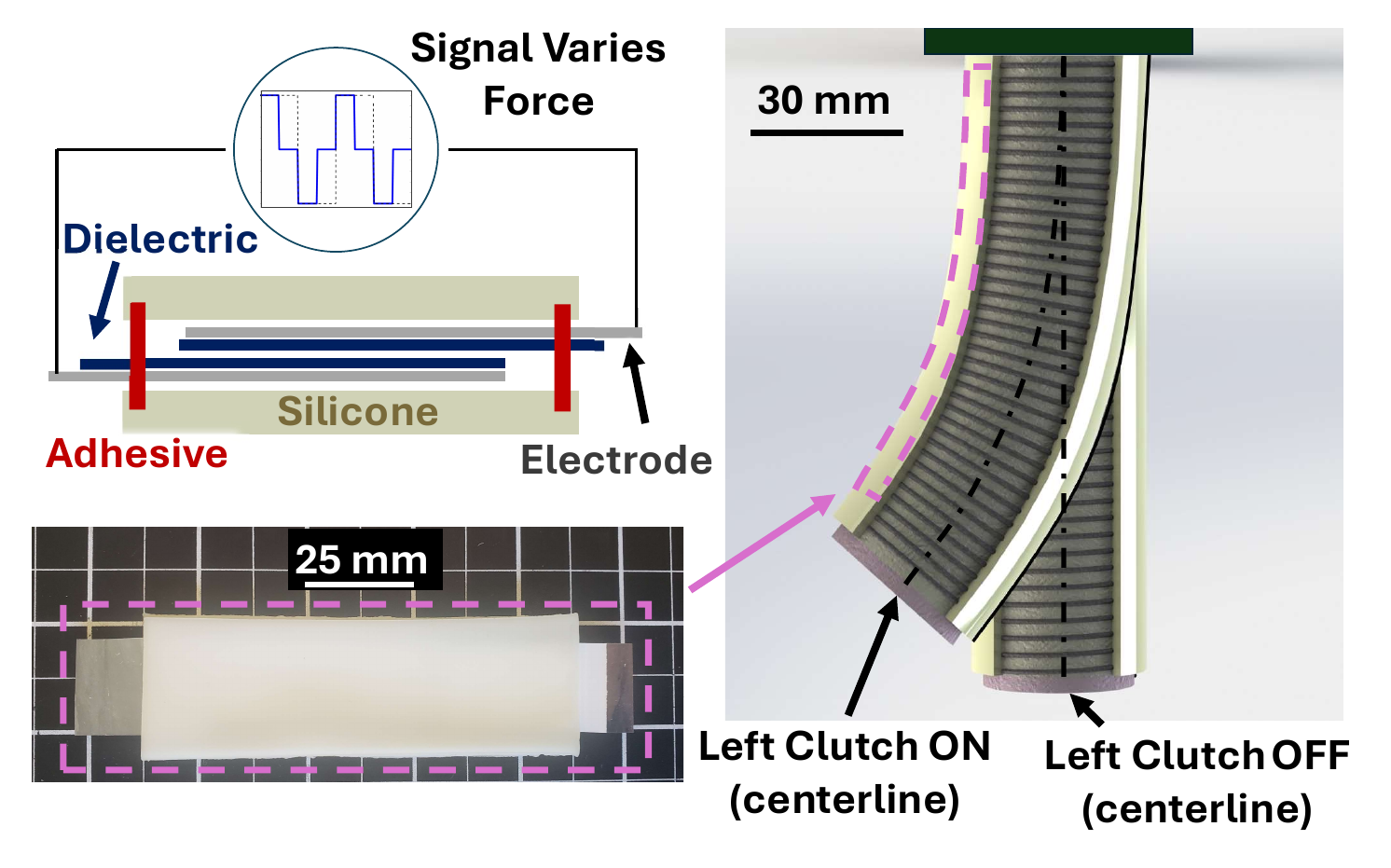}
      \caption{(left) Silicone-sheathed electrostatic clutch. On-state properties are controlled electronically via pulse-width modulation of high voltage square-wave. System consists of a variable silicone layer connected with Sil-Poxy adhesive to a 50 $\mu m$ electrode layer of AL-sputtered BOPET with a 25~$\mu m$ layer of Luxprint dielectric. (right) Antagonistic clutch system attached to the left side of a SPA to provide position control during inflation. Inflation with fully activated clutch superimposed over inflation with deactivated clutch.}
      \label{Figure1_sheathing}
\end{figure}

Other groups have applied antagonistic forces by actively programming the stiffness of surrounding materials \citep{yang2021,firouzeh2015} or applying external forces such as adding tendons \citep{lee2016}. The resulting systems go through various inflation trajectories depending on the activation of the antagonistic devices and corresponding strain limitation. A trunk-shaped actuator can be converted from a single degree-of-freedom (DoF) to a 3-DoF workspace with the inclusion of three antagonistic materials spaced around its circumference \citep{yang2021}. 

External strain limitations often require stiff materials to provide mounting points on the SPA \citep{lee2016}. An alternative approach is to mount the antagonistic components in sleeves, which can distribute the forces along the SPA \citep{Galloway2013}. Passive sleeves have allowed for significantly varied kinematic behavior \citep{Sedal2018} and pressure-dependent friction \citep{devlin2020}. However without active components, these sleeves have yet to allow for real-time modulation of shape or position. 


Electroadhesive (EA) clutches are strong, low-profile components that  offer adaptability for soft robots without compromising their flexible nature \citep{campbell2022}. EA clutches consume less power than motor-driven tendons while providing electrical control that is faster and more energy efficient than stiffening approaches that rely on phase change \citep{levine2021,diller2018,Hinchet2019}. They are stronger than fluidic or jamming-based solutions, with some clutches holding over 20~$N cm^{-2}$ \citep{Hinchet2019}, and avoid complications of additional pumps, compressors, valving, or tubing. However, EA clutches have binary (active/inactive) control authority. While this is valuable to rapidly move between discrete states \citep{campbell2022}, it is not always useful for applications where force application persists through the extension of the clutches. Force applied during motion utilizes sliding friction force between the clutch plates and is dependent on the electrostatic force of the clutches. If the clutch force can be modulated, then this sliding friction force is also variable.  To function in this `variable friction' regime, Hinchet et al.~added an additional sliding layer on top of their dielectric and varied the magnitude of their input voltage \citep{hinchet2022}. However, in another recent work by Feizi et al. \citep{feizi2022}, a similar effect is achieved by introducing a form of pulse-width modulation (PWM) into the square-wave input. In this way, the need for an additional sliding layer and change in the magnitude of voltage was avoided. This offloading of complexity from the physical system to the electrical control can also benefit soft systems, which already contain a substantial amount of physical complexity.

EA clutches are sensitive to external contamination, and their holding force is reliant on precise orientation. This is often overcome in practice by surrounding the soft components with rigid structures to ensure functionality \citep{feizi2022}. In applications that rely on fully compliant robots, this is not an option. Instead fabric sleeves have been employed to provided protection and clutch pretension \citep{hinchet2022}. While fabric sleeves are a natural solution for human interaction, they are not always easy to integrate into SPA. Silicone-based strain limiters, on the other hand, have been shown to integrate well into silicone-based SPA \citep{kim2019}. They also allow for design versatility, as silicone can be cast in a wide range of shapes with highly variable thickness to accommodate actuator needs. The Ecoflex used in our system provides superior stretch compared to fabrics, matching the strain of the SPA, while alternative silicones could be used to provide greater rigidity. 

This work demonstrates the design and modeling of a sheathed-clutch system that can be easily implemented as an active sleeve for silicone SPA. These sleeved clutches provide antagonistic tension forces that are varied using PWM of electrical inputs. We demonstrate fabrication strategies for contaminant-resistant EA clutch systems that can be adhered directly to silicone robots. We model the silicone sheathing to choose appropriate  stiffness for our applications and use a varied control strategy (similar to \citep{feizi2022}) to alter the force response of our clutch system in real time. We then demonstrate the utility of this clutch system in a dynamic catching task that relates increased clutch force energy dissipation to a reduced displacement throughout the catch. Finally, we demonstrate endpoint position control of a pneumatic prismatic joint actuator. 




\section{Modeling}
Designing a sheathed clutch for specific force applications requires an understanding of the holding forces in both the clutch and the silicone sheathing.
Clutches have been well modeled \citep{diller2018,levine2021}, and both fracture and friction models have been used in clutch design \citep{levine2022}. We will assume a friction-based response, specifically with the inclusion of an air gap as described in Section \ref{EA_Theory}:
\begin{equation} \label{clutch_friction_eqn}
F_{clutch}=\frac{\mu\cdot\epsilon _0\cdot  A}{2}\cdot \left(\frac{\epsilon _r\cdot V}{d}\right)^2
\end{equation}
where $\mu$ is the coefficient of friction between clutch pads, $d$ is the distance between the pads, $\epsilon_0$ is the permittivity of free space, $\epsilon_r$ is the relative permittivity of the EA pad dielectric, $A$ is the surface area of the electrode, and $V$ is the potential difference between clutch plates.

Modeling hyperelastic elastomers depends on assumptions about both material response and boundary conditions. We will assume  a diagonal deformation gradient \textbf{F} (only principal stretches $\lambda_1, \lambda_2, \lambda_3$), the incompressibility of silicone and uniaxial expansion ($\lambda = \lambda_1, \lambda_2=\lambda_3 = \frac{1}{\sqrt{\lambda}}$), and homogeneous, path independent (hyperelastic) expansion. Due to the relatively high stiffnesses of adhesives and clutch plates, we consider stretch only in the length of Ecoflex between adhesives.


Our model uses the Gent strain energy equation \citep{gent1996}: \hbox{$W = \frac{-\mu J_m}{2} ln(1-\frac{I_1-3}{J_m})$}. Where W is the strain energy, $\mu$ is the shear modulus, $I_1 = \lambda_1^2 + \lambda_2^2 + \lambda_3^2 = \lambda^2 + 2/\lambda $ is the first invariant (the trace) of the left Cauchy-Green Deformation tensor ($\mathbf{B} = \mathbf{F}\mathbf{F}^T$), and $J_m$ is an empirically solved material property defining the max extension of the molecular chains.


Our assumptions yield the following, simplified, constitutive relations for principal stresses:
$\sigma_i = \frac{\delta W}{\delta \lambda_i} \lambda_i+p$. Here $p$ is an undefined hydrostatic pressure responsible for enforcing incompressibility, and \textbf{I} is the identity matrix. Using the lack of restriction in the non-axial directions of sheath expansion ($\sigma_{2}=\sigma_{3}=0$) to solve for $p$, the final relation between $\sigma_{1}$ and $\lambda$ is:
\begin{equation}
\label{Stress-Stretch-Uni}
    \sigma_{1} = \frac{\mu J_m}{J_m-I_1+3} (\lambda^2- 1/\lambda) 
\end{equation}

With the relationship between stress and stretch fully defined by material properties $\mu$ and $J_m$, the force is dependent on these properties and \hbox{silicone cross-section area, $A$:} $F_{sheath} =  \sigma_{11}A = f(\lambda, \mu, J_m, A)$. Using superposition, we can define the tensile antagonistic force of the sheathed clutch system simply as the following:
\begin{equation} \label{Combined_Force}
    F_{max} = F_{sheath} + \zeta F_{clutch}
\end{equation}
where $F_{sheath}$ is fully defined by our design, but $F_{clutch}$ will be controlled via an electrical input varying the variable $\zeta$.

By implementing PWM control, we are able to change $\zeta$ from a binary variable to an empirically determined analog value between 0 and 1 and adjust the force response to our needs without altering the electrostatic pressure in Eq.~\ref{clutch_friction_eqn}.

\section{System fabrication and control}

The two components of the system, the silicone sheathing and the biaxially-oriented polyethylene terephthalate (BOPET) clutches, are fabricated separately and bonded together with Sil-Poxy silicone adhesive. While we fabricate planar orientations in this work, they also function in a partially curved (`tubular') state similar to  \citep{sun2022}, as seen when they are adhered to a cylindrical actuator (Fig. \ref{IPAM Results}).

\subsection{Silicone Surface Properties}

Platinum-cure silicone can be formed into arbitrary flat shapes with gravity molding, but softer silicones (in, particular EcoFlex 00-30) are relatively tacky, resulting in a  high coefficient of friction when sliding along the BOPET backing of an EA clutch. If untreated, this high friction affects off-state stiffness and causes buckling in the clutch plates. Cornstarch (baby powder) is a commonly used lubricant for silicone systems, and applying it to the contact surface of the silicone greatly reduces this friction. However, EA clutches rely on friction as well, and the presence of powder lubricants contaminates the plates so they cannot hold large forces. 


Chemical alteration of the silicone surface properties allows for reduced surface friction without negatively affecting clutch performance. 
Oxidation of the cured silicone followed by fluorination was found to significantly lower the friction coefficient. However, Smooth-On's SLIDE surface diffuser was found to be easier to apply and even more effective. This simple silicone treatment enables the effective use of clutches in contact with the silicone surface. We add 3\% by weight of SLIDE to all silicone  used for sheathing, and then allow the silicone to rest outside its mold for at least 24 hours prior to adhering clutches. In this work we use 30x100mm rectangles made from 5-10mL of \hbox{Ecoflex 00-30} per piece. 

\begin{figure}[ht!]
\centerline{\includegraphics[width=0.9\textwidth]{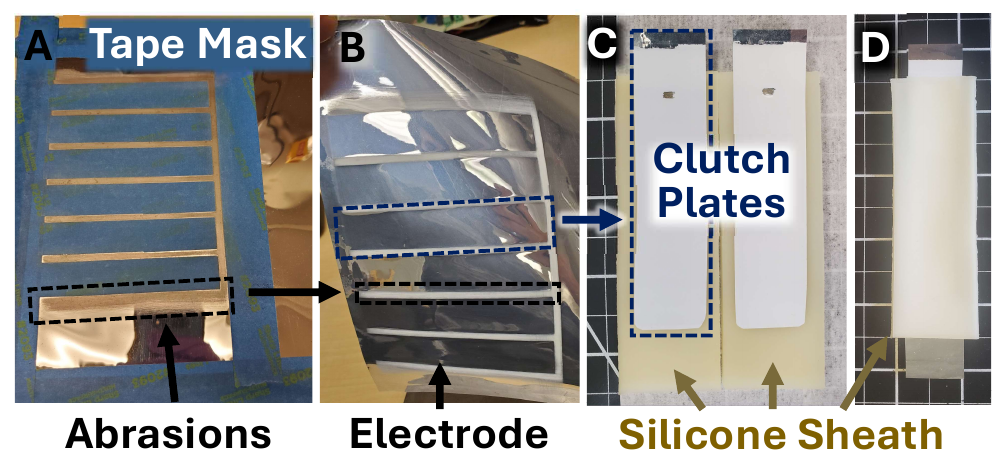}}
\caption{Clutch fabrication process: A. Masking and abrading the aluminum-sputtered BOPET. B. Resulting clutch sheet ready to be cut (back of sheet). C. Cut clutch plates on silicone sheathing with holes for adhesive contact. D. Assembled sheathed clutch.}
\label{Clutch_Fab}
\end{figure}

\subsection{Clutch Fabrication}

We implement the clutch fabrication process from \citep{diller2016} with an added abrasion step to remove metallization at the edges of the clutch to prevent shorting across the edges of the plates. We apply painter's tape (blue in Fig.~\ref{Clutch_Fab}A) to the active electrode regions and abrade the aluminum from the surrounding BOPET with a Scotch-Brite pad. We then coat a $25 \mu m$ film of Dupont Luxprint 8153 over the entire sheet. After curing, the clutch areas are clearly distinguishable and ready to be cut (see Fig.~\ref{Clutch_Fab}B). We apply copper tape directly to the edges of clutch electrodes that remain free of Luxprint to make electrical connections to the clutch.

\subsection{System Synthesis}

We mold two rectangular pieces of silicone per system, one for each clutch plate. We adhere the clutch plates to their respective silicone near the exposed electrode, with most of the clutch plate free to slide relative to the silicone. Once the adhesive has cured, we scrape a small area of Luxprint and aluminum off the clutch plate to expose the BOPET underneath (Fig.~\ref{Clutch_Fab}C). A thin layer of adhesive is applied to this exposed BOPET and the nearby silicone for each plate. These areas are then adhered to the exposed silicone of the other piece, creating a sandwich as seen in Fig.~\ref{Figure1_sheathing}. After curing, we seal the two pieces of silicone around the remainder of their circumference with a thin layer of Sil-Poxy to keep contaminants away from the clutch plates.

\begin{figure}[ht!]
\centerline{\includegraphics[width=0.8\linewidth]{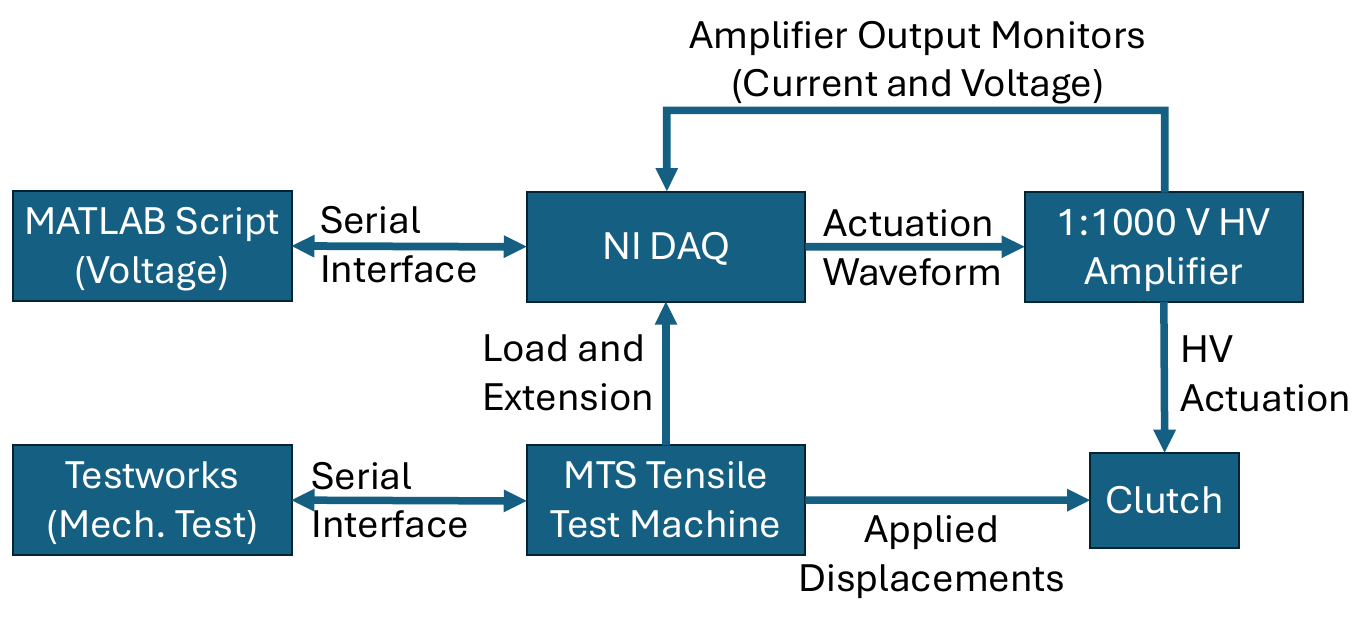}}
\caption{Clutch Control: Block diagram of electrical control for simultaneous high-voltage PWM and tensile testing}
\label{Clutch_Control}
\end{figure}

\subsection{PWM Clutch Control}

The clutches are powered with a bipolar 300~V amplitude square wave, and pulse-width modulation is applied to that square wave to create an actuation waveform (Fig.~\ref{PWM-Force} inset). 
This waveform is calculated in MATLAB, which interacts with a National Instruments (NI) Data Acquisition (DAQ) system via serial communication. That NI DAQ creates the electronic waveform, which is then amplified 1000 fold by a high volt (HV) amplifier before reaching the clutch plates. This process, and its use in the material testing shown in \hbox{Fig. \ref{Clutch_Response},} is represented by the block diagram in Fig.~\ref{Clutch_Control}.

\section{Sheathed Clutch Characterization}

\begin{figure}[ht!]
\centerline{\includegraphics[width=0.85\textwidth]{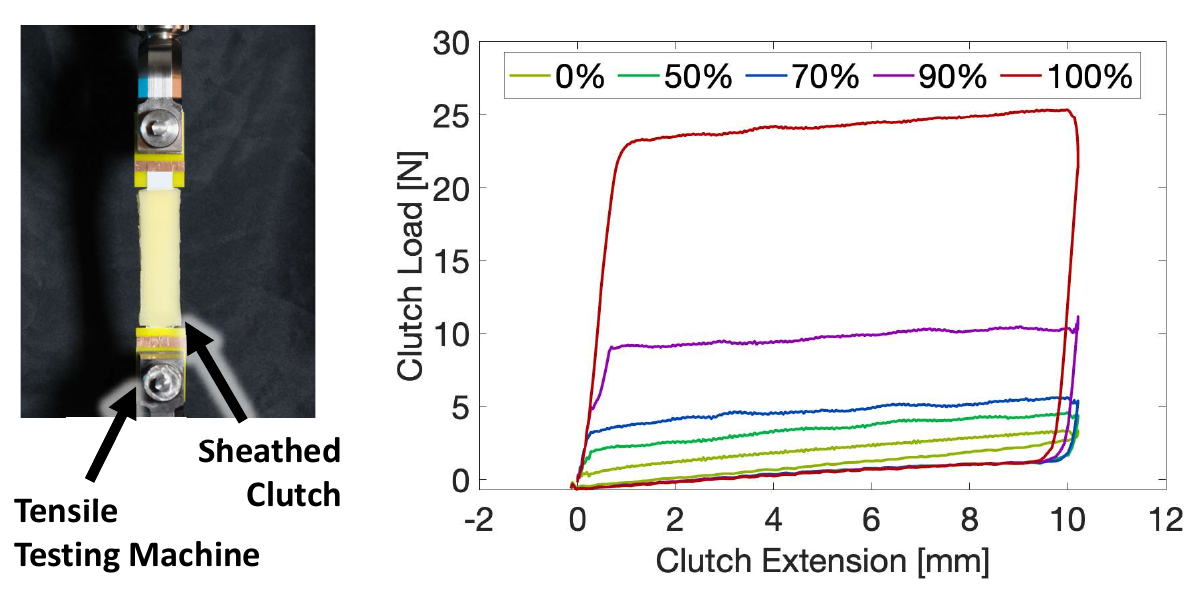}}
\caption{(left) Experimental setup for clutch force characterization. (right) Clutch force response during loading and unloading at various PWM duty cycles (duty cycle indicated by color).}
\label{Clutch_Response}
\end{figure}

The variance between on-state and off-state force-displacement relations is measured by testing the system in a uniaxial test system (MTS Criterion). The clutches are adhered to acrylic plates, which are mechanically clamped to the MTS. Clutches are activated with the high-voltage input and then pulled at a constant rate of 250~mm/min. Displacement continues as the system slips and reaches a steady-state force (subtracting the force contributed by the sheathing). When the system reaches a displacement of 10~mm, the clutch is unloaded at 250 mm/min back to its original length. By comparing this force-displacement curve (Fig.~\ref{Clutch_Response}) to the force-displacement curve of a deactivated clutch, we isolate the component of the system load borne by the clutch, and can thus measure the clutch holding force.

In a full-power loading at 300~V, the maximum tensile holding force of the clutch system reaches about 25~N. 
At this extension, 10~mm, the sheathing contributes approximately 3~N and the clutch contributes approximately 22~N. While there is a theoretical decrease in holding force as the clutch plates move past one other (decrease $A$ in Eq. \ref{clutch_friction_eqn}), the positive slope of the load-extension curve throughout loading in \hbox{Fig. \ref{Clutch_Response}} implies that the stress of the silicone contributes more force than the clutch loses over the tested range.



We determine the empirical relationship between activation duty cycle and holding force ($\zeta$ in Eq.~\ref{Combined_Force}) for each PWM duty cycle ranging from 0 to 100\% in increments of 10\%. 
Example cycles with incrementally increasing duty cycles are shown in Fig.~\ref{Clutch_Response}. The holding force was found to vary non-linearly with activation duty cycle, as seen in Fig. \ref{PWM-Force}. 

\begin{figure}[ht!]
\centerline{\includegraphics[width=0.75\textwidth]{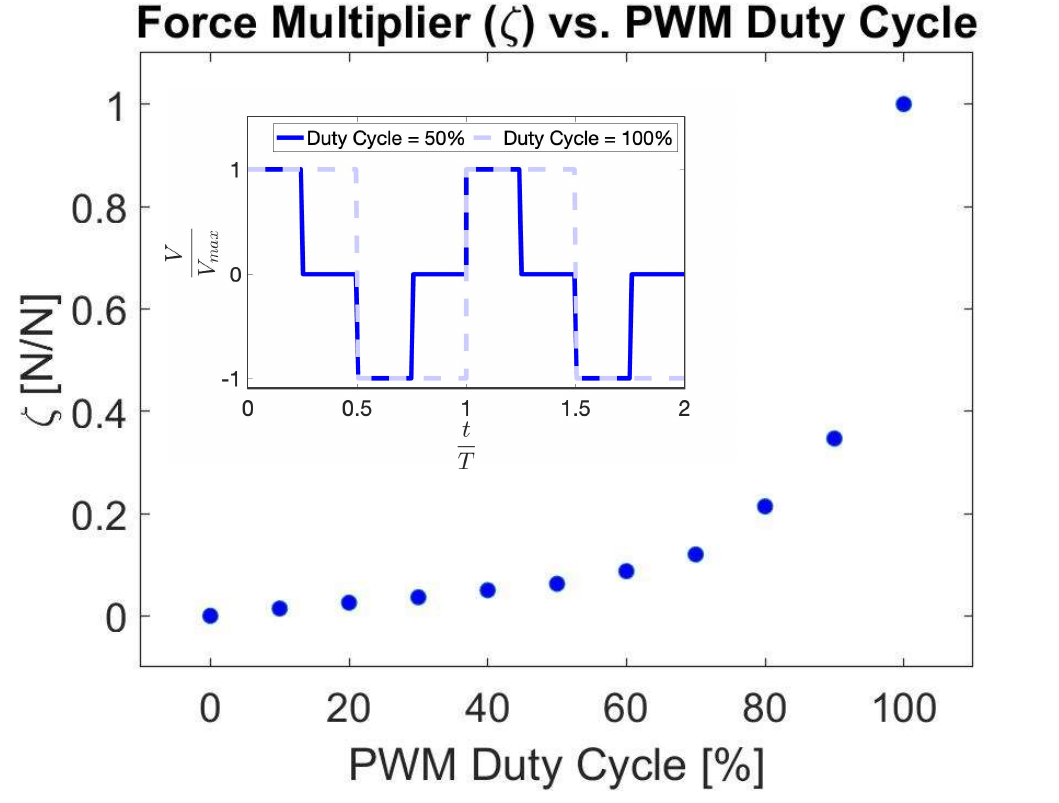}}
\caption{Empirical relation between clutch force multiplier from Eq.~\ref{Combined_Force} ($\zeta$) and PWM duty cycle. (inset) Example waveform for PWM output at 50\% and 100\% duty cycle.}
\label{PWM-Force}
\end{figure}

\section{Applications}
\begin{figure}[t!]
\centerline{\includegraphics[width=0.85\linewidth]{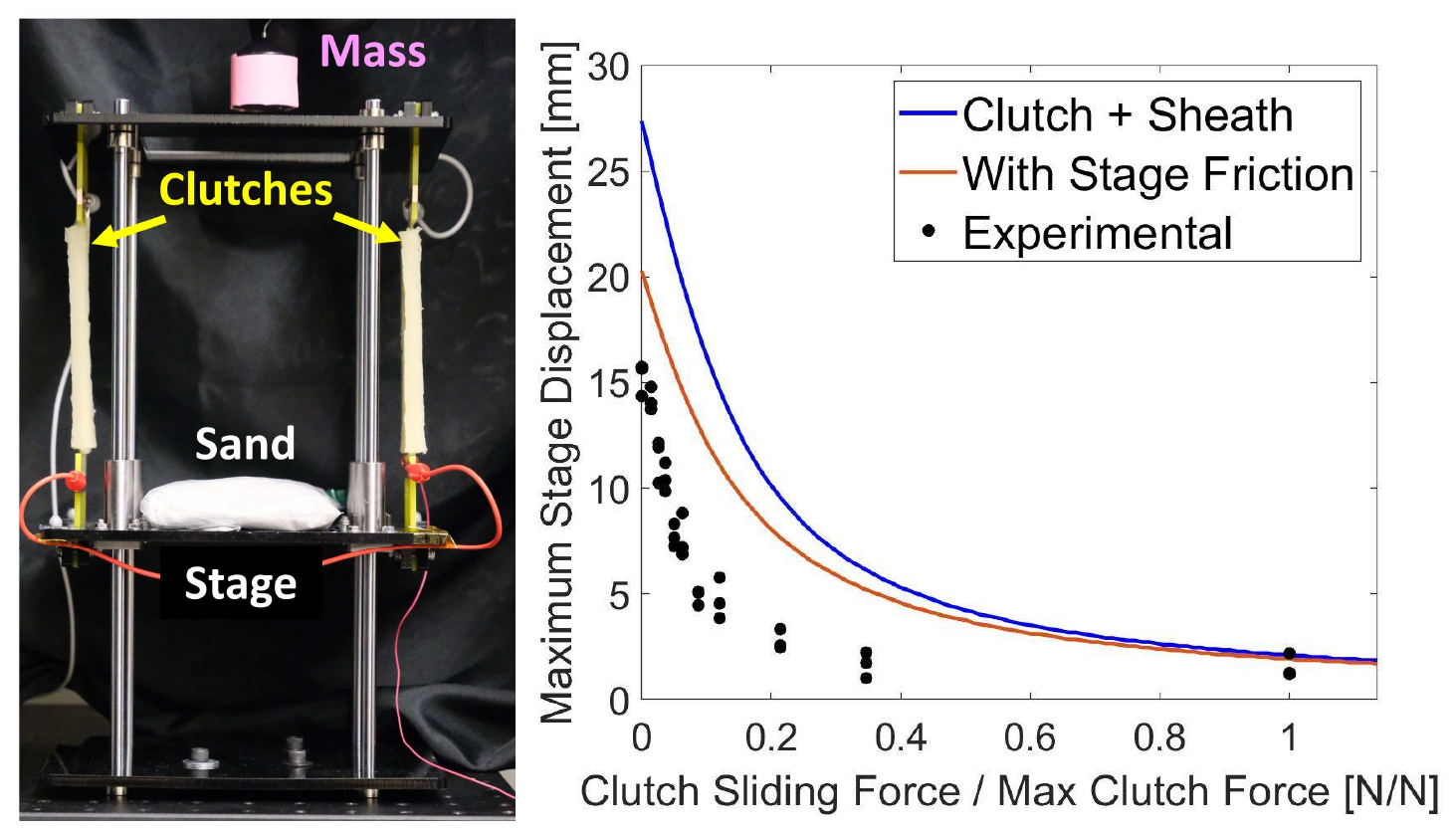}}
\caption{(left) Experimental setup for dynamic energy dissipation with sheathed clutch. (right) Maximum stage vertical displacement as a function of clutch activation force as modeled via energy methods (Eq.~\ref{Stage-Energy}, lines) and seen in experimental results (points). The blue line estimates dynamic response based on modeling of the clutch and sheath, while the orange line adds the work done by a constant force stage friction.}
\label{Catching_Setup}
\end{figure}

\subsection{Variable Non-conservative Energy Dissipation}
\subsubsection{Experimental Setup}
To validate our theoretical and empirical modeling, we test the clutch in a highly constrained, highly dynamic catching task. We constructed a 1 degree-of-freedom (DoF) test stage that is restrained against gravity by two sheathed clutches. The clutches are aligned such that the stage DoF is out-of-plane from the clutch plates. The clutch systems dissipate energy throughout the motion of the stage. Modulating this energy dissipation alters the distance through which the stage drops during a catch. The sheathing is pre-tensioned by the weight of the stage (\tildehack700~g), bringing the system to an initial steady state. We use two 3~mm thick pieces of Ecoflex 00-30 for the sheathing such that they provide enough tensile force to minimize initial clutch system elongation and thereby maintain clutch overlap area.  

We drop a 200~g mass onto the stage from a height of 200~mm and measure the displacement of the stage in the room's reference frame using a VICON motion-capture system. Displacement is measured relative to the position of the stage when the mass is dropped with downward as positive. The mass lands on a bag of microsphere sand to ensure that it remains in contact with the stage. Three trials are preformed for different activation duty cycles of the clutch system, ranging from 0\% to 100\% duty in increments of 10\%.

\subsubsection{Modeling}
This stage system is modeled as a simple inelastic collision (zero coefficient of restitution), where the momentum of the mass is transferred into the mass and stage as they move together after impact. With deactivated clutches, all the kinetic energy goes into the sheathing (plus system deformation and friction), which provides an underdamped, oscillatory, response. If the clutches are activated, they act as a non-conservative decelerating force whose magnitude is limited by the duty cycle of the PWM. The kinetic energy ($KE$) of the stage system can be described as a function of the stage's downwards displacement, $d$, and trends towards zero due to the nonconservative work done by the clutches and stage friction:

\begin{equation}
\label{Stage-Energy}
    KE_{Stage} = KE_{initial} - E_{friction} - E_{sheath} - E_{clutch} \\
\end{equation}
\[ = m \cdot g \cdot h_0 \cdot \frac{m}{m_{total}} + m_{total} g d   - E_{friction}  \] \[- (E_{sheath(d)}-  E_{sheath(initial)})  - d F_{clutch} \]

\noindent Where $m$ is the mass of the dropped mass, $m_{total}$ is the sum of $m$ with the stage mass, $h_0$ is the height above the stage from which the mass is dropped, $g$ is the standard acceleration of gravity, and $F_{clutch}$ is the product of its max clutch holding force with $\zeta$.
Energy ($E$) in the sheath is the integral of the sheath force, $F_{sheath}$, over the displacement: $E_{sheath} = \int_{d_0}^{d} F_{sheath} \,dz $. Published values of material constants \citep{steck2019} for incompressible EcoFlex 00-30 were used in sheath energy calculations such that the calculations could be used in sheath design.

The initial kinetic energy of the system is modeled as an inelastic collision, with momentum conserved. The initial kinetic energy of the mass and stage is therefore the gravitational potential energy of the mass reduced by the factor $\frac{m}{m_{total}}$ through this collision. Friction is modeled as a constant, restrictive force and estimated to be 2.5~N based on system measurements. 

For a damped response (no oscillation), enough energy has to be dissipated in the clutches ($E_{clutch}$) that the restorative force from the sheathing is less than the gravitational force of the combined mass of the stage and dropped mass ($m_{total}$) when the energy (Eq.~\ref{Stage-Energy}) reaches zero ($0 \geq F_{Z} = F_{sheath} - m_{total}g$). $F_{sheath}$ can be directly calculated from the corresponding principal stress (Eq.~\ref{Stress-Stretch-Uni}) and cross-sectional area $A$:  $F_{sheath} = \sigma_{11}A$.

This is critically different from a regular damped response in that the restorative force doesn't have to go to zero for the system to stop. Because $F_{clutch}$ is not dependent on velocity, the system stops without oscillation so long as \hbox{$ -F_{clutch} < F_{Z} \le 0$} and Eq.~\ref{Stage-Energy} = 0.

\subsubsection{Results}


The experimental stage consistently displaces less than predicted by theory. This is, in part, due to forms of energy dissipation not included in our model such as the viscous energy losses in the silicone and stage bearings. We also fail to take into account elastic energy in the cured Sil-Poxy adhesive. Still, the general trend between force and displacement matches, as seen in Fig.~\ref{Catching_Setup}. 

Oscillations occurred in all three trials for duty cycles less than or equal to 40\%. They were by far the most pronounced for 0\% and 10\%, which oscillated above the starting position (in all three 0\% trials and one 10\% trial). Oscillations also occurred in one trial at 50\% and one trial at 70\%. Theory predicts oscillations ($F_Z > 0$) at normalized clutch forces below \tildehack0.32, which is consistent with a duty cycle between 80\% and 90\% based on our initial characterization.

\subsubsection{Discussion}

There is a perceptible gap between estimated and experimental stage displacement during the drop testing. Specifically, more energy is removed from the system in experiments than was estimated by the model.

Given the discrepancy at the y-intercept, this error is most likely due to the passive response of the system. We attribute this to two main sources. First, the theoretical strain energy of Ecoflex~00-30 is used in this work, with a shear modulus of 17~kPa (Gent model) \citep{steck2019}. Our uniaxial testing for the modeling in Section~\ref{sec: Theory} determines a shear modulus of nearly twice this value at 31.7~kPa for our Ecoflex~00-30. Therefore the sheathing was likely taking substantially more energy from the system per unit strain than our initial model indicates. The second, likely less critical, factor is the energy dissipated into non-axial motion. Video analysis shows shaking of the clutches and stage that is not accounted for in the modeling of stage or clutch friction.

\subsection{SPA Strain Limitation}

\begin{figure}[ht!]
\centerline{\includegraphics[width=\linewidth]{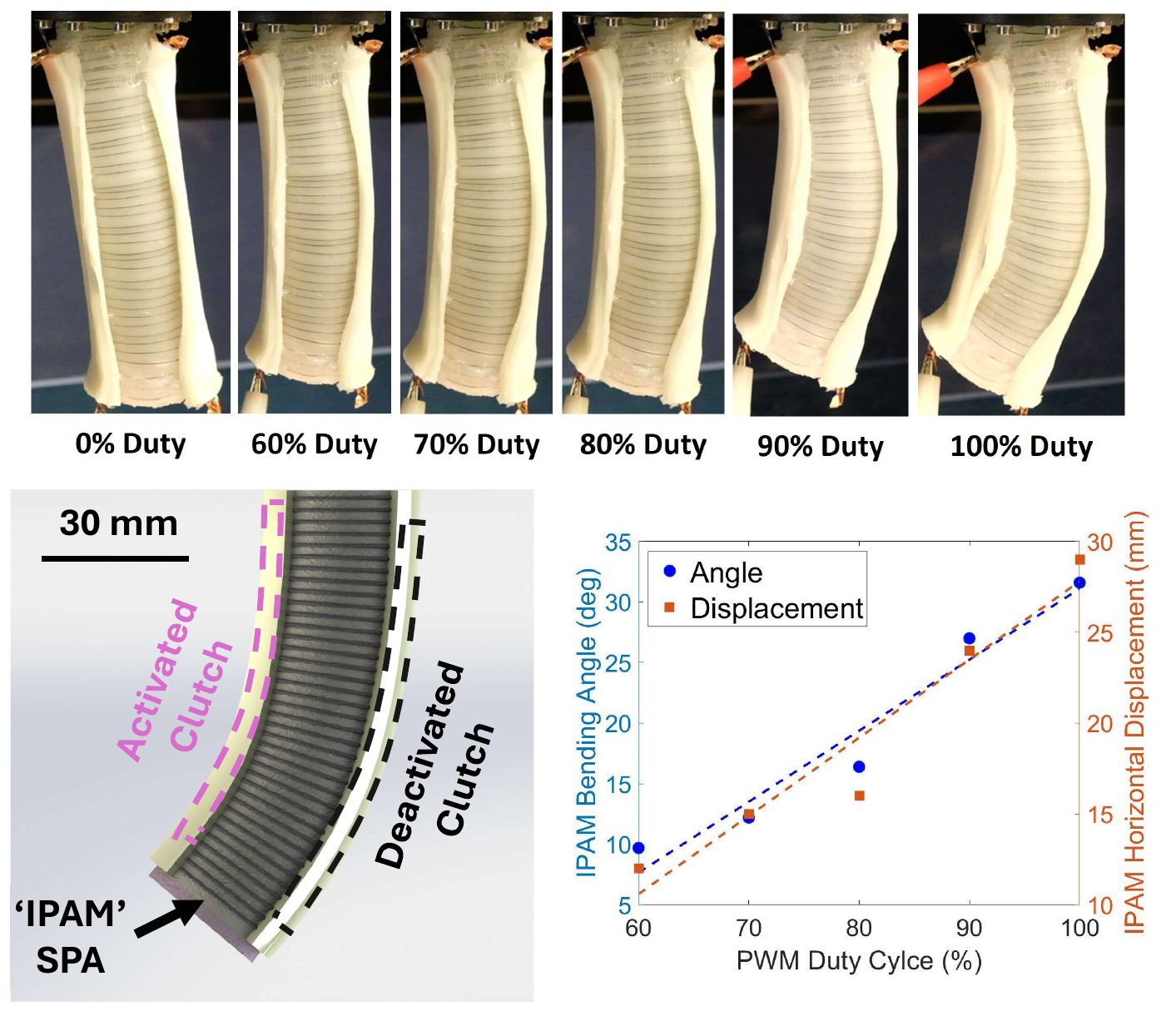}}
\caption{(top) SPA inflated to approximately 25~kPa with different clutch duty cycles. (left) Diagram of SPA expansion with fully activated clutch. (right) PWM effect on bending angle and horizontal displacement.}
\label{IPAM Results}
\end{figure}

\subsubsection{SPA Fabrication \& Experimental Setup}

To test the strain limitation potential of the sheathed clutch, we attached it to a low-pressure, silicone, inverse pneumatic artificial muscle (IPAM). We injection-molded the body of the IPAM with EcoFlex 00-30 in a 3D-printed mold to create a cylinder with wall thicknesses of 1.5~mm. After it cured, we wrapped inextensible fiber (Power Pro 40lb braided line) around the body by rotating the silicone on a drill, similar to  \citep{hawkes2016}. We secured the fiber to the body at both ends with Sil-Poxy and cast an additional, thin, layer of EcoFlex 00-30 on top of the fibers by blade-casting while the system rotates. The system was left spinning during curing to provide even distribution of silicone. Finally, we used Sil-Poxy to adhere a harder silicone (Smooth‑Sil 936) to the flat of the cylinder as a cap.

The sheathing is made of two 1.5~mm thick pieces of Ecoflex 00-30 to minimize off-state antagonistic forces. To adhere the sheathed clutches to the IPAM, we use Sil-Poxy along the entire area of the sheath. We secured the sheath to the IPAM with zip-ties during curing to ensure conformal contact. After curing, we attached a second clutch system to the opposite side in the same manner to balance out the sheath's passive constraint and allow for single-DoF extension when clutches are inactive.

We actuate the IPAM with a 6~V DC pump to an internal gauge pressure of 25~kPa, which we monitor via a Qwiic MicroPressure sensor in a connected pressure chamber. Immediately prior to and throughout actuation, we activate one of the two sheathed clutches to duty cycles corresponding to the characterized values in Fig.~\ref{PWM-Force}. We record the extension with a head-on view. Video frames are used to determine pressure at which the clutch transitions from a static hold to a stick-slip regime, as well as the bending angle and horizontal position at 25~kPa (Digimizer 6.3.0). 

\subsubsection{Results}


Low duty cycles, below 60\%, had a minimal effect on the trajectory of the SPA. Starting at 60\% duty cycle, there was a direct correlation between clutch holding force and the pressure at which it slipped. The clutches began slipping at 
the following pressures for 60-100~\% duty cycles respectively: 5, 6, 7, 10, 15~kPa.

Slipping allows for the clutch system to stretch while still applying an antagonistic force throughout inflation. Once we held internal pressure constant at 25~kPa, slipping no long occurred and there was a unique bending angle and horizontal position for each duty cycle. We will consider bending angle of the IPAM as the clockwise angle from the base plate to the cap. The horizontal displacement of the center of the IPAM cap is measured relative to the position with an inactive clutch. Measurements are taken horizontally in a reference frame fixed at the mount with leftward motion as positive. Zero is considered as the angle or displacement of the IPAM at 25~kPa with the left clutch completely inactive. The right clutch is never activated. Results for angle and displacement are seen in Fig.~\ref{IPAM Results} with linear trend lines. 

The strain-limiting ability of the sheathed clutch is variable in that it slips and then continues to limit strain at a different length. The relation between SPA internal pressure and clutch stress after slip is therefore complex and its exploration is left for future work. 

\section{Conclusions and Future Work}
This work presents design and fabrication techniques for a clutch system that is directly useful for antagonistic forces in soft pneumatic actuation. We calculate off-state constitutive response for sheathing design, and relate active force control to PWM electrical inputs. The system functions in accordance with modeling expectations in the dynamic catching demonstration. We then show how PWM of a clutch can control the effective stiffness of a material to enable position control of the end point of an IPAM actuator.

Future work is required to better characterize PWM control of on-state electrostatic adhesion. We will aim to develop a theoretical understanding of the empirical relations found in \citep{feizi2022} and Fig~\ref{PWM-Force}. Though the shape of the relation is similar to a quadratic that we would expect from a voltage change, there appears to be a lesser holding force associated with the modulation. Further modeling of the effect of this mode of strain limitation on a SPA would also be helpful in determining the best use-cases.

There is significant potential for augmenting SPA trajectory and force outputs with this clutch system. Beyond the shape morphing shown previously \citep{campbell2022}, the use of these antagonistic systems can theoretically increase force output for SPA and enable large-scale object manipulation. Modeling advances for the combined soft system will likely be required to enable these modalities. We also hope to  expand the versatility of the system by adding repeatable external adhesion to the clutch system, similar to \citep{kim2019}. Further work improving the holding performance of EA clutches will also enable higher-pressure SPA applications. 


\chapter{MODELING AND DESIGN FOR AXISYMMETRIC SOFT PNEUMATIC ACTUATORS} \label{ch: Membrane_Design_Section}

Sections 1-7 of this chapter are primarily adapted from:

Gregory M Campbell, Gentian Muhaxheri, Leonardo Ferreira Guilhoto, Christian D Santangelo, Paris Perdikaris, James Pikul, and Mark Yim. Active learning design: Modeling force
output for axisymmetric soft pneumatic actuators. arXiv preprint arXiv:2504.01156, 2025\\

Which is currently under review for Robotics and Automation Letter: Special Issue (RA-L SI) `Interdisciplinarity and Widening Horizons in Soft Robotics'.

My contributions involved leading the design and implementation of the research approach, the design and fabrication of the test platforms and data-collection software, part of the characterization data collection, design and data collection for mass lifting, part of the data post-processing, and the majority of the writing.\\

\section{Introduction}
    
    
    







Soft actuators are promising for physical human-robot interaction in large part due to their compliance. 
Successful control of a soft pneumatic robot requires careful characterization of the soft manipulator, its fluidic elastomer actuators, and the elements that supply fluid energy to predict these reactions \citep{marchese2016}. Characterization of relevant elements for a built system is laborious, and even intractable for soft actuators with many design parameters. This paper  presents soft pneumatic actuator design characterization for actuation trajectories involving applications with external forces.

Researchers have developed a robust understanding, via sophisticated modeling \citep{melly2021}, of hyperelastic silicone materials and their reactions to external forces \citep{feng1973, feng1975}. This understanding has allowed others  to characterize the response of inflated silicone membranes to external forces through analytical solutions \citep{yang2021} and energy methods \citep{herzig2021, shi2023}. Energy methods allow general characterization but require solving sets of ordinary differential equations numerically, which is time-consuming and scales poorly during exploration of a parametric design space. For partially restrained (anisotropic) membranes, some have instead relied on simplified load estimation from contact-area assumptions \citep{ambrose2023}. Contact-based force transforms are valuable for their ease and speed of calculation, but their assumptions break down for larger actuators and strains. It is preferable to combine the strengths of both these methods and fully characterize the actuators in the same manner that broader design spaces have been characterized for shape targeting without loading \citep{pikul2017, forte2022}.

We take inspiration from research that has used Kevlar strain limiters to reinforce and shape the extension of silicone membranes \citep{sholl2021, ceron2018}. Unlike the membranes discussed above, the trajectory of these Fiber-Reinforced Elastomeric Enclosure (FREE) is entirely defined by the inextensible deformation of the fiber elements. The design space for force application with single-expansion FREE’s has been explored generally for slim cylindrical actuators with stiffer rubbers \citep{connolly2017}, including in the presence of external loadings \citep{sedal2021, bishopMoser2015}. These slender actuators can struggle under compressive loading due to buckling \citep{thomalla2022}, while wider-based balloons have been shown to be useful for lifting \citep{sholl2021, campbell2022, ambrose2023, devlin2020} and are less susceptible to catastrophic buckling due to their  low slenderness and expanding cross-sectional area. Softer rubbers also allow the actuators in this work to operate at low pressures relative to the FREE community, with max pressures of 7.5~kPa.


Sequential experimentation and active learning provide a means of  collecting new data to minimize overall error of a machine learning model \citep{ren2021}. Such models are able to provide accurate predictions at very fast speeds at inference time (forward pass on order of ms). Learning has been used for inverse design of shape response for strain-limited membranes \citep{forte2022}, but data-driven methods of characterization have underperformed \textit{energy} methods in the presence of external forces \citep{sedal2021}. We aim to leverage active learning to explore the parameterized design space in a data-efficient way and reduce global model uncertainty. This will allow for the design of actuators targeted at specific lifting applications and reduced model error compared to theoretical, energy-based, methods.

In this work we characterize a parameterized class of actuator to design soft pneumatic actuators with optimized lift response. To understand the response of the inflated system, we solve energy minimization relations that estimate the inflated shape of the actuator in the presence of a known external force. We then use active learning to collect an efficient dataset that spans the prescribed parameter space and that includes strains beyond the linear-elastic region of elastomer deformation. We train our neural network model to interpolate and predict force response between collected data and design parameters. We prescribe a target lift (height and force) trajectory for a single pressure sweep and obtain a membrane design as output. We demonstrate the utility of this model by using it to lift a mass along targeted trajectories as well as to  maximize lift height.


\section{Theoretical actuator modeling}
\label{sec: Theory}

\subsection{Mass-Spring and Finite-Element Modeling}
Membranes can be effectively modeled as many low-mass particles connected by extensible springs with material properties defined by the bulk properties of the material. This is effectively used for co-design of shape structure and control \citep{bhatia2021}. Similar systems with more complex constitutive relations form Finite element analysis (FEA), which has been used successfully for predicting the equilibrium shapes of reinforced membranes undergoing pneumatic inflation \citep{forte2022,campbell2022}.
While comparable FEA could potentially estimate output forces for our pneumatic actuator use-case, each distinct contact area would require an individually-run simulation. While this is achievable, the scaling is poor compared to the theory-based modeling described in the following section.

\begin{figure}[ht!]
\centerline{\includegraphics[width=0.5\textwidth]{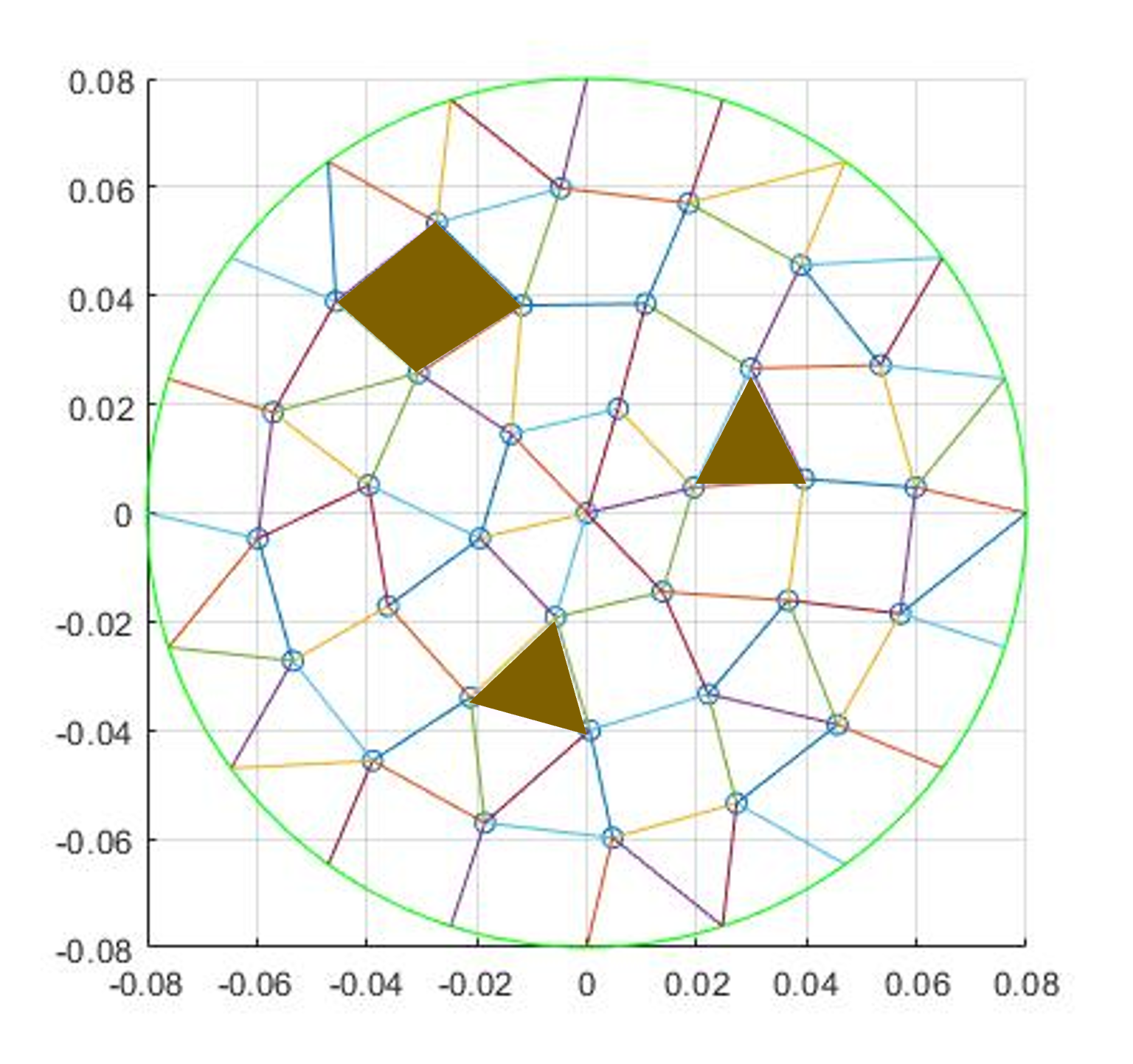}}
\caption{Theoretical layout used in mass-spring methods for inflated membranes. Points represent masses, lines represent springs, and empty regions (such as the shaded regions) represent pressure areas}
\label{FEM}
\end{figure}

In a practical application, we expect to be measuring contact area many times per second (on the order of 30~Hz based off specifications for cameras used in relevant works \citep{yin2022}). There would not be time to run anything approaching an FEA at this refresh rate, but we may be able to solve relevant stretch theory equations quickly enough. And if we develop a look-up table or machine learning equivalent that is solved a priori, we can certainly reference these values fast enough for real-time control.

\subsection{Hyperelastic Silicone Biaxial Expansion Theory}
In our scenario, we can consider the higher-order consitutive properties of silicone as equibiaxial bubble inflation of an in-plane circular silicone membrane (without any embedded stabilization). The initial $\pi*r^2$ surface area will stretch equally both radially and circumferentially as it expands into an oblate spheroid to reach force equilibrium with the pressure pushing from below. This bubble inflation has been studied through the lens of material science, and visual sensing can successfully record stretch values during expansion \citep{yin2022,rosset2014}. Thinking of this membrane instead as an actuator, we are more interested in the relation of pressure (proportional to actuator force) and stretch (contributes to actuator expansion) than in the relation of stress to either. We are also less concerned with whether or not our equibiaxial expansion is 'rigorous', and will assume the entire area expands in this manner.

Though Reuge et al., Rosset et al., and others do an excellent job outlining the theory of this expansion, I will reiterate it here. We assume a diagonal deformation gradient \textbf{F} (only principal stretches), the incompressibility of silicone (constant volume), and path independent (hyperelastic) expansion. The combination of a diagonal deformation gradient and incompressible equibiaxial expansion allows us to define our circumferential and radial stretches as equal ($\lambda_1 = \lambda_2 = \lambda$) and our change in thickness to be $\lambda_3 = 1/\lambda^2$. We then develop the relation of pressure to stretch by equating each pressure and stretch to stress individually.

The relation of pressure and stress for a (hemi)spherical thin-walled vessel is well known, but we can confirm by applying a force balance on the silicone cut in a plane just above and parellel to the aperture of the pressure chamber. The air pressure applies over the interior circular area, while the silicone stress applies over the thickness of the vessel:
\[ P\pi r^2 = 2\pi rt\sigma_{11} \]
where r is the radius of the aperture, $t=\frac{t_0}{\lambda^2}$ is the membrane thickness (where $t_0$ is initial membrane thickness), P is the guage pressure of the air in the pressure chamber, and $\sigma_{11}$ is the first principal stress (radial stress) in the silicone. This then reduces to the following:

\begin{equation}
\label{Stress-Pressure}
 P = \frac{2t}{r}\sigma_{11}
\end{equation}

To relate stretch to the stress, we first turn to the Gent strain energy equation:
\[W = \frac{-\mu I_m}{2} ln(1-\frac{I_1-3}{I_m})\]
where W is the strain energy, $\mu$ is the shear modulus, $I_1 = \lambda_1^2 + \lambda_2^2 + \lambda_3^2 = 2\lambda^2 + 1/\lambda^2 $ is the first invariant (the trace) of the right Cauchy-Green Deformation tensor, and $I_m$ is an empirically solved material property defining the maximum extension of the molecular chains.


The corresponding stress-strain relations are then solved based on the constitutive relation, which holds for incompressible materials that do not rely on invarients $I_2$ or $I_3$:
\[ \mathbf{\sigma} = 2 \frac{\delta W}{\delta I_1} \mathbf{B}-p \mathbf{I}\]
where $\mathbf{B} = \mathbf{F}\mathbf{F}^T$ is the left Cauchy-Green tensor, \textbf{$\sigma$} is the Cauchy stress tensor, p is an undefined hydrostatic pressure responsible for enforcing incompressibility, and \textbf{I} is the identity matrix.

Setting $\sigma_{33} = 0$ due to the lack of restriction in the direction of membrane thickness allows us to solve:
\[ p = \frac{\mu I_m}{\lambda^4 (I_m-I_1+3)}\]
which then gives us our final relation between $\sigma_{11}$ and $\lambda$:
\begin{equation}
\label{Stress-Stretch}
    \sigma_{11} = \frac{\mu I_m}{I_m-I_1+3} (\lambda^2- 1/\lambda^4)   
\end{equation}

Combining Eqn. \ref{Stress-Pressure} and Eqn. \ref{Stress-Stretch} gives us our desired relation of pressure as a function of stretch for equibiaxial expansion:
\begin{equation}
\label{Pressure-Stretch}
    P = \frac{2t_0 \mu I_m}{r(I_m-I_1+3)} (1 - \frac{1}{\lambda^6})   
\end{equation}

Notably, this equation now allows us to relate pressure to stretch via the two empirically solved material values required for Gent's strain energy model and two design variables: membrane thickness and aperture radius.

\begin{figure}[ht!]
      \centering
      \includegraphics[width=.87\linewidth]{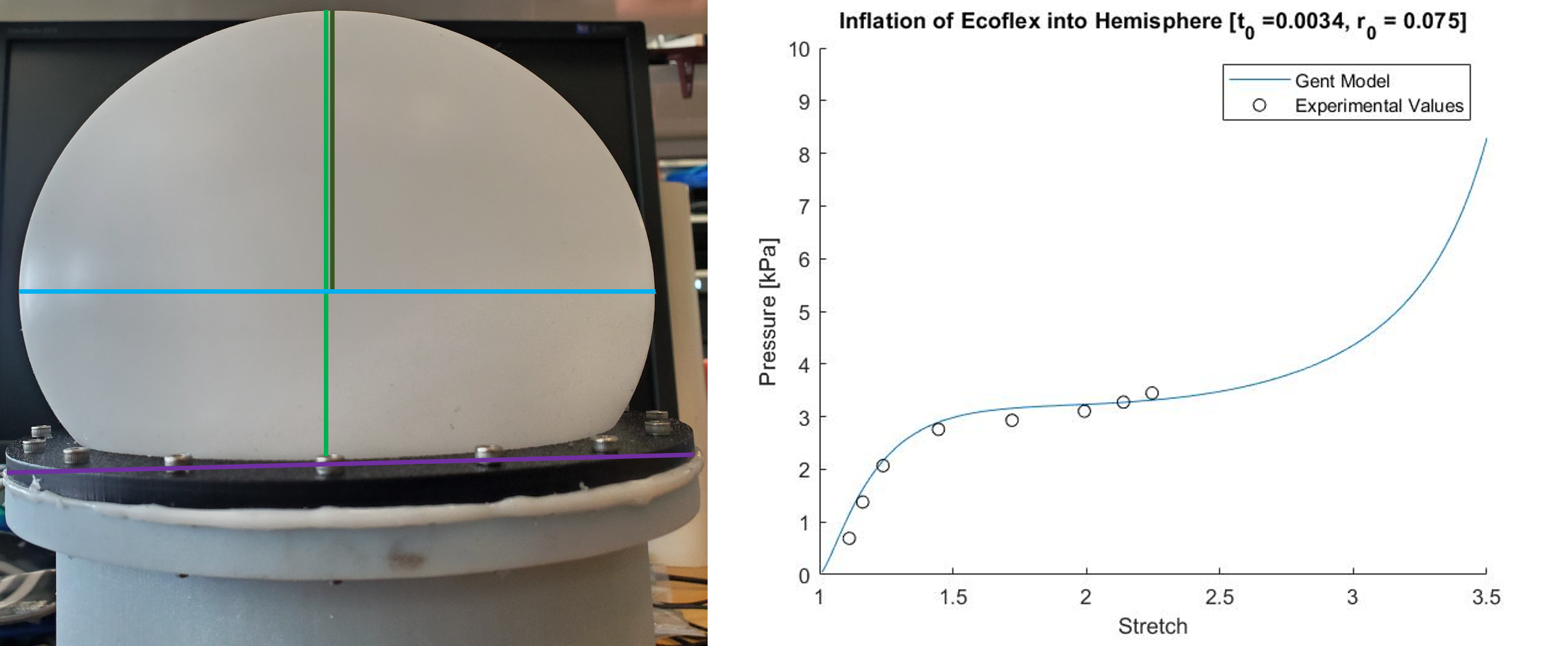}
      \caption{left: measurements used to solve for surface area of inflated membrane for stretch calculations, right: plot of pressure-stretch relation described in eqn. \ref{Pressure-Stretch} compared to experimental data from inflation of an unrestricted silicone membrane}
      \label{Silicone Verification}
\end{figure}

In reality, r increases when the bubble reaches stretches over $\sqrt{2}$, which lead to larger raddii at the equator of the oblate spheroid than the aperture. We can estimate the radius of the membrane as a function of stretch by solving for the surface area of the spheroid that best matches the area of silicone available ($\lambda^2*SA_0$). Using this geometric reasoning and published material values for Ecoflex 00-30 \citep{steck2019}, we are able to experimentally verify this theory as seen in Fig. \ref{Silicone Verification}.

\subsection{Contact and Energy methods}
Generally speaking, as an external object comes into contact with the surface of an elastomeric membrane, it restricts local expansion. This surface restriction, alone, is enough to disrupt the geometric assumptions made in the previous section. However these type of restrictions have been examined both theoretically \citep{Chou2005} and practically in work with FREE actuators \citep{sedal2021, sholl2021}. For our simple case, we instead consider this as a boundary condition that limits expansion only in the affected section of membrane. This also allows us to consider an external pressure applied at that location, which can correspond to the lifting of a mass.

The combined result of the stretch restriction and external pressure is a shape output that differs greatly from our geometric assumptions required for eqn. \ref{Pressure-Stretch}. This shape is solved using an energy-minimization calculation as follows.

\subsection{Elastomeric Thin Membrane}

We wish to determine the shape of a thin membrane upon inflation, with an external force applied  at membrane radius, $r$, ranging from $0\leq r \leq r_0$ as shown schematically in Fig.~\ref{fig:Membrane schematic}. 
\begin{figure} [htb]
    \centering
    \includegraphics[width= 0.5\linewidth]{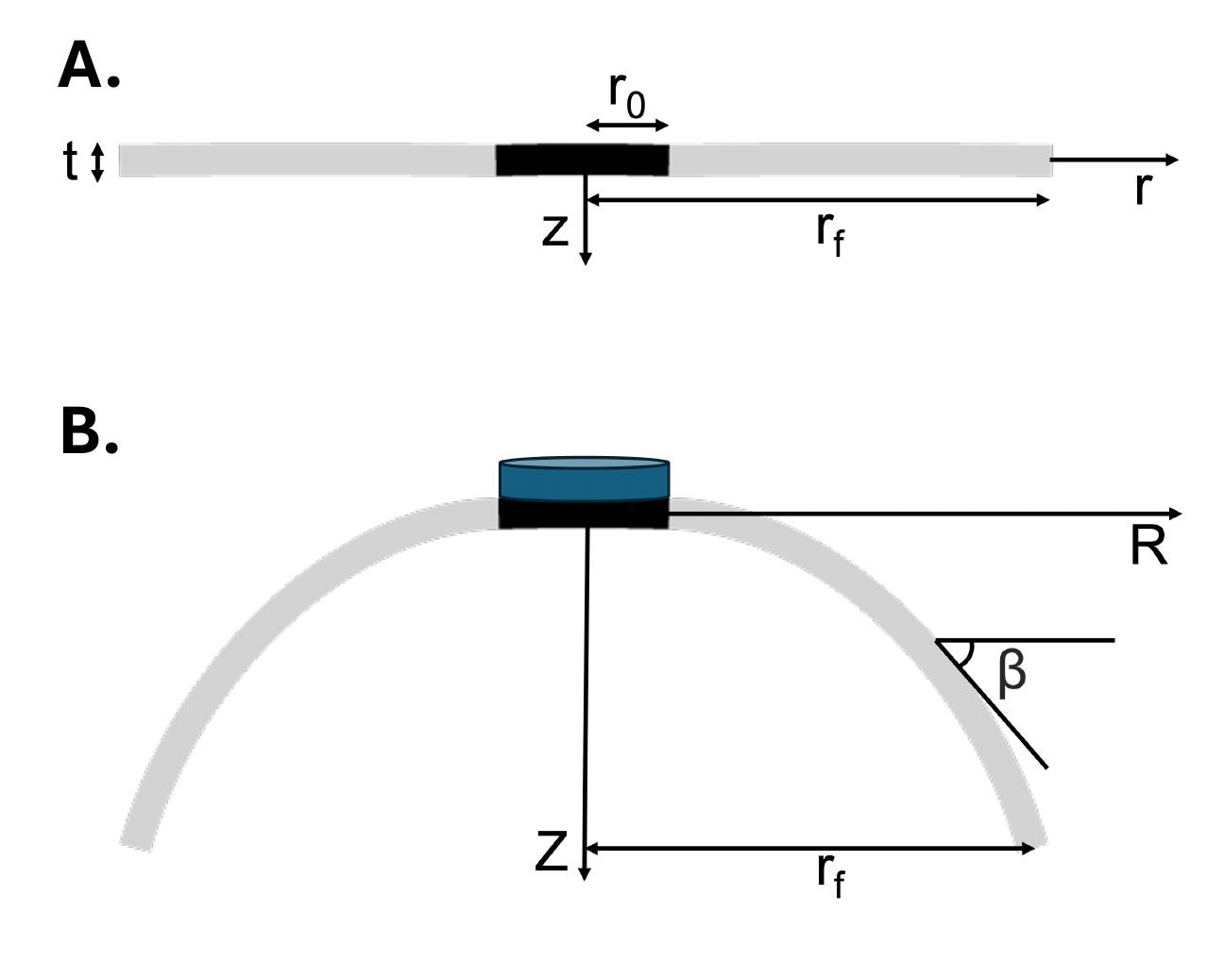}
    \caption{Schematic of the membrane (gray) with strain limiter (black) in the A. undeformed state, and B. deformed state while in contact with external force (blue).}
    \label{fig:Membrane schematic}
\end{figure}Before deformation, the shape of the membrane that is not in contact with the force, using cylindrical coordinates, is given by $r_0\leq r \leq r_f$, and membrane height $z=0$. The region $0\leq r \leq r_0,$ which is in contact with the applied force, consists of low-strain material and we assume that there is no deformation in that region, only a translation of the region along the $z$ direction. We  assume  the membrane stays axisymmetric after the deformation and its shape is given by,
\begin{equation}
    \label{eq:membrane parametrization}
    R=R(r),\:\:\: Z=Z(r).
\end{equation}
From this, we can write the deformation gradient as, 
\begin{equation}
    \mathbf{F} = R'(r)\mathbf{\hat{r}} \otimes \mathbf{\hat{r}} + (R(r)/r) \mathbf{\hat{\theta}} \otimes \mathbf{\hat{\theta}} + Z'(r) \mathbf{\hat{z}} \otimes \mathbf{\hat{r}},
\end{equation}
where primed variables indicate a derivative with respect to $r.$ The eigenvalues of the right Cauchy-Green tensor, namely of $\mathbf{C}=\mathbf{F^T F},$ are the principal stretches of the membrane,
\begin{equation}
    \label{eq:stretches}
    \lambda_1(r)=\sqrt{R'(r)^2+Z'(r)^2},\:\:\:\lambda_2(r)=\frac{R(r)}{r},
\end{equation}
where $\lambda_1(r)$ represents the amount of material stretch along the meridian direction, while $\lambda_2(r)$ represents the amount of material stretch along the latitude direction. The third principal stretch comes from our assumption of incompressibility, $\lambda_1(r) \lambda_2(r)\lambda_3(r)=1 \xrightarrow{}\lambda_3(r)=1/(\lambda_1(r) \lambda_2(r)),$ where $\lambda_3(r)$ represents the material stretch in the direction normal to the membrane (thickness direction). We write an energy for the system that assumes full contact between the contact plate and the strain-limiting contact region as shown in Fig.~\ref{fig:Graphical_Abstract}B. 
\begin{equation}
\begin{aligned}
    \label{eq: energy}
    E=\int_{r_0}^{r_f}\left(\underbrace{2\pi\: r \:t \:W(\lambda_1,\lambda_2)dr}_{\text{elastic energy }} - \underbrace{\pi\:p R^2\:Z' dr }_{\substack{\text{work done by} \\ \text{pressure difference}}}\right)~+ \\ \int_0^{r_0} \left( \underbrace{-F Z' dr}_{\text{external force work}}+2\pi\: r \:t \:\tilde{W}(\lambda_1,\lambda_2)dr - \pi\:p R^2\:Z' dr\right),
\end{aligned}
\end{equation}
\noindent where $W$ and $\tilde{W}$ are the strain energy per unit undeformed volume for the elastic material and the low-strain material respectively, while $p$ is the gauge pressure inflating the membrane. In this study, we use the Gent model for the strain energy density function \citep{zhou2018}, which we can write as $W = -\frac{\mu J_m}{2} \ln(1 - \frac{\lambda_1(r)^2 + \lambda_2(r)^2 + \lambda_3(r)^2 - 3}{J_m})$. This model contains two constants, the shear modulus $\mu$ and the extension limit constant $J_m$. We find the elastic material constants using uniaxial testing, similar to \citep{Marechal2020}: $\mu=31.7~kPa,$ $J_m= 39.6$. We estimate sufficiently stiff values for the strain-limiting material based on data in \citep{pikul2017}: $\tilde{\mu}=1~MPa,$ and $\tilde{J_m} = 25.$ 

We first set the first variation of the energy to zero,
\begin{align}
\delta E &= \int_{r_0}^{r_f} D \left(
    \frac{\partial W}{\partial \lambda_1} (\frac{\partial \lambda_1}{\partial R'} \delta R'
    + \frac{\partial \lambda_1}{\partial Z'} \delta Z')
    + \frac{\partial W}{\partial \lambda_2} \frac{\partial \lambda_2}{\partial R} \delta R
\right) dr \nonumber \\
&\quad - \int_{r_0}^{r_f} \pi p \left(2 R Z' \delta R + R^2 \delta Z' \right) dr
- \int_0^{r_0} F \delta Z' = 0.
\label{eq: first variation}
\end{align}

Where $D=2 \pi r t$. We then integrate, by parts where necessary, to obtain the equations of equilibrium and apply geometric relations based on $\beta(r),$ which we take to be the angle between the tangent to the membrane and the horizontal line at point $r,$ as shown in Fig. \ref{fig:Membrane schematic}B, \hbox{$R'(r)=\lambda_1(r) \text{cos}(\beta(r)),$} $\:\:\:Z'(r)=\lambda_1(r) \text{sin}(\beta(r))$. After simplification, we write the equilibrium equations for the $r_0\leq r \leq r_f$ region as:
\begin{align}
&\frac{d\lambda_1}{dr}=\frac{W_2-\lambda_1 W_{12}}{r W_{11}}\text{cos}\beta+\frac{\lambda_2 W_{12}-W_1}{r W_{11}},\label{eq: equilibrium equations1}\\ &\frac{d\lambda_2}{dr}=\frac{\lambda_1 \text{cos}\beta-\lambda_2}{r},\label{eq: equilibrium equations2}\\
&\frac{d\beta}{dr}=\frac{\tilde{p}r \lambda_1 \lambda_2-W_2 \text{sin}\beta}{r W_1},\label{eq: equilibrium equations3}
\end{align}
where $\tilde{p}=p/t, \:W_1=\partial W/\partial \lambda_1, \:W_2=\:W/\partial \lambda_2,$ and \hbox{$W_{12}=\partial^2 W/\partial \lambda_1 \lambda_2.$}

The second term  in Eq. \ref{eq: energy} only contributes to the boundary conditions at $r_0$ since there is no deformation in the $0\leq r < r_0$ region. The boundary conditions that come from the integration by parts give us a condition on the angle $\beta,$ at $r=r_0,$ so the initial boundary conditions needed to solve Eqns. \eqref{eq: equilibrium equations1}, \eqref{eq: equilibrium equations2}, and \eqref{eq: equilibrium equations3} are given by, 
\begin{align}
    \label{eq:initial conditions}
    &\lambda_1(r_0)=x; 
    \lambda_2(r_0)=1;
\beta(r_0)=\text{ArcSin}\left(\frac{\pi p\:r_0^2\lambda_2^2-F}{2\pi t\:r_0 W_1}\right)\Big|_{r=r_0}
\end{align} 
The condition on $\lambda_2$ comes from the fact that the contact area and the plate are in full contact and there is no extension in that area. The condition on $\beta$ comes from the boundary conditions obtained during energy minimization. We provide values for $p$ and $F,$ and solve the condition on $\lambda_1$ by using the shooting method to find the value of $x$ that satisfies the condition $\lambda_2(r_f)=1.$ This final condition states that the membrane is fixed at the ends. Integrating the equilibrium equations from $r_0$ to $r_f$ obtains the shape of the deformed membrane. Thus, we find the height of the membrane for a given pressure and force value. Solving this relationship across many pressure-force pairings creates a theoretical model that can estimate forces for pressure-height pairings within, or nearby, the set.

\subsection{Concentrically strain-limited Thin Membrane}
To span our actuator design space, we want to find the deformed shape of a membrane made up of elastic material that also includes strain-limiting rings (see Fig.~\ref{fig:Graphical_Abstract}A). To do this, the energy in Eq. \eqref{eq: energy}\: will involve an extra integral for each additional piece we add to the system, with the strain energy density function $W$ being used for the elastic material and $\tilde{W}$ being used for the strain-limiting rings. We have to solve Eqns. \eqref{eq: equilibrium equations1}, \eqref{eq: equilibrium equations2}, and \eqref{eq: equilibrium equations3} for each of the material pieces separately, with the initial boundary conditions at each piece related through the boundary condition at the point of contact as well as the final boundary condition $\lambda_2(r_f)=1.$ 

Solving this final boundary condition is unsuccessful, however, for 25\% of membranes because of solver divergence brought on due to numerical stiffness for the heterogeneous material membranes. The successful subset of solutions are used and the results are discussed in Sec. \ref{subsec: ringed model performance}.

\section{Design space and experimental modeling}

\subsection{Role of Experiments}


The underlying mechanics of a soft pneumatic actuator can be considered as a full pneumatic system, a macroscale material structure, or even a microscale material structure. The theoretical discussion above applies physics at the macroscale, and this affects the behaviors we ultimately aim to design for at the systematic level. However for systems with complex strain limitation, we see a gap between macroscale analytical, and even FEA, predictions and experimental output \citep{yang2021}. We see this explicitly in Ch.~\ref{ch: Shape Morphing}, where FEA results differ from experimental by a RMSE ranging from 7~\% of maximum height in the least constrained case to 13~\% of maximum height in the most constrained case. This is exacerbated in the presence of external loading, with Sedal et al. reporting mean errors over 17\% of their maximum force even for their improved continuum model \citep{sedal2021}. This gap is likely due to the more microscale structures, for instance air gaps or how individual polymer chains interact with stiffening elements. In part due to these microscale interactions, experimental characterization and tuning are still required even for individual designs \citep{marchese2016}. The goal in this work is to eliminate the subsequent effort of accounting for these microscale phenomena theoretically by using experimental data that naturally incorporates microscale phenomena. We specify simplified state vectors that align with our output goals and which we can precisely measure during experiments.

Experimental data also represents an important real-world benchmark for our theoretical and simulated results. While FEA has been used as ground-truth in past works \citep{yang2021a}, these results are subject to user-chosen mechanical properties and model parameters. Well-documented experimental results can provide a more appropriate baseline to which we can compare future analytical or simulation results, even when experiments still differ from how actuators may function outside the lab (see Section~\ref{sec: Peturbations}). In this work, we evaluate both our theoretical and learning-based models with a k-fold cross-validation using the collected experimental data.

\begin{figure*} [!t]
    \centering
    \includegraphics[width=\textwidth]{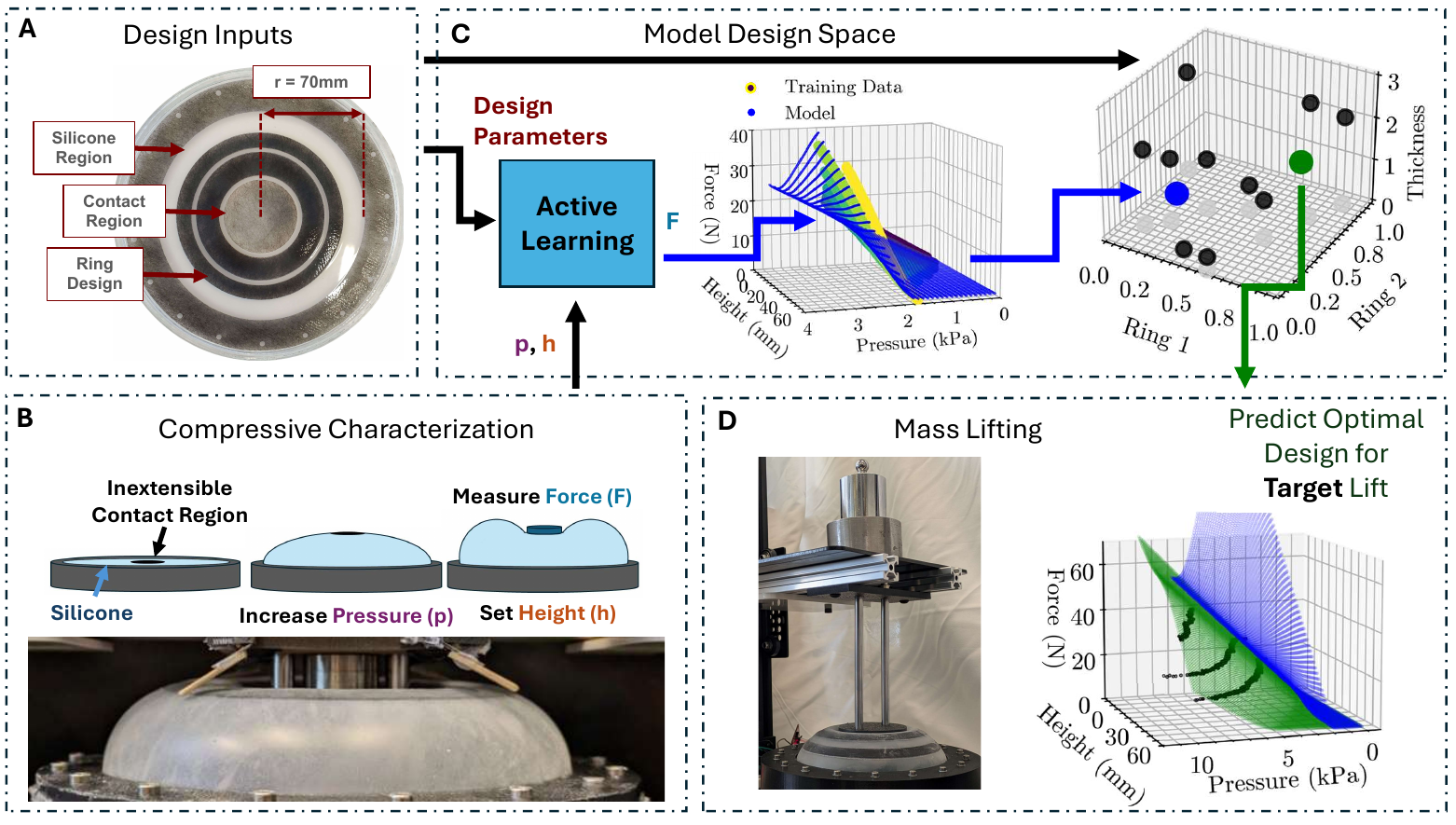}
    \caption{A. Design parameters for example membrane. B. Compressive Testing (top) Testing procedure for each membrane: expansion from a flat plane into a set height load-cell. Procedure is repeated for different heights. (bottom) Example of physical test measuring force for varying pressure at a set height. C. (left) Pressure, height, and six design parameters predict force output for a given membrane. (middle) Model (blue) overlaid with force-pressure training data at varying heights (0-70mm from purple to yellow). (right) 3-D visualization of 6-D design space. Example training set points in black, model from C in blue, model from D in green. D. Optimization for lifting task. (left) Physical testing to verify lift trajectories at a given mass. (right) Model planes for (blue) training parameters and (green) design parameters optimized to hit target trajectories (trajectories in black).}
    \label{fig:Graphical_Abstract}
\end{figure*}

\subsection{Design Space}

While soft pneumatic actuators can take a very large number of configurations, this work focuses on a class of soft pneumatic actuator defined as a thin (between 1 and 3~mm), circular membrane of radius 70~mm made from EcoFlex 00-30 rubber and reinforced with up to two axisymmetric rings (Soft n' Shear fabric). We ensure at least 10~mm between the outer and inner radius of a ring and contact areas between 25.4 and 38.1~mm. Our design space is an axisymmetric, finite, subset of all membrane-based SPA. 
An example membrane is seen in Figure \ref{fig:Graphical_Abstract}A-B. 

\subsection{Data Collection}
We fabricate membranes using gravity molding of Ecoflex 00-30. Lasercut strain-limiting rings and contact regions (Soft n' Shear) are applied  to the uncured silicone after degassing and before curing. Membranes are mounted on a 3D-printed pressure chamber, which houses an air pressure sensor (MPRLS0025) and an ESP32 microcontroller to wirelessly transmit pressure data. Air pressure is supplied by a  \hbox{4.5~V} DC air pump (ZR370-02PM) and released by a 12~V solenoid valve (Plum Garden). A piezoelectric load cell measures forces via an acrylic contact plate the same radius as the membrane's contact region and load is limited to move vertically  with linear ball bearings.

We perform automated testing for each membrane where the contact plate is positioned vertically by linear actuators (Homend) and verified by a time-of-flight sensor (VL53L0X) between trials. A single trial consists of the activation of the pump and the subsequent inflation of the actuator. Inflation continues for ringed membranes until the internal gauge pressure reaches 4.3~kPa. Then the pump is deactivated and the solenoid valve releases air until internal pressure reaches atmospheric. If the membrane comes into contact with anything except the contact plate, the trial is completed. Three trials are performed per contact plate height, with eight heights, 0-70~mm, per membrane. 
If the membrane does not burst, testing is repeated to a maximum pressure of 6.1~kPa. Ringless membranes were given no pressure limit and allowed to inflate until contact with the test rig.

Data from each trial includes time, force, pressure, time of flight height data (left height, right height), flow rate, and contact with test rig (binary), material type, nominal thickness, radius, contact radius, test year, test month, test day, curing rack (A/B), contact plate nominal height, trial number, thicknesses (from destructive testing of some membranes), and ring data. While deflation data is recorded in some cases, exclusively inflation data is used in model training and verification. 
The data is also sorted into a learning-friendly format as a dictionary with key values matching membrane design parameters. 
Dictionary data in the form of a .pkl file can be found in the Github repository (see Data Availability section).

\subsection{Model Architecture}

\begin{figure}[!htb]
    \centering
    \includegraphics[width=0.95\linewidth]{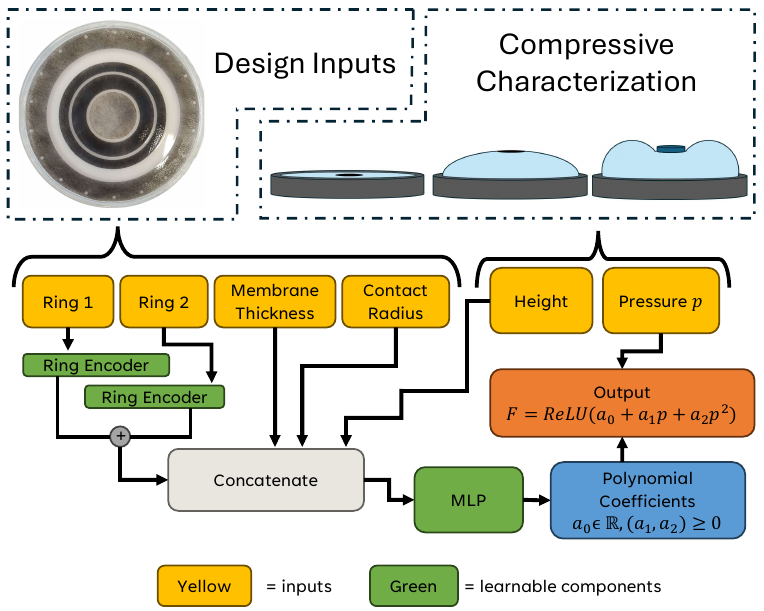}
    \caption{Model Architecture. (Top) Characterization data, pressure and height, and design inputs, ring radius and width, membrane thickness, and contact radius, are inputs to the actuator model. (Bottom) The  model solves for  force output 
    relative to pressure. 
    }
    \label{fig:model-architecture}
\end{figure}


Our neural network model predicts the external force $F\in \mathbb{R}$ generated by pneumatic membrane actuators under varying conditions. The model accepts three primary inputs: (1) a membrane design vector $M\in \mathbb{R}^6$ encoding six design parameters, (2) object displacement height $h\in\mathbb{R}$ from the actuator's initial plane, and (3) internal gauge pressure $p\in\mathbb{R}$. These inputs fully characterize the actuation task, allowing the model to interpolate between collected pressure-height-force data for tested membranes (Figure \ref{fig:Graphical_Abstract} D) and predict force responses for untested membrane designs.

The membrane design vector $M$ encodes the following parameters: contact radius, membrane thickness, and four ring-related parameters (position and width for each of two potential rings, with NaN values indicating ring absence). Example vector formatting is shown in Table \ref{tab: Mems}.

Our approach follows the Operator Learning framework \citep{li2020, lu2021}, where neural networks operate on functions in potentially infinite-dimensional spaces. Specifically, given membrane design $M$ and height $h$, our model outputs a function $F_{M,h}:\mathbb{R}\to\mathbb{R}$ that computes external force for any given pressure: $F_{M,h}(p)$ represents the force applied by membrane $M$ at height $h$ under pressure $p$.

\textbf{Ring Encoder Design.}
To handle both ringed and ringless membrane designs within a unified architecture, we developed a specialized Ring Encoder that converts ring parameters (or their absence) into consistent latent representations. The encoder operates as follows:
\begin{itemize}
    \item For valid ring parameters represented as $v\in\mathbb{R}^2$ (position and width), the encoder applies a linear transformation: $e = Tv$, where $T\in\mathbb{R}^{d\times 2}$ is a trainable weight matrix projecting to dimension $d$.
    \item For missing rings (indicated by NaN), the encoder maps directly to a fixed but trainable vector $e_{NaN}\in\mathbb{R}^d$.
\end{itemize}

To ensure injectivity and prevent information loss, we set $d\geq3$, leading $e_{NaN}$ to lie outside the image of matrix $T$. We explored dimensions $d \in \{3, 12, 24\}$ during ablations, with a single shared architecture across all ring encoders to maintain consistent representations. The individual ring representations are summed, making the model invariant to ring ordering (Ring 1 vs. Ring 2) and adaptable for additional rings. The latent representation is passed into an initial MLP (we explored [layer, neuron] shapes of [8,8] and [16,5] during ablations) or passed directly into the base MLP.

\textbf{Force-Pressure Relationship Modeling.}
The encoded ring information is concatenated with membrane thickness, contact radius, and height, then fed into a Multi-Layer Perceptron (MLP). This network computes polynomial coefficients that characterize the system's force-pressure response. Individual data points are weighted during training such that each membrane test, not each data-point, contributes evenly.

We impose several physical constraints:
\begin{itemize}
    \item \textit{Monotonicity}: Coefficients multiplying pressure terms ($a_1, a_2$ for quadratic polynomials) are constrained to be non-negative (more pressure does not lead to less force).
    \item \textit{Physical realism}: The constant term $a_0$ may be negative, but a ReLU activation is applied to the final polynomial output, preventing negative force predictions and ensuring consistency with experimental data. That is, if at a given pressure the membrane is unable to lift the object, the model outputs 0 instead of a negative value.
\end{itemize}

During ablation studies, we evaluated several function representations:
\begin{itemize}
    \item Learnable basis functions (DeepONet strategy \citep{lu2021}).
    \item Explicit polynomial assumptions (of degree 1,2,3).
\end{itemize}
The constrained polynomial representations (i.e., degree 2, shown in Fig.~\ref{fig:model-architecture}) proved superior, as the DeepONet formulation exhibited relative overfitting.

\textbf{Training Details}
All networks were trained for 10,000 iterations using the Adam optimizer \citep{kingma2017} on an Nvidia RTX A6000 GPU, requiring approximately 20 seconds per network during ablation experiments.

\subsection{Active Learning}


In order to carry out Active Learning (AL) \citep{ren2021}, we must first be able to quantify the epistemic (model) uncertainty of our predictions under the operator learning framework \citep{guilhoto2024}. We do this by using a Randomized Prior Network (RPN) ensemble of independent networks \citep{osband2018, yang2022}. Each of the $N\in\mathbb{N}$ members of the ensemble combines a prior and a trainable network to output a different prediction $F_{M,h}^1(p),F_{M,h}^2(p),\dots, F_{M,h}^N(p)$, which are averaged in order to compute the final prediction $F_{M,h}(p)=\frac{1}{N}\sum_{i=1}^NF_{M,h}^i(p)$. The epistemic uncertainty is computed as the standard deviation of these predictions, with a small value indicating agreement between members of the ensemble (low epistemic uncertainty), and a large value indicating disagreement (high epistemic uncertainty). The active learning procedure is then carried out by selecting the membrane $M\in\mathbb{R}^6$ for which predictions on average have the highest possible value of epistemic uncertainty. Testing such a membrane and training the model with this newly acquired data then decreases the uncertainty for this membrane and other designs similar to it, since the model is now trained with the experiments from this specific design. We carry out this procedure iteratively, until enough membranes are acquired.

In particular, we carry out active learning in the parallel setting, where at each iteration $q=2$ membranes are obtained simultaneously. We determine the best pair of membranes to collect by selecting the designs that present largest uncertainty in predictions at selected heights $h_1,\dots, h_{N_h}$ and pressures $p_1,\dots,p_{N_p}$. This is quantified via the acquisition function
\begin{equation}\label{eq: al-acquisition}
    \alpha(M_1, M_2) = \sum_{l=1}^N\max_{k\in\{1,2\}} \left[ \sqrt{\sum_i \sum_j \left(F_{M_k,h_i}(p_j)-F_{M_k,h_i}^l(p_j)\right)^2} \right]
\end{equation}
which is maximized using the L-BFGS optimizer with several starting points, then taking the best optimized value overall. Note that the expression in Eqn. (\ref{eq: al-acquisition}) is easily extendable to settings where we wish to collect an arbitrary number $q\in\mathbb{N}$ of membranes at each iteration, instead of only two.

\subsection{Uncertainty}
\label{sec: uncertainty}
Our acquisition function aims to reduce epistemic (model) uncertainty across the design space, but little consideration is given to the aleatoric (data) uncertainty of our characterization in this algorithmic approach. We perform three characterization trials for each set height on each membrane characterization and use only the third trial in model training and testing. It is necessary to discount the first trial because of plastic deformation (the Mollins effect \citep{diani2009}). We choose to also ignore the second trial to reduce the effect of aleatoric uncertainty on our error metrics. We also use only inflation data, as opposed to the force-pressure relations during both inflation and deflation. We see hysteresis up to about 2~N between inflation and deflation that may further increase error outside of characterization conditions.

Aleatoric uncertainty still enters our data, especially in the form of fabrication imprecision (membrane thickness, ring positioning, silicone mixing ratio), and set height imprecision (measured within 1~mm). Presumably the use of multiple trials for each height would only increase the associated aleatoric error further. We attempt to eliminate the worst of the test imprecision by trimming our dataset of obviously flawed tests, but with such a small number of tests, any associated error will affect the precision of our model. Further aleatoric uncertainty will likely emerge as conditions grow in complexity, for instance in non-axisymmetric, multi-DoF lifts.

We expect there is still an opportunity to decrease epistemic uncertainty further and in turn reduce NN model error, at the cost of additional testing. We stopped characterization testing once our acquisition function started recommending 1-ring membrane designs, indicating that the 2-ring design space may be well characterized. The magnitude of error for NN model ringless membrane predictions (without considering the ringed dataset) relative to the baselines indicates that this region still suffers from epistemic uncertainty. Given the accuracy of theoretical modeling in the ringless domain, this could be an excellent opportunity to use multi-fidelity data streams to further reduce epistemic uncertainty.

\section{Results: Characterization \& Modeling}

\begin{figure} [!ht]
    \centering
    \includegraphics[width=\linewidth]{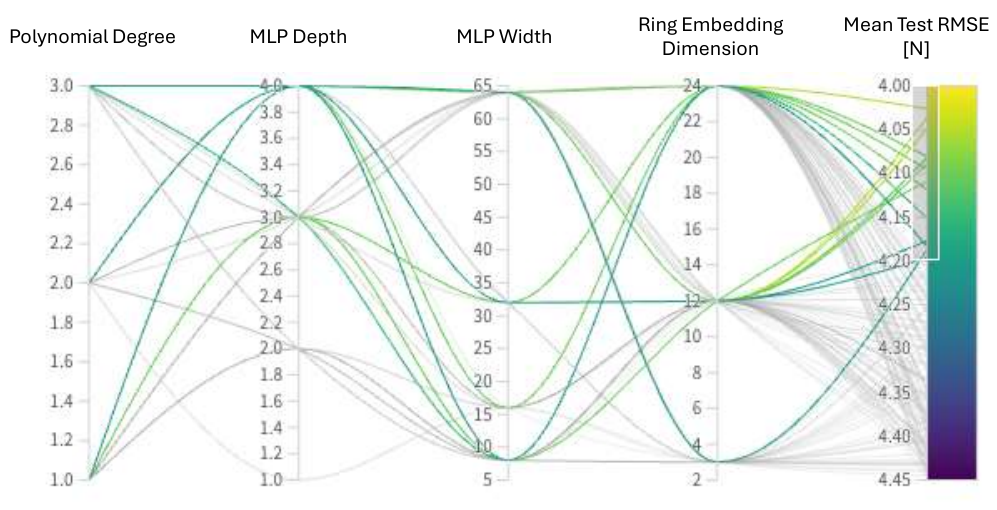}
    \caption{Model RMSE [N] of best performing model hyperparameters tracked to the corresponding parameters: output force polynomial degree relative to pressure, multi-layer perceptron depth and width, and ring encoder MLP width.}
    \label{fig: hyper-sweep}
\end{figure}

\subsection{Actuator Performance}
The membranes central to these actuators use shore hardness 00-30 silicone rubber. This combined with the 70~mm actuator radius allows them to operate at pressures under 7.5~kPa and apply forces reaching 103~N. Maximum force occurred with a ringless membrane of thickness 2~mm and contact radius 38.1~mm at a height of 50~mm. We characterized each actuator's displacement up to 70~mm when possible, and heights up to 79~mm were reached during mass lifting. Without taking into account the membrane properties, the maximum contact area tested, $45.6~cm^2$, would only be expected ($F=p*A$) to provide approximately 34~N of force. 

We fabricated 28 different membranes, of which 22 were ultimately used in model training. Other membrane data was discarded primarily due to stiffening element delamination or development of a hole in the membrane. These modes of failure occurred exclusively at boundaries between unstiffened and stiffened regions of silicone. Fracture due to expansion past the rubber's strain limitation did not disqualify a membrane, though we removed the trial involving the fracture from that membrane's dataset. The resulting trimmed dataset included 188,318 individual data points. Individual membrane designs each had between 1,906 and 17,044 data points (median: 7,116, inter-quartile range: 4,583 to 12,780). 


\subsection{Model Performance (Ringless)} \label{subsec: ringless model performance}
We compare the performance of the MLP-based pipeline to the theoretical (energy-minimization) model and a naive baseline regression. To create the second-degree polynomial baseline regression for ringed membranes, we concatenate all monomials up to order two for ring parameters, membrane thickness, contact radius, heights, and pressures. We fit a linear regression of this concatenation to the force outputs. We repeat these steps for a ringless membrane model, this time ignoring ring parameters in the concatenation.



We perform a $k=3$ $k$-fold validation with pressure, height, and membrane design parameters as inputs and force as an output across the six ringless membranes. That is, for each of the $k=3$ folds, we pick 2 membranes as our test set and use the remaining 4 for training, cycling through the 3 different groups and taking the average test error across the 3 folds. The methods predicted force with the following root mean square error (RMSE): Neural Network (NN): 6.4~N, Theoretical model: 5.9~N, Regression baseline: 8.0~N. With the addition of the ringed data, the NN RMSE dropped to 5.1~N. 
The maximum force seen in the lift trials was 103~N, of which these RMSE represent between 5.0\% (5.1~N) and 7.8\% (8.0~N).

\subsection{Model Performance (Ringed)} \label{subsec: ringed model performance}
We optimize model hyper parameters for the entire dataset by sweeping through: form of $F_{M,h}:\mathbb{R}\to\mathbb{R}$, width/depth of network, and latent ring representation. We compare the RMSE in model force relative to relevant test-set characterization data in a $k=11$ $k$-fold cross-validation across 22 membranes. 

The best-performing hyper-parameters result in a RMSE of 4.0~N (4\% of max seen force) across ringed and ringless membranes. 
The model generally performs best with $F_{M,h}$ as polynomial degree 1, MLP depth of 3 or 4, and ring embedding model dimension of 12 or 24. MLP width doesn't have a large impact on result. Neither does the existence or parameters of a separate MLP for ring pre-processing.

It is difficult to baseline these values, as both the theoretical model and regression approach used for ringless trials fail to effectively deal with the parameterized ring values in all cases. Solving the theoretical model was successful for 75~\% of ringed membranes and the average RMSE across both ringed and ringless membranes was 4.4~N. The polynomial regression needed to be optimized separately with different numbers of input variables for ringed and ringless membranes, and the average RMSE across both sets was 9.4~N.

\section{Design for Open-Loop Applications}


Once the NN model is trained, we use it to predict actuator performance in lifting tasks. We define a lift trajectory by the change in height of the object being lifted, the force applied by the actuator, and the pressure inflating the actuator. Our model therefore predicts the effect of design parameter choices on lift trajectories. Experimental lifts are performed with a 1
degree-of-freedom (DoF) test stage that is constrained by gantry plates. Masses are placed on the gantry and force is transmitted to the actuator via a contact plate (see Fig. \ref{fig:Graphical_Abstract}.D). 

We first verify model predictions on two membranes (one with and one without rings) from the training set. We model the expected lift trajectories for three target masses, 1.5~kg, 2.5~kg, and 4~kg, and choose three points along each trajectory as target way-points. The points are chosen by the associated heights: 5, 40, and 50~mm. These heights were chosen because they represent areas of the trajectory that vary greatly between designs. We perform each lift and solve the error (weighted L2 norm in pressure and height) between the experimental trajectory and each way-point. Force is not included in this error metric because it is prescribed by the chosen mass. The total error between a membrane and the target points is defined as the root-mean square of the nine (3 mass x 3 way-point) errors:
\begin{equation} 
    \label{eq:RMSE}
    RMSE = \sqrt{\frac{1}{n}  \sum_i\left[ \left(\frac{(p_{exp}-p_{target(i)})}{p_{max}}\right)^2 + \left(\frac{(h_{exp}-h_{target(i)})}{h_{max}}\right)^2 \right]}    
\end{equation} 
\noindent where $p_{max}$ and $h_{max}$ are the limits of the space in which we model membrane forces: 10~kPa and 50~mm. $p_{exp}$ and $h_{exp}$ represent the respective pressure and height values from the experimental trajectory that minimize the individual error.

\begin{figure} [!htb]
    \centering
    \includegraphics[width= \linewidth]{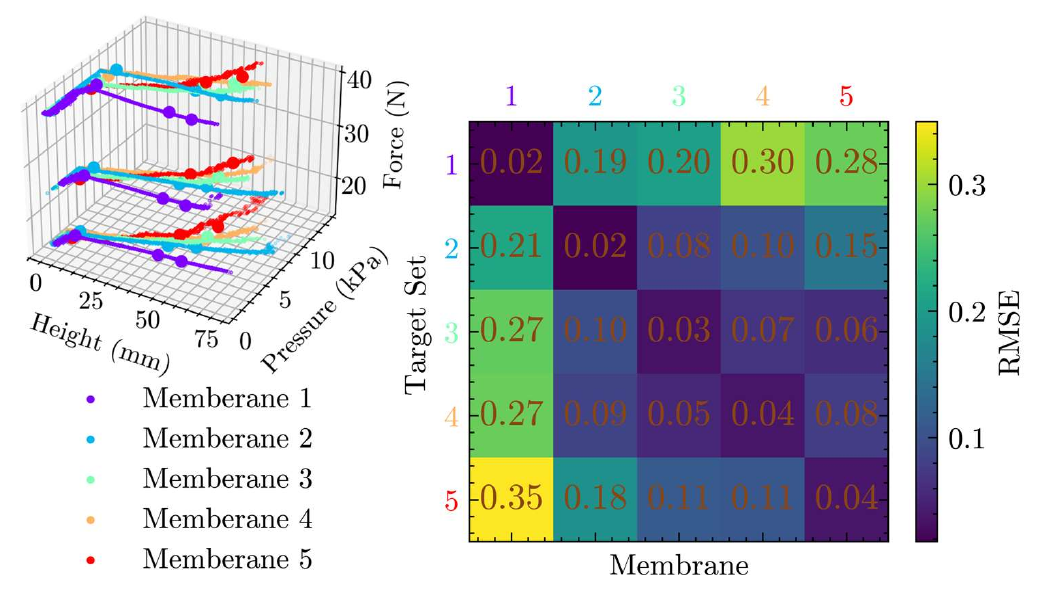}
    \caption{Experimental trajectories: (left) Pressure-height-force data from lifts at three different masses for each of five membranes (lines) with target points for each trajectory (circles). (right) RMSE between experimental trajectories and target way-points for each pairing of membrane and target.}
    \label{fig: Exp_Traj_Error}
\end{figure}

The scaled RMSE for the in-set ringless and ringed membrane trajectories are both 0.02. We then designate trajectories for three membrane designs not in the training set. We choose these membranes specifically to span the pressure-height-force state space. We designate nine way-points for each membrane based on the model. We perform mass testing and error calculation as designated above. The scaled RMSE for these membranes range between 0.03 and 0.04. Parameters for all five membranes are listed above the dashed line in Table~\ref{tab: Mems}. Ring radius describes the radius at the center of the ring, ring width is the distance from that radius to the inner or outer radius ($r_{outer} - r_{inner} = 2 \cdot width$). For testing, the thickness values are restricted between 2.0 and 3.0~mm to minimize popping events. All other parameter ranges remain the same.
The experimental trajectories and their target way-points are shown in Fig.~\ref{fig: Exp_Traj_Error}. The associated errors are also shown in Fig.~\ref{fig: Exp_Traj_Error} - the main diagonal elements represents error along the trajectory for which each membrane was designed, off-diagonal elements represent each membrane's error related the the other membranes' trajectories. The decrease in error for a targeted membrane relative to the average of the other membranes' errors ranges from 69\% to 93\%.

\begin{center}
\begin{threeparttable}
\caption{Mass lift membrane design parameters [mm]}

\begin{tabular}{>{\centering\arraybackslash}p{1.5cm}|
                >{\centering\arraybackslash}p{1.25cm}|
                >{\centering\arraybackslash}p{1.25cm}|
                >{\centering\arraybackslash}p{1.25cm}|
                >{\centering\arraybackslash}p{1.25cm}|
                >{\centering\arraybackslash}p{1.25cm}}
\hline
\makecell{ \\Thickness} & \makecell{Contact\\Radius} & \makecell{Ring\_1 \\ Radius} & \makecell{Ring\_1 \\ Width} & \makecell{Ring\_2 \\ Radius} & \makecell{Ring\_2 \\ Width} \\
\hline
2.0  & 25.4 & nan   & nan  & nan   & nan  \\
2.0  & 25.4 & 49.0 & 5.0 & 62.0 & 5.0 \\
2.3  & 29.6 & 37.6 & 5.0 & 62.0 & 5.0 \\
2.0  & 28.0 & 45.6 & 5.0 & 60.3 & 6.7 \\
2.0  & 38.1 & 47.6 & 6.4 & 62.0 & 5.0 \\
\hdashline  
2.0  & 25.4 & 33.4 & 5.0 & 46.4 & 5.0 \\
2.0  & 31.9 & 46.0 & 5.0 & 59.0 & 5.0 \\
\bottomrule

\end{tabular}
\label{tab: Mems}
\end{threeparttable}
\end{center}

\vspace{6pt}

Using the inherent gradients and fast solve-time of the trained model, we are able to optimize for specific lift goals. As an example, we maximize lift height at given targets of pressure and force. We define the following posterior function~$\Pi$: $\Pi = -k_{force}*F_{error} - k_{pressure}*p_{error} + k_{height}*h_{min}$

\noindent where $k$ are scaling factors and $h_{min}$ is the 'smooth min' score (via the LogSumExp operation of three height values). Local minima are found using gradient descent and compared across 2,500 random starting points. We choose two sets of target points,  with forces across the three target masses (14.7 to 39.2~N). Target pressures are set at 6.9~kPa (set A) and 8.3~kPa (set B). The optimized parameters are listed below the dashed line in Table \ref{tab: Mems}.


We fabricate membranes based on the results of the optimization and test them as described for the five membranes above. We search the lift trajectories for points closest to the target force and pressure then record lift heights for each membrane at these points. A score is given based on the three heights, with a higher score matching a larger height. For each set of target points, the newest membranes had the highest score. Optimal Membrane~A reached a score of 29.4. Among the five prior membranes, four reached all target points for set A and they averaged a score of 14.4. Optimal Membrane~B reached a height score of 40.8 for target set B. Two of the five prior membranes were able to reach the set B target pressure-force combinations, these two averaged a score of 25.7.


\section{Discussion}
This paper uses a machine learning model to inform our design decisions within a parameterized design space for a force application task. We model the theoretical mechanisms governing the expansion of silicone-based SPA and develop a custom Neural Network (NN) architecture for exploring and quantifying the effects of the parameterized design space. The outputs of this model were verified and compared against two other types of models using a dataset gathered with automated experimentation. The trained model is used to define trajectory waypoints for membranes both in and out of the training set, confirmed with experimental mass lifts, and then to optimize for a specific lift output (maximize height). 


\subsection{Class of SPA}

The SPA we characterize in this study provide performance that could enable affordable means of force application for human motion. An actuator with the footprint of an adult head and a contact area smaller than an adult fist, backed by a 5~V battery and a portable air pump, is shown to provide forces over 100~N over a workspace of over 50~mm. Using just 22 membrane characterizations, we are able to train a model that provides 4.0~N RMSE across the entire parameterized design space. For a given force, we can design an actuator that will reach multiple points along a chosen pressure-height trajectory within approximately 4\% error in height and pressure. These metrics indicate the potential for performing pre-meditated lifting trajectories via a simple pressure sweep.

Within the raw data we see the stiffening elements do not increase maximum force output or workspace of the actuators within the tested pressure range, and that they increase the relative pressure needed for a given force or height output. Adding more stiffening elements also increases the potential for membrane fracture due to increased boundary areas. Though neither was studied extensively in this work,  this may decrease both the safe operating pressure for the membranes and their cycle life. While far more lift trajectories with ringed membranes are possible, a system designed using only ringless membranes may survive rougher usage.

Some of the most interesting performance within the collected dataset occur within unconstrained membranes. These membranes converge in pressure and force across different heights. This convergence is due to the hyperelastic constitutive relations of the silicone, and could be an important characteristic in lifting with this class of actuator. This convergence point also represents a difficulty for theoretical simulations, and comparative error was much higher at forces above it. This convergence, and the relatively higher forces per unit pressure from unconstrained membrane compared to strain-limited membranes, both contributed to the higher model errors on this seemingly simpler subclass of actuator.


\subsection{Modeling}

We can predict the way in which embedding concentric stiffening elements will alter the lift trajectory of a silicone membrane, proven by the small predicted error described in Section \ref{subsec: ringed model performance}. It is important to contextualize the performance of our NN model with respect to state of the art theoretical models and off-the-shelf data-regression algorithms. We note that the simpler regression methods are not able to handle combined ringed and ringless data in the way that our NN model does. This and their large errors for ringed cases may indicate that they aren't a good choice for complex design parametrization. Similarly, the rigidity of the boundary condition on the energy minimization approach prevents its success in some ringed cases. Therefore for the ringed design space chosen for this work, the complexity of a NN appears to be warranted. Furthermore, once the NN model is trained, it is just as quick, if not quicker, to query as the other options.

The use of a NN model also avoids a few of the drawbacks of classic modeling methods. Finite Element Analysis (FEA), for instance, has an inherent tradeoff between computation time and accuracy via mesh density. The output of an FEA is also a very data-rich model geometry, which would then need to be post-processed, further increasing computation time. Theoretical modeling inherently relies on bulk material property approximations, which can lead to error due to changes in behavior at different scales or varying properties between different loading conditions (for instance uniaxial and equibiaxial as shown in \citep{rosset2014}).

The NN model design approach brings its own drawbacks, despite its advantages in error reduction. The NN fails to consider any physics of the system, except for those that we’ve coded into parameter limits and output constraints. This means that the results (specifically the output coefficients) are subject to overfitting relative to the input data and have no guarantee of generalizing outside the training distribution. This training distribution is specific to our inputs and sensitive to the decisions we make regarding parameter ranges, experimental conditions, and fabrication limitations (i.e., minimal distance between strain-limiting rings). This limitation is highlighted in many of the optimization outputs aligning with our maximum or minimum limits on parameters. Specifically, the model recommended designs where the rings were as close as possible to either the contact area or the outer housing. Our manufacturing techniques do not allow for  high precision in the placement of the rings, but automated methods might allow us to relax design parameter constraints. This would allow us to answer, for instance, what distance between rings maximizes lift height at different loadings. Further exploration of learning-based approaches in Appendix~\ref{ch: BO_Taguchi} exemplify other issues, including the model's reliance on initialization data and the existence of local minima to an optimization function.

There are also broader issues to consider with an increased reliance on learning methods. The NN represents a complex and difficult-to-interpret function between our design inputs and our force-pressure relationships. While this reduces error relative to an assumption-heavy simplified function, it also deprives us as researchers of important design intuition, making us reliant on the model for making informed design decisions. This lack of intuition makes it difficult to intelligently question the accuracy of the model, especially in domains outside the training set. We must be careful to look at the output parameters that appear optimal for specific tasks to gain further intuition about the importance of different design features. 

The training of machine learning models is also expensive. The training is computationally resource-intensive, which is an unavoidable cost in terms of electrical power. We have also chosen to train purely on experimental data, which is extremely expensive in terms of user time compared to analytical or simulated data. The trained network is efficient relative to the numerical methods used in our theoretical model, but training remains substantially less efficient than purely analytical methods. Notably, experimental data also suffers from aleatoric uncertainty, as discussed in Section~\ref{sec: uncertainty}.

While the results of this particular membrane class remain encouraging, the ability to effectively model any parameterized actuator class with a relatively small (n=22) dataset has wider implications within SPA and soft robotics generally. SPA researchers are using a variety of of multi-material designs to enable different interactions with the world. Active learning could help speed the exploration of these design spaces and lead to precisely tailored designs via inverse design. Classical modeling methods, such as FEA, should still be considered for initial analyses, as they do not require any prior dataset besides material properties. They can also be performed with high throughput to replace experimental tests for training a model that can then perform inverse design (as in \citep{forte2022}). Regardless of the methods, offloading control complexity from the pneumatic input to the mechanical design can enable inexpensive and untethered robots with useful force and motion outputs. 

\section{Conclusions \& Future Work}

We collect an active learning enabled n=22 dataset comparing force, height, and pressure across the design space for 70~mm radius concentric ring strain-limited silicone soft pneumatic actuators. Actuators in this design space are found to apply forces over 100~N and reach heights of over 70~mm. This dataset allows for an empirical model with less error than theoretical energy methods and naive curve-fit models. The empirical model is fully differentiable, which we leverage for design optimization in a height-maximization mass lifting task.

While there is potential for these actuators to be useful individually given their relatively high force and displacement outputs, we foresee a larger potential impact from their use in parallel. Because this model allows us to design for a given force/height pairing at a given pressure, we can connect multiple actuators to a single pressure source and use their embodied response to prescribe their lift trajectories. If we can model the, for instance, rigid body they are lifting, we can also model the kinematics of the lift from a single pressure source. While membranes displayed low hysteresis in our characterization testing, additional work will be required to see what additional force-height-pressure planes can be reached once a set of actuators is attached in parallel.

This model also allows for us to estimate the effect of (de)activating specific strain limiters, so long as the initial and final strain states remain in the design space. This, combined with variable limiters like those used in Chapters \ref{ch: Shape Morphing} and \ref{chap:sheathed}, could be used to alter lift trajectories in real-time for open- or closed-loop lifting. 

While we consider the characterization data modeled in this study to be the most pertinent to soft pneumatic actuator lifting, there were additional datastreams recorded that will be shared on Dryad, including volumetric flow and video of membrane expansion. We hope this data can be useful for better understanding the properties of inflated membranes undergoing concentric one-dimensional loading or as a comparison for improving analytical analyses.

A possible future direction of research using this data is to develop a multi-fidelity model that leverages both experimental and simulated data. Such a model could use large amounts of high-throughput simulated data and calibrate predictions based on a small number of real-world experiments. Such a model has the potential to drastically reduce uncertainty and increase overall predictive accuracy.

\section*{Data Availability}

The majority of the work above was coded in Python, including in Jupyter Notebooks. This code, and a .pkl file containing a dictionary of all the test data used herein, is available on Github: \url{https://github.com/gmcampbell/SPA\_Design}. 

The complete dataset, including tests not included in this paper and video of testing when applicable, is freely available on Dryad: \url{https://doi.org/10.5061/dryad.jsxksn0mt}.

\section{Scope and Perturbations}
\label{sec: Peturbations}

We are encouraged by the results of NN modeling for lift trajectories in the presence of external contact, especially the low prediction errors and distinction between membrane trajectories seen in Fig. \ref{fig: Exp_Traj_Error}. But how narrow is the class of SPA that we have characterized? In this section, we briefly investigate the extent to which lift trajectories change if we relax the following constraints: the embedded contact strain limitation matches the contact plate; the contact plate is round; the membrane is lubricated with cornstarch.

We fabricate two new membranes matching the design parameters of the first (unringed-trial~2) and last (ringed-trial~1) entries in Table \ref{tab: Mems}, except without the central contact strain limitation (i.e., new Contact Radius = 0). We then perform mass lifts with these new membranes, varying the lift constraints on contact plate and lubrication. In one instance (trial~5), we also lift with the initial baseline membrane and a rectangular contact plate. The large circular contact plate is 31.9~mm in radius, the small circular contact plate is 25.4~mm in radius, and the rectangular contact plate is 50~mm x 37~mm. Specific permutations are outlined in Table~\ref{tb: membrane ablations}, with plot colors referencing the results in Fig.~\ref{fig: membrane ablation}.

\begin{table}[!ht]
\centering
\caption{Perturbation Study - Key}
\begin{tabular}{|c|c|c|c|c|c|}
\hline
\textbf{Trial} & \textbf{Ringed} & \textbf{\makecell{Contact Strain \\Limited} } & \textbf{Lubricant} & \textbf{\makecell{Contact Plate \\Shape} } & \textbf{Plot Color} \\
\hline
1 & 1 & 1 & 1 & O (lg) & \tikz\fill[black] (0,0) circle (0.1cm);\\
\hline
2 & 0 & 1 & 1 & O (sm) & \tikz\fill[black] (0,0) circle (0.1cm);\\
\hline
3 & 1 & 0 & 1 & O (lg) & \tikz\fill[blue] (0,0) circle (0.1cm); \\
\hline
4 & 1 & 0 & 1 & $\square$ & \tikz\fill[red] (0,0) circle (0.1cm);\\
\hline
5 & 1 & 1 & 1 & $\square$ & \tikz\fill[green] (0,0) circle (0.1cm);\\
\hline
6 & 0 & 0 & 0 & O (sm) & \tikz\fill[magenta] (0,0) circle (0.1cm);\\
\hline
7 & 0 & 0 & 0 & O (lg) & \tikz\fill[cyan] (0,0) circle (0.1cm);\\
\hline
8 & 0 & 0 & 1 & O (lg) & \tikz\fill[orange] (0,0) circle (0.1cm);\\
\hline
9 & 0 & 0 & 1 & O (sm) & \tikz\fill[violet] (0,0) circle (0.1cm);\\
\hline
\end{tabular}
\label{tb: membrane ablations}
\end{table}

\subsection*{Results and Discussion}

Comparing the rectangular contact plate lifts to the corresponding circular contact plate lifts (trials~5 and 4 to trials~1 and 3, respectively), we see very little deviation between lift trajectories. It is possible that the higher pressure at first height increase and slower increase of pressure with height is completely due to the fact that the overall area of the square contact is smaller than that of the circular contact. In trial~4, however, we do see a particularly large discrepancy when lifting the largest, 4~kg, mass. We stopped this test early due to pinching at the corner of the rectangular contact plate. Specifically, the strained silicone at two separate locations on the membrane met and prevented additional expansion between those two locations. 

\begin{figure}
    \centering
    \includegraphics[width=0.8\linewidth]{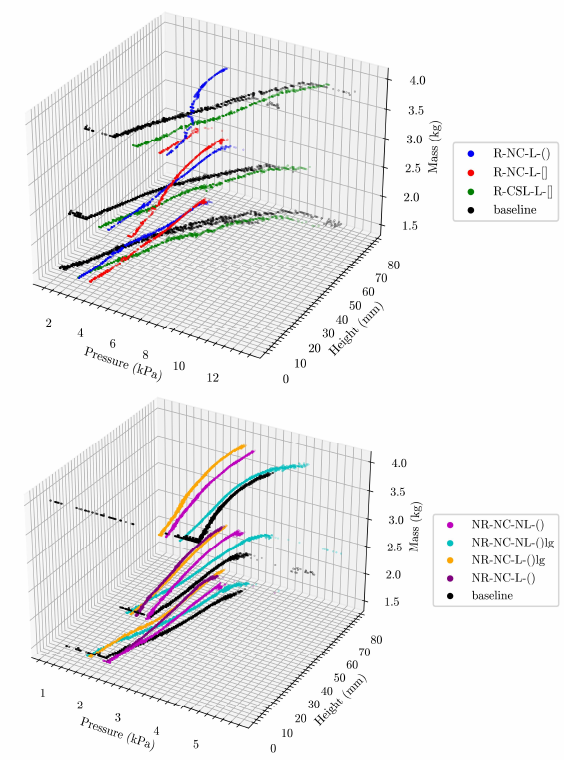}
    \caption{Mass lift results for ablation study across contact strain limitation, lubrication, and contact plate shape. (Top) Ringed membranes. (Bottom) Unringed membranes.}
    \label{fig: membrane ablation}
\end{figure}

Comparing the ringed trials without contact strain limitation (3 and 4) to those with it (1 and 5, respectively), we again see higher pressure at first height increase and slower increase of pressure with height. This variation is the hardest to conceptualize and warrants further investigation. Within unringed trials, we see a gap between the contact strain-limited (trial~2) and unlimited (trial~6) that minimizes for larger weights

The discrepancy between lift trajectories for the lubricated trial~9 and the strain-limited trial~2 implies the membrane area in contact with the plate is still able to strain in trial~9, slipping relative to the plate. This was observed directly during testing. trial~9 exhibits a similar initial lift pressure for the 1.5~kg mass lift but remains at a lower pressure throughout all lifts, similar to the ringed example. Unfortunately the membrane ruptured during the 4~kg test of trial~9, and that data cannot be compared.

The unlubricated result in trial~7 performed very similarly to the corresponding lubricated trial~8 at low masses and pressures, but became more similar to the contact strain-limited, lower contact area, trial~2 at higher pressures or higher mass lifts. The relative performance of trial~7 then implies that the local elastomeric strain within the contact area can be restricted in the unlubricated case in the presence of large internal pressure combined with external force, though perhaps only for a region smaller than that of the contact plate area.

\subsection{Conclusions and `Real' Lifting}
\label{sec: sim2real gap}
The precision of our membrane design modeling appears sensitive to minor perturbations in use-case, which will exacerbate the errors calculated from characterization data. These perturbation discrepancies warrant additional sensing and parameterization to address contact area shape, and investigation into the effect of pinching for shapes with non-continuous edges. These results also indicate that this modeling parameterization falls short of what would be required for less controlled (`real-world') lifting, where contact patches will vary widely.

Measuring and maintaining contact during lifts outside the lab is an open area of difficulty for this type of modeling. These preliminary ablation results are insufficient to  model surface interactions (i.e., tangential friction) and the associated effect on local strain, but these effects will need to be overcome in order to move to human lifting. The effect of external object material properties and perhaps even environmental factors such as humidity add further complexity. Predicting if a contact patch creates pinching, as we saw in trial~4, could also be pivotal to overcoming the sim-to-real gap.

Regarding application of SPA, a class of actuator without contact strain limitation would be useful in a much wider variety of lifts. Perturbation results show that there is indeed a distinct lift trajectory when this contact strain limitation is removed. Furthermore, the mass of the lift and relative pressure is found to be very important for determining the relative effect of lubrication for these instances. It is possible that for some lifts, we could create a transform from the solved model (i.e., baseline lifts) and non-constrained lifts. In particular the similarity of trajectories in trial~2 and trial~7 for the 4~kg mass lift is encouraging.

\section{Parallel membrane lifting}
\label{sec: parallel lifting}

A natural extension of this membrane characterization and inverse design is its application in a shared pressure, multi-membrane system. Co-design between individual membranes can allow for parallel force application and lift trajectories from a single, shared pressure source. We take a preliminary look at this application here, noting potential difficulties with multiple static equilibria. 

\begin{figure}[!ht]
    \centering
    \includegraphics[width=0.8\linewidth]{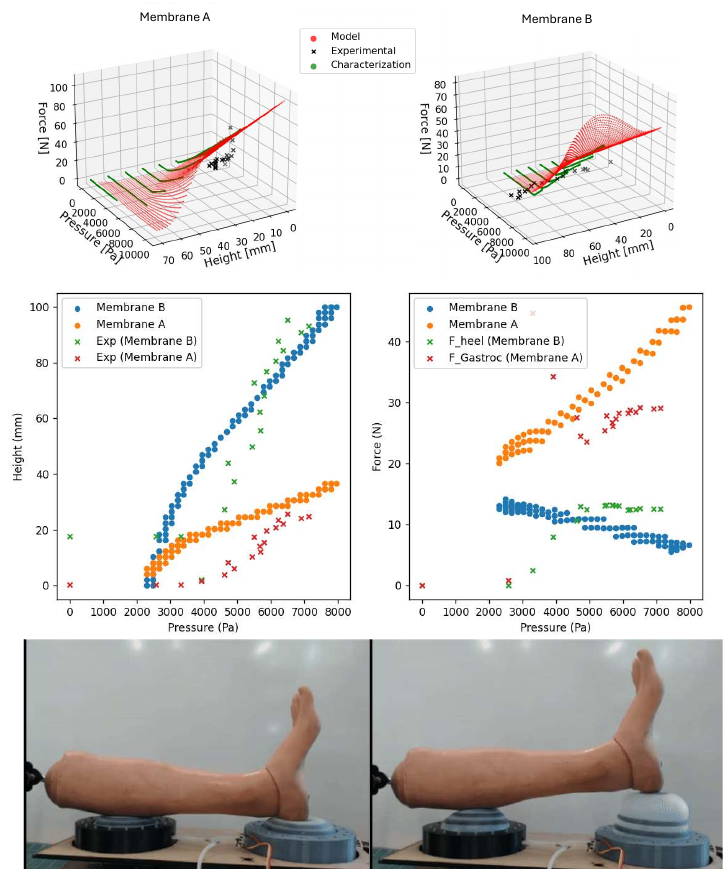}
    \caption{Shank lift experiment using two co-designed membranes. (Top) Modeled trajectory (red) compared with characterization data (green) and lift trajectory (black) for Membranes A and B. (Middle) Modeled trajectory (circles) compared with experimental results (xs) for height (left) and force (right) versus pressure. (Bottom) Low pressure (left) and high pressure (right) states for Membranes A and B (relative left and right) connected to a single pressure source to lift shank.}
    \label{fig: shank lift}
\end{figure}

\subsection{Leg lift}

We test a two-actuator system on a human-scale lift. We mount a 5~kg manakin shank onto a rotary joint. Kinematics of the shank were assumed to follow the perfect revolute joint with even mass distribution throughout the limb. We then positioned two pressure chambers such that membrane contacts would occur at the calf (Membrane~A) and heel (Membrane~B). Two membranes were co-designed using a model trained on all available membrane characterization data to minimize the maximum force experienced at either membrane throughout a single pressure sweep and to maintain contact with the leg throughout. Designs were constrained to membranes that matched the contact area of the calf and heel for Membranes A and B, respectively. 

The optimal membrane parameters are as follows (note Membrane~B was recycled for the mass lifts in model verification):

\begin{table}[!ht]
\centering
\caption{Shank lift membrane parameters}
\begin{tabular}{|c|c|c|c|c|c|c|}
\hline
\makecell{ \\Membrane} & \makecell{ \\Thickness} & \makecell{Contact\\Radius} & \makecell{Ring\_1 \\ Radius} & \makecell{Ring\_1 \\ Width} & \makecell{Ring\_2 \\ Radius} & \makecell{Ring\_2 \\ Width} \\
\hline
A & 3.0 & 38.1 & 46.1 & 5 & 59.1 & 5 \\
\hline
B & 2.0 & 25.4 & 49.0 & 5 & 62.0 & 5 \\
\hline
\end{tabular}
\label{tab:shank_mems}
\end{table}

Membranes are mounted to two 3D-printed and laser-cut mount and port (PLA and acrylic) pressure chambers, which are connected together via pneumatic tubing. This connection is then split to a pressure sensor (MPRLS0025) and an additional line that is further split to a single pneumatic pump and a compressed open-cell foam (McMaster) pneumatic resistor. The pressure chambers are placed on scales that restrict motion with linear ball bearings while measuring vertical force via load-cell (HX-711 breakout). Force and pressure data are monitored via I2C with an ESP32 microcontroller that then passes the data via serial to our Python recording script.

We activate the 5~V pump until Membrane~B is fully inflated (roughly 100~cm) and then manually deactivate the pump, allowing pressure to release through the pneumatic resistor. After the lift test, we perform automated characterization testing for each membrane. Position is calculated from lift video using the Tracker App \citep{tracker2025}.

A comparison between characterization data, model prediction, and lift data is shown in Fig. \ref{tab:shank_mems}. The 2D and 3D trajectory comparison shows that the expected trajectory is followed generally for both actuators, though it is closer for Membrane~B. Looking at error in force, Membrane~A averaged 15.6~N difference between measured and predicted actuator force, while B averaged only 7.3~N.

\subsubsection{Discussion}

Our predictions for actuator force output in the leg lift scenario erred from reality by two to four times the calculated error of our model alone. Yet we were able to predict single-membrane, one-DoF lifts to within 2~\% RMSE. In this section, we will discuss assumptions and shortcomings that explain the difference between performance in the highly controlled lift experiments and this step towards human lifting. 

A number of assumptions would have to hold for the membrane pressure response in the leg lift to match that in controlled experimentation. We assume that the membranes maintain identical, quasi-static pressures and that we can measure these pressures precisely at a sensor upstream of the chambers throughout inflation. We assume simple kinematics of the leg and that we can measure this precisely from video recordings. We assume the leg is perfectly rigid and its mass is evenly distributed. We assume the contact area of the calf and heel remains constant and perfectly matches the corresponding contact regions of the respective membranes. We assume that force is transmitted completely through the membranes, and not at all through the rigid membrane housings. We assume that, due to the small angle assumption, the angular lift is equivalent to lifting straight upward. Finally, we assume a single pressure response for each membrane.

The error in the 2.5 to 4.5~kPa region of the trajectory could be best explained by the leg remaining in contact with the rigid membrane housing (beneath the membrane) during the initial portion of the lift. This explains both the flat height curve (not lifting off the rigid contact) and the significant error in force prediction (Membrane A able to `lift' far beyond its predicted pressure response due to superposition of forces from the housing). The large initial force on Membrane A implies that load distribution at initial contact was significantly different from what was assumed during modeling. This alternative distribution is possible only because of the external contacts mentioned above and the geometry of the leg. This real-world consideration will need to be accounted for in system modeling, as opposed to membrane modeling.

Many other factors contribute to error in the model, and one such breakdown of assumptions occurs in the estimation of the contact areas between leg and actuators. Careful consideration of the photographs in Fig.~\ref{fig: shank lift} reveals that the contact between shank and membrane is not only non-circular but varies greatly throughout the lifting task. As discussed in Section~\ref{sec: Peturbations}, changing the contact patch affects lift trajectory, even if the strain limitation patch of the membrane remains constant. Both the shape and size of the contact varying throughout the lift likely contributes significantly to variation between this trajectory and the modeled lift. More precision in the parameterization of the contact patch, for instance by additional sensing, could mitigate this error.

Other model assumption breakdowns likely contribute to error, as well. We see a divergence between the faster inflation (shown in Fig.~\ref{fig: shank lift}) and slower deflation trajectories that could be partially explained by the dynamic flow of air leading to variable pressure between the sensor and the membranes. We use inflation (as opposed to deflation) data from membrane characterization, but during that testing, the pressure sensor is within the pressure chamber, eliminating the potential for this measurement error. More precise height measurement (for instance, time-of-flight or even Vicon) could also guarantee precision for future trials.

Despite its shortcomings, this lift represented the most exciting part of my thesis work \footnote{Actually, I lifted my own leg with this same setup. That was the most exciting part.}. This system of two actuators not only lifts a human leg equivalent with a small 5~V pneumatic pump but also enables us to design and optimize the trajectory of that lift by differentiating a model trained only on individual membranes. With additional work to tie in redundancies and application across varying human kinematics, this bolsters my hope that SPA can have real-world impact.

\subsection{Two mass perturbations}

\begin{figure} [!ht]
    \centering
    \includegraphics[width=0.65\linewidth]{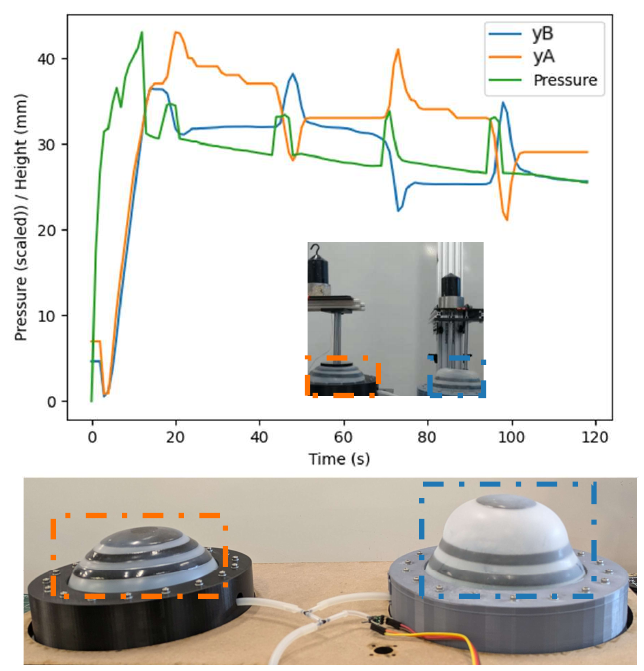}
    \caption{Equilibrium experiment for multiple membranes connected to a single pressure source each lifting 2~kg in parallel. (Top) Effect of external perturbations on internal pressure (green) and equilibrium position of membranes (blue and orange). (Bottom) Membranes used in equilibrium experiment, shown in unladen, inflated configuration at shared internal pressure. (Inset) Membranes performing lift experiment - prior to perturbation.}
    \label{fig:2mem}
\end{figure}

We perform another experiment to investigate whether multiple trajectories exist for a system of multiple membranes. We inflate both membranes while they each lift 2~kg of mass on a 1-DoF test rig (with rig mass included, they lift 2.5-3~kg). We then perturb a membrane by physically displacing the balloon downward in an attempt to reduce strain. We first press Membrane~A at approximately 20~seconds, then Membrane~B at approximately 45~s, then A, then B. Ignoring the slow pressure leak, we find the equilibrium heights of both membranes shift as a result of the perturbations, implying that there are at least two valid sets of membrane heights for this system of two actuators. Given that we saw low hysteresis for single-membrane systems during inflation and deflation, this is likely attributable to the system integration itself.

The existence of multiple system equilibria complicates the scaling of our individual membrane modeling. Future work should investigate which system factors contribute to the additional equilibria and how to control which set of trajectories is followed. This direction is potentially exciting, as it could enable further versatility from these systems.


\chapter{CONCLUSION}

In this thesis, we have made progress towards the design of useful elastomeric soft pneumatic actuators (SPA) with both active and passive strain limiters. With the low-cost designs and simple pneumatic inputs we use throughout this work, it is our hope that these contributions broaden the path towards accessible SPA.

We first investigate electroadhesive (EA) clutches as a means of rapid strain limitation - developing a prototype SPA for shape variation and low-mass manipulation. This prototype highlights the scope of  shapes achievable by externally strain-limited elastomeric surfaces, as well as many of the practical drawbacks of EA strain limitation for bilaterally expanding elastomer. Some of these drawbacks are addressed in the next chapter when we design the passive strain response of sheathed EA clutches and characterize their active response to more complex electrical inputs. The prototype applying this sheathed clutch to an externally limited tubular SPA gives context for potential applications in position control.

We then add external force application to our inflation trajectories by focusing on a class of concentrically strain-limited elastomeric surfaces lifting in one degree of freedom. Using strain energy theoretical methods with characterized constitutive relations, we can predict the pressure-height-force relations across our design space. Using active-learning and automated experimentation, we validate our theoretical simulations and efficiently train a lightweight neural-network-based model that outperforms them. This differentiable model allows us to optimize design parameters for target lifts, which we again validate experimentally. 


\section{Future Work}

\subsection{Generalizing soft pneumatic actuator modeling for design}

There is an inherent trade-off between model breadth and model accuracy for complex systems. In Chapter \ref{ch: Membrane_Design_Section} of this thesis, we have shown that machine learning techniques can outperform theoretical methods for a parameterized design class of SPA. However, this is a very specific result. To model an identical class with a different hardness silicone, the baseline theoretical simulation should require only an update of material parameters. Our pipeline, on the other hand, would require an entirely new dataset - potentially months of experiments. Our differentiable model offers increased accuracy and the ability for inverse design, but these benefits need to be generalized if they are to fully inform design decisions.

To generalize our model beyond the limitations of our parameterized design space, I suggest we define a minimal element that is relevant across all rows of Table~\ref{tb: Lit Review}. This has been achieved theoretically for a subset of fiber-reinforced tubes and surfaces in prior work \citep{connolly2017, sedal2021, sholl2021} and is detailed for a subset of fabric-reinforced surfaces in this thesis. Inverse design within this super-set of functional strain-limited elastomers could allow for better-targeted embodied response and more effective transitions with the (de)activation of active strain-limiting elements. To this end, a differentiable architecture like the NN used in this thesis may be a viable path forward, if it has access to sufficient training data.

To improve the accuracy of this generalized model, it will be important to incorporate experimental data in addition to the theoretical. Multi-fidelity modeling achieves this precisely, incorporating trends that are overlooked in theoretical models alone. A well-trained network could then move towards generating hybrid designs across rows of Table~\ref{tb: Lit Review} to solve challenges in soft lifting. This type of model could even be incorporated into existing soft robotic simulation platforms to  decrease the sim-to-real gap.

\subsection{Lifting beyond one degree of freedom}

As mentioned at the end of Chapter~\ref{ch: Shape Morphing}, it is helpful when dealing with SPA external contacts to simplify complex actuator shapes down to a single output degree of freedom. This greatly simplifies the modeling, but also makes the resulting lifts much less versatile. As we investigate at the end of Chapter~\ref{ch: Membrane_Design_Section}, using multiple of these simplified actuators in parallel along a rigid body can lead to interesting manipulation trajectories assuming that we know the kinematics of the body being lifted.

Advances could be made along this direction of rigid manipulation with timely activation of additional membranes, for instance when a specific state (i.e., internal pressure) is reached. This can be done actively (for instance via deactivation of an EA clutch) or passively (for instance with a well-designed soft pneumatic valve as we investigate in Appendix~\ref{ch: BO_Taguchi}). Complex manipulations can then be planned during actuator system design and executed repeatably on the multiple-actuator system without need for any additional SPA modeling.

As we apply these systems to human lifting, in particular, we can alter designs for increased regions of stability and redundancies among actuators. Inverse design for this stability is perhaps the biggest area of potential improvement for soft robotics over current commercial systems (i.e., HoverTech). If we can quickly query lift states that are nearby our target lift to predict how many will still provide restorative forces, we can judge how stable our lift would remain during a perturbation.

\subsection{Sensor-informed systems}

Preliminary results from model-informed lifting based on strain-limited membrane characterization indicates that additional data streams could improve real-time modeling for this class of actuator. In our leg lift (Section~\ref{sec: parallel lifting}), we rely only on pressure data. The relationship between forces and heights over the pressure sweep are then theoretically set by membrane designs and system kinematics. We can theoretically update our prediction of how the SPA will respond to the pressure input in real-time by measuring non-modeled variables or modeled inputs that may change over the course of a lift.

Sensing for contact region will be important for quantifying the difference between expected and actual contact area, which affects associated pressure response. We embed time-of-flight sensing into a pressure chamber for this class of SPA in Chapter~\ref{ch: Shape Morphing}, and in other works we use multimodal sensing to output depth (as well as proximity) data during contact \citep{yin2022}. This type of sensing would fit seamlessly into the system used for characterization in Chapter~\ref{ch: Membrane_Design_Section}, though would scale poorly as we move to multiple membranes lifting in parallel.

With the implementation of membrane surface sensing, we could also us complete membrane geometry as a high-dimensional input for machine learning. With membrane design parameters and a real-time surface geometry as state inputs, pressure, force, and height could all be modeling outputs. Similar to the anisotropic strain we used for directional lifting in Chapter~\ref{ch: Shape Morphing}, relative strain between different areas on the membrane could then inform direction of lift as total strain informs the magnitude. In a training context, the input state would change to include membrane design parameters, contact patch, pressure, and high-dimensional surface geometry, while the output state would include multidimensional force and height. Alternatively, input could remain as pressure and membrane design parameters and the model could attempt to learn the entire membrane geometry, more similar to \citep{forte2022}, to imply height and force.

Low-dimensional approximations of membrane geometry may be sufficient for training and would avoid the requirement for expensive equipment that scales poorly for shared-pressure (i.e., multi-membrane) systems. For instance, spacial membrane strain data could serve as embedded sensing to eliminate the need for an enclosed time-of-flight sensor. Reducing sensing complexity even further, we could monitor multidimensional force output with a load-cell and model lift trajectory based on pressure and force. Alternatively, the angle $\beta$ could be an excellent one-dimensional state input to model the force-height relation for a given pressure(though this is limited to one-dimensional lifts). Each of these data streams scale linearly with the number of membranes in the system, trading increased complexity and computation for adaptive, higher-accuracy models.

The next step after embedding sensing could be to increase control authority to react to sensing inputs, either by active strain limitation or additional pressure inputs. This closed-loop system could then be controlled for more precise trajectories, similar to \citep{marchese2016}, in the presence of external force. Notably, the sensing discussed above would all be embedded in the membrane or associated frame, allowing the system to theoretically be used outside a motion tracking setting. The disadvantage of this closed-loop system is the significant engineering complexity compared to the single pressure sensor open-loop system used in this thesis. While these closed-loop systems may be interesting in a lab setting, it will be a long time before they will be able to compete with the precision of equally complex rigid systems for useful applications outside the lab.





\section{Outlook}
Low cost and untethered solutions for soft pneumatic actuators (SPA) should aim to off-load control complexity into the mechanical membrane - designing for an embodied response to changes in pressure or force. Otherwise, as in classical robotics, variable control remains dependent on the quality of the actuator (pump). In this thesis, we focus on solutions specifically within strain-limited elastomeric surfaces, but this is not the only viable medium. The poor fatigue life of elastomers with embedded strain-limiters undergoing large strains, in particular, leads me to believe that the ultimate solution may lie elsewhere. 

We (and others) have shown electroadhesive (EA) clutches as a potential strain limiter for SPA, but it is worth emphasizing that they are currently viable only alongside crumpled plastics (not undergoing strain) or the softest elastomers. For EA clutches to become viable in tandem with stiffer materials and corresponding higher pressures, they will need continued advances in the magnitude and reliability of their holding force. As they continue to develop for higher forces and lower voltages, I foresee EA clutches becoming a very versatile active strain limiter that is attractive for SPA in terms of cost, power usage, and footprint.

Machine learning will continue to permeate more and more aspects of mechanical design (and human decision making generally). Active learning seems particularly exciting for gaining efficiency in mechanical design, so long as the design space can be parameterized and limited effectively. As we incorporate these methods, we need to be careful to acknowledge the limitations and biases of the tool. And for rapid prototyping especially, it is best when models can be validated with physical experiments. If these datasets can then be made open-source, there will be even greater opportunity for well-trained models in the future.


\end{mainf}

\begin{append}

\chapter{ACTIVE LEARNING FOR DESIGN OF EXPERIMENTS} \label{ch: BO_Taguchi}

This section includes ongoing work that will be submitted as:

Gregory M. Campbell, Yi Cao, Hannah Escritor, Zihao Zhou, and Mark Yim. Active Learning for Rapid Prototyping: Model
Uncertainty in Design of Experiments.

My contribution to this work involved leading the design and implementation of the research approach, the design and implementation of the testing and optimization software, partial design and implementation of the hardware test systems, data post-processing, data analysis, and the majority of the writing.\\

\section{Introduction}

The success of active learning (AL) for actuator design in Chapter~\ref{ch: Membrane_Design_Section} implies that machine learning can enable mechanical design optimization generally. However, without a baseline or metric of improvement, it is difficult to quantify the success of such methodology. In this ongoing work, we aim to provide quantitative analysis on the performance of Bayesian Optimization (BO) relative to principled Design of Experiments (DoE). We also analyze relative decrease in epistemic uncertainty, which we quantify by model uncertainty, as new training data is provided to the model.

Rapid prototyping and additive manufacturing provide a time-efficient means of product innovation and development, but they often suffer from a pronounced disconnect between theory and reality compared to carefully machined prototypes. While theoretical models give insight into the mechanism and broad effects of design parameters, manufacturing imprecisions, and ductile materials negatively affect theoretical precision in many applications \citep{yang2015}. Prototypes and experimentation become instrumental to the discovery of smooth functionality and optimization, and efficient experimentation can both relieve the cost of this endeavor and further accelerate development. 

Researchers have employed a principled Design of Experiments (DoE) to quantify and query parametric design spaces - adding efficiency when capturing data is difficult or expensive. These approaches aim to span the design space with relatively few experiments, sometimes including statistically motivated tools \citep{montgomery2012}. Similar methods have been developed to include robustness to noise factors that are outside the researcher's control, a process termed Robust Parameter Design (RPD) \citep{robinson2004}. One RPD tool in particular, Taguchi orthogonal arrays, are easily applied for varying numbers of parameters and levels, allowing for their continued relevance in the 21st century \citep{ilzarbe2008, arboretti2022}. This method notably discretizes all continuous parameters, allowing for the mixed investigation of continuous, integer, and categorical elements. 

Parallel to DoE, the growth of machine learning (ML) has allowed for precise surrogate models to approximate the effects of design choices in even the most non-linear of interactions \citep{campbell2025, feizi2022}. These surrogate models allow for interpolation between sampled design combinations and very fast speeds at inference time. ML has been applied to data collected through DoE primarily to increase prediction accuracy, with the most common ML algorithms using neural network (NN) \citep{arboretti2022}.

Sequential experimentation (adaptive sampling) and active learning provide means of  collecting new data to minimize overall error of a ML model \citep{ren2021}. Beyond model improvement, Bayesian Optimization (BO) uses targeted acquisition functions with a Gaussian Process Regression surrogate to maximize the likelihood of improving a specific output metric during each experiment. BO has become widely accessible via open-source software \citep{greenhill2020}, notably BoTorch \citep{balandat2020}. 
BO approaches substantially reduce the number of experiments required for optimization \citep{wang2022}, including within the areas of mechanical design of rapid-prototyped elements \citep{gongora2020} and manufacturing hyperparameters using continuous parameterizations \citep{deneault2021}. Using BoTorch, Meta's Adaptive Experimentation Platform (Ax) has provided an accessible platform for BO \citep{bakshy2018} that inherently combines continuous elements with categorical parameters, but its efficacy in mechanical design applications has not been thoroughly explored.

In our first case study, we co-design categorical, integer, and continuous design and manufacturing parameters to maximize a single objective combination of gear ratio and backdrivability for a compound Wolfrom bilateral gearbox \citep{matsuki2019}. However, instead of using traditional machining of metal gears as they were designed using machine learning techniques in \citep{matsuki2019}, we develop a 3D printing compatible version. Gear-based metamaterials have been proven effective for stiffness-tuning \citep{fang2022}, and could also benefit from this rapid-prototyping approach.  3D printing is much more accessible and lower cost but presents significant challenges that are difficult to model with standard gear design.

In the second case study, we vary categorical and continuous design parameters to maximize multiple objectives relevant to the crack pressure and steady-state pressure difference across an injection-molded silicone check valve. We again impose system-match specifications (tubing, pump, pressure chambers, release valve), limiting the effectiveness of existing theoretical models \citep{laake2022, laake2024}. Specifically, the use of a pneumatic pump couples the control of pressure and flow, while set system components require that all changes be applied at the valve specifically, unlike in \citep{laake2022}.

We apply surrogate model uncertainty quantification to analyze the effect of input data on model training progression for both case studies. We also apply parameter sensitivity analysis to highlight important features in each case. We provide context for the relative advantages of BO and Taguchi methods in rapid prototyping for mechanical design, and highlight questions that are important to future applications.

\section{Modeling \& Parameter choice}
\label{sec: BO_Theory}

\subsection*{Gears}
We wish to design a compound Wolfrom bilateral gearbox with high back-driving efficiency at high reduction ratio, for the purpose of versatile robot actuation. This compound gearbox is composed of two coaxial planetary gearboxes. The sun gear is present only on the forward-drive input side. We refer to the forward-drive input side of the compound gearbox as gear 1 and the output side as gear 2. 
The geometry of each individual gear in the compound gearbox is fully specified by its module M, tooth number Z, and profile shift coefficient X. Assigning fixed module values a priori, there are then a total of 10 free design parameters, namely: \[X_{p1}, X_{p2}, X_{r1},X_{r2},X_{s},Z_{p1}, Z_{p2}, Z_{r1},Z_{r2},Z_{s},\] where the subscript r denotes the ring gear, the subscript p denotes the planet gear, and the subscript s denotes the sun gear. However, these 10 parameters are not pairwise independent. The coaxial geometry design imposes geometry constraints that allow us to render 4 of the 10 parameters dependent on the remaining 6. Furthermore, we will fix the design of ring 1 throughout the study for ease of experimentation, which reduces the number of free design parameters further from 6 to 4. 
The original 10 design parameters are subjected to the coaxial geometry constraint:
\begin{equation}
\label{eq: radii_eq}
     r_a = r_b =r_c
\end{equation}
where \(r_a, r_b, r_c\) are respectively the center distances between $s$ and $p_1$, $p_1$ and $r_1$, and between $p_2$ and $r_2$. 
Let subscript t denote the tip diameter and subscript r denote the root diameter, and we introduce a variable clearance C between the meshing gears. Then 
\begin{equation}
\label{eq: r_a}
     r_a = r_{s_t}+r_{p_r}+C_{s\rightarrow p1}
\end{equation} 
\begin{equation}
\label{eq: r_b}
     r_b = r_{r1_r}-r_{p1_t}-C_{p1 \rightarrow r1}
\end{equation} 
where $C_{s \rightarrow p1}$ denotes the clearance between sun and planet one and $C_{p1 \rightarrow r1}$ denotes the clearance between planet one and ring one. 

For profile shifted spur gears, radius is related to the profile shift coefficient \cite{matsuki2019}, with $+1$ for tip and $-1$ for root:
\begin{equation}
\label{eq: profile_shifted_r}
     r_i = \frac{m_i}{2} (Z_i+2x_i \pm 1) 
\end{equation} 

Substituting Eqn. \ref{eq: r_a} \ref{eq: r_b} \ref{eq: profile_shifted_r} into Eqn. \ref{eq: radii_eq}, we can expand $r_{a}=r_{b}$ as 
\begin{multline}
\label{eq: r_a r_b}
   \frac{m_1}{2} \big[(Z_s+2X_s+1)+(Z_{p1}+2X_{p1}-1)\big] + C_{s\rightarrow p1} =\\
   \frac{m_1}{2} \big[(Z_{r1}+2X_{r1}+1)-(Z_{p1}+2X_{p1}+1)\big] - C_{p1\rightarrow r1}
\end{multline}
where $m_1$ is the module of the sun, ring one, and planet one gears. 
Likewise, $r_{b}=r_{c}$ can be expanded as 
\begin{multline}
\label{eq: r_b r_c}
   \frac{m_1}{2} \big[(Z_{r1}+2X_{r1}+1)+(Z_{p1}+2X_{p1}+1)\big] + C_{p1\rightarrow r1} =\\
   \frac{m_2}{2} \big[(Z_{r2}+2X_{r2}+1)-(Z_{p2}+2X_{p2}+1)\big] - C_{p2\rightarrow r2}
\end{multline}
Rearranging and simplifying Eqn. \ref{eq: r_a r_b} and \ref{eq: r_b r_c} we obtain 

\begin{equation}
\label{eq: r_a r_b r_c simplified1}
 Z_{p1}+2X_{p1}= \frac{1}{2} [(Z_{r1} +2X_{r1})-(Z_{s}+2X_{s})]- \frac{C_{s\rightarrow p1}+C_{p1\rightarrow r1}}{m1}
\end{equation}

\begin{multline}
\label{eq: r_a r_b r_c simplified2}
 Z_{p2}+2X_{p2}= \frac{m_1}{m_2} [(Z_{p1} +2X_{p1})-(Z_{r1}+2X_{r1})]+(Z_{r2} +2X_{r2}) +2(C_{p1\rightarrow r1}-C_{p2\rightarrow r2})
\end{multline}
As shown in Equations 7 and 8, the coaxial constraint Eqn.\ref{eq: radii_eq} renders $Z_{p1},Z_{p2},X_{p1}, X_{p2}$, dependent on the remaining 6 design parameters and the variable clearances. In addition, we fix the values of $Z_{r1}$ and $X_{r1}$ throughout the Taguchi trials for ease of experimentation, which leaves our free design parameters:
\[X_{s}, X_{r2}, Z_{s}, Z_{r2}\] 
These four parameters are the primary design variables in our Taguchi array for the compound Wolfrom gearbox and fully determine the right-hand side 
Eqn~\ref{eq: r_a r_b r_c simplified2}. We can then solve for the values of the remaining independent variables $X_{p1}, X_{p2}, Z_{p1}, Z_{p2}$. Because the number of gear teeth must be an integer value and we restrict X to be between -1 and 1 in this study, there are then only four valid combinations of $Z_{p}$ and $X_{p}$. Among them we define the standard solution to be $Z_{p_s} = floor(Y)$ and $X_{p_s} = \frac{1}{2}(Z_{p_s}-floor(Y))$ where Y denotes the right-hand side of 
Eqn \ref{eq: r_a r_b r_c simplified2}. The remaining 3 valid solutions contain $Z_{p}$ values that are integer offsets from the standard solution $Z_{p_s}$, quantified by offset parameters $p_{1~offset}$ and $p_{2~offset}$ that take the integer range $[-1,2]$. We include these offset parameters as secondary design variables in our Taguchi array, along with the variable clearances. We enforce all clearance values to be equal and define a single clearance variable $Cl$. 

In addition to the above design parameters, we also study the effects of manufacturing choices. We define the Taguchi variables material and lubricant with 2 and 4 levels respectively, along with the gear thickness variable $g_{t_h}$ that takes integer values between 3 and 6 mm. These variables primarily affect friction between meshing gears, which is a critical factor influencing the driving efficiencies of the gearbox. We assume an average coulomb friction constant \cite{yada1997}, which directly affects drive efficiency, as calculated in \cite{matsuki2019}.

We define the basic driving efficiency $\eta_0$ of a pair of involute gears as the ratio of power output to power input, and simplify the expressions using no-slip contact condition on the pitch circle. 
\begin{equation}
\label{eq: F_efficiency}
\eta_0 = \frac{\omega_{out}\tau_{out}}{\omega_{in}\tau_{in}}=\frac{f_{out}}{f_{in}}; 
\end{equation}
where $f$ is the tangential force, $\omega$ is the rotational velocity, and $\tau$ is the gear torque from the driving (in) and driven (out) gears. 

From kinematics we have the following gear ratio equation, which also helps define efficiency: 
\begin{equation}
\label{eq: gear ratio}
G_r = \frac{1-I_2}{1+I_1}; 
\end{equation}
where 
\begin{equation}
\label{eq: ratio I}
I_1 = \frac{r_{r1}r_{p2}}{r_{r2}r_{p12}}=\frac{Z_{r1}}{Z_{s}};~~ 
I_2 = \frac{r_{r1}r_{p11}}{r_{s}r_{p12}}=\frac{Z_{r1}Z_{p2}}{Z_{r2}Z_{p1}}; 
\end{equation}
\textit{$r_{p11}$ and $r_{p12}$ are respectively the pitch circles of planet one when engaged to sun and ring one.} \\

In steady-state operation, the net torque acting on the gears is zero, allowing us to solve for all forces at the contact points as a function of radii, basic driving efficiencies, and the input torque. Reducing these equations, see \cite{matsuki2019}, yields the back-drive efficiencies $\eta'$ of the Wolfrom gearbox as follows:

\begin{equation}
\eta' = \frac{\tau_{\text{out}} \omega_s}{\tau_{\text{in}} \omega_{r2}} = \frac{(1 + I_1) \eta_a (\eta_b \eta_c - I_2)}{\eta_c (\eta_a \eta_b + I_1)(1 - I_2)}
\label{eq:eta_backward_75}
\end{equation}
for $I_2<1$, and 
\begin{equation}
\eta' = \frac{(1 + I_1) \eta_a (1 - \eta_b \eta_c I_2)}{(\eta_a \eta_b I_1)(1 - I_2)}
\label{eq:eta_backward_76}
\end{equation}
for $I_2 >1$. Where:
\begin{equation}
\eta_i = 1 - \mu \pi \left( \frac{1}{z_{i1}} + \operatorname{sgn} \frac{1}{z_{i2}} \right) \epsilon_i, \quad i \in \{ a, b, c \}
\label{eq:eta_i}
\end{equation}
and $a, b, c$ are defined as:\\
\[a_1 = s, ~a_2 = p_1, ~b_1 = p_1, ~b_2 = r_1, ~c_1 = p_2, ~c_2 = r_2\]


Our final Taguchi array contains the following 10 variables. The numerical range or level of the variables are given in Table I. 
\[Z_{s},Z_{r2},X_{s},X_{r2},p1_{offset},p2_{offset},g_{t_h}, Cl, material, lubricant \]

\begin{figure}
    \centering
    \includegraphics[width=0.45\linewidth]{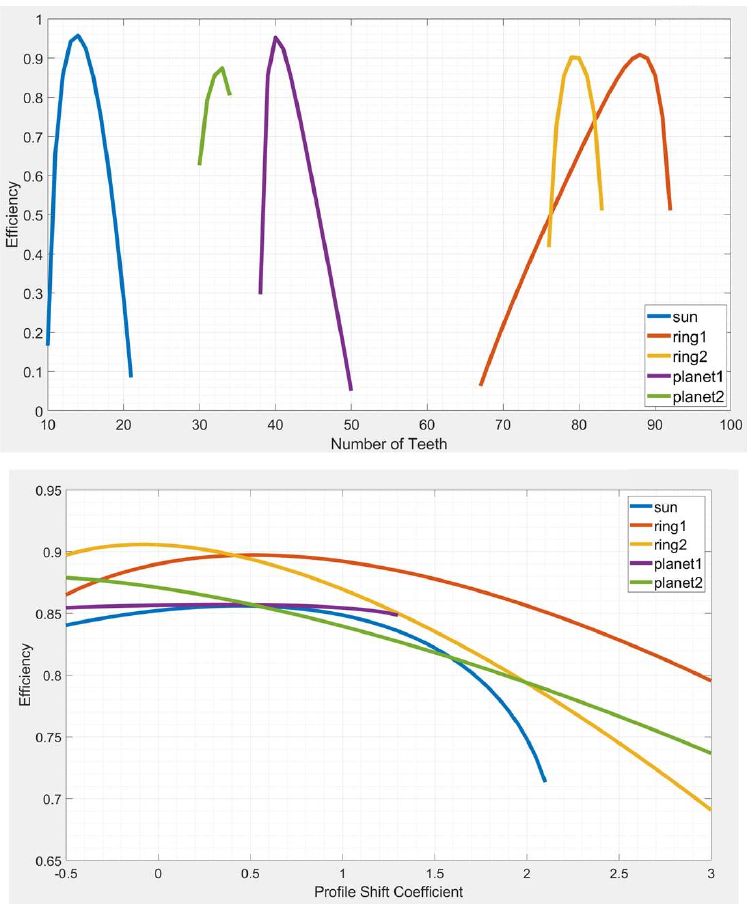}
    \caption{One-factor-at-a-time analysis of gear efficiency for (top) number of teeth and (bottom) profile shift coefficient }
    \label{fig:eff_prof}
\end{figure}

We can use Eqn.~\ref{eq:eta_backward_75} and \ref{eq:eta_backward_76} to predict driving efficiencies to inform our parameter ranges. Fig.~\ref{fig:eff_prof} shows the response of back-drive efficiency to change in teeth number and profile shift coefficients of individual gears from a standard set-up of\\
\[
\begin{aligned}
&z_s = 12, \quad z_{p1} = 39, \quad z_{r1} = 90, \quad z_{p2} = 32, \quad z_{r2} = 81 \\
&x_s = 0.48, \quad x_{p1} = 0.76, \quad x_{p2} = 0.54, \quad x_{r1} = 2.0 \\ 
&x_{r2} = 1.21
\end{aligned}
\]

\subsection*{Valves}

We wish to design a soft check valve that achieves a crack pressure of 10~kPa and a steady-state pressure of 5~kPa for the pneumatic capacitance and resistance of our specific two-chamber test system. These values are assigned to enforce a delayed activation of the second pneumatic chamber and a targeted offset in relative inflation trajectories.

The valve consists of a single silicone body with dimensions determined by injection molding. The functional properties of the valve depend on mechanical and geometric properties of the silicone body, and also on fluidic properties of the pneumatic system \cite{laake2024}. 
We hold constant the fluidic properties of the pump, piping, pneumatic capacitance, and pneumatic resistance. 

The mechanics of the dome are modeled as the superposition of a linear spring with stiffness $k_s$ and a rotational spring $k_R$. The apex of the dome, node $P$, is defined by the position $x$, which is considered 1-dimensionally and radially towards the center of the dome. The equilibrium position of $x$ is determined from a force balance between the pneumatic pressure ($\Delta p$) applied over the area of the dome ($A_V$) and the mechanical spring force $F_s$, which is defined as the derivative of the silicone dome's elastic energy, $E$, with respect to $x$ \cite{laake2024}:

\begin{equation} 
\label{eq: valve_spring}
\begin{aligned}
F_S = \frac{dE}{dx} = 
&- k_S \left( 
\frac{ \sqrt{h_{dome}^2 + b_0^2} - \sqrt{x^2 + b_0^2} }{ \sqrt{x^2 + b_0^2} \ } 
\right)x \\
&+ \frac{k_R}{b_0} \left[
 \frac{\arctan\left(\frac{h_{dome}}{b_0}\right) + 
\arctan\left(\frac{x}{b_0}\right)}{\frac{x^2}{b_0^2} + 1} \right]
\end{aligned}
\end{equation}
where $h_{dome}$ is the no strain dome height and $b_0$ is the constant dome radius. Crack pressure is then defined by the pressure that supplies sufficient force to drive $x$ to 0: $\Delta p_{crack}*A_V - F_S(x_*) = 0$, 
where $x_*$ is the valve crack position of $x=0$ and $\Delta p_{crack}$ is the crack pressure.

We begin with design parameters $h_{dome}$ and $b_0$ based on the printable files provided with \cite{laake2022}. Fixing the valve radius to allow for consistent testing equipment, the curvature of the elliptical dome is then described completely by the dome height. The material constants $k_s$ and $k_R$  parameterize based on the silicone material and the thickness of the valve, which we set to be constant for the entire piece. Geometry and length of the cuts also affect the material constants. Geometry represents the number of cuts from the center, while the length is the length of each of those cuts measured in the tangential plane of the dome from the point $P$.

$k_s$ and $k_R$ are solved experimentally in \cite{laake2024}. Using their reported values of $k_s =  2.9 \times 10^7 \frac{N}{mm}$ and $k_R = 3.6 \times 10^8 \frac{N mm}{rad}$ for Dragon Skin 20, we solve for a cracking pressure of \hbox{$\Delta p = 1.5 \times 10^9~kPa$} for a 3~mm $h_{dome}$ and 3.75~mm $b_0$. This result is orders of magnitude higher than the experimental value, or the pressure capabilities of our system ($< 100~kPa$). Rather than re-solving these constants for every possible combination of materials, thickness, and cut, we focus our experiments on matching the pressure characteristics relevant to our use.

\subsection{Parameters}

\begin{table}[ht!]
\centering
\begin{minipage}[t]{0.48\columnwidth}
\centering
\caption{Gear Parameters}
\label{tb:parameters}
\renewcommand{\arraystretch}{1.1}
\begin{tabular}{|m{2.4cm}|m{3cm}|}
\hline
\textbf{Parameter} & \textbf{Range/Options} \\
\hline
\multicolumn{2}{|c|}{\textbf{Integer Parameters}} \\
\hline
\( z_{\text{sh}} \) & [6, 9] \\
\( z_{\text{r2}} \) & [76, 85] \\
\( g_{\text{t\_h}}~[mm] \) & [3, 6] \\
\( p1_{\text{offset}} \) & [-1, 2] \\
\( p2_{\text{offset}} \) & [-1, 2] \\
\hline
\multicolumn{2}{|c|}{\textbf{Choice Parameters}} \\
\hline
material & PLA, ABS \\
lubricant & None, LI, Oil, PTFE \\
\hline
\multicolumn{2}{|c|}{\textbf{Float Parameters}} \\
\hline
\( x_{\text{s}} \) & [-0.8, 0.8] \\
\( x_{\text{r2}} \) & [-0.25, 1.0] \\
\( Cl [mm] \) & [0, 3] \\
\hline
\end{tabular}
\end{minipage}
\hfill
\begin{minipage}[t]{0.48\columnwidth}
\centering
\caption{Valve Parameters}
\label{tb:silicone_parameters}
\renewcommand{\arraystretch}{1.1}
\begin{tabular}{|m{2.5cm}|m{3cm}|}
\hline
\textbf{Parameter} & \textbf{Range/Options} \\
\hline
\multicolumn{2}{|c|}{\textbf{Choice Parameters}} \\
\hline
material & DS10, DS20, E0030, E0050 \\
cut & triple, x, slit \\
\hline
\multicolumn{2}{|c|}{\textbf{Float Parameters}} \\
\hline
t~[mm] & [0.2, 1.5] \\
$h_{dome}$~[mm] & [1.0, 3.8] \\
$l_{cut}$~[mm] & [0.75, 2.0] \\
\hline
\end{tabular}
\end{minipage}
\end{table}

The 10 gear parameters and 5 valve parameters are described in Table \ref{tb:parameters} and \ref{tb:silicone_parameters} respectively. Note that `h' in $z_{sh}$ and $g_{t\_h}$ denotes a half-value, restricting the actual values (sun gear teeth and gear thickness) to even integers. Gear thickness is restricted this way to interface correctly with purchased rods and bearings, which are themselves discrete sizes.

PLA [Matter Hackers] and ABS [Overture] plastics both use 1.75~mm filament. Lithium grease (LI) [Lucas], mineral oil (oil) [Hampton Bay], and Polytetrafluoroethylene lubricant (PTFE) [3-IN-ONE] are spread evenly over the gears by manual spinning prior to testing. Dragon Skin (DS) and Ecoflex (E) silicone rubbers are both mixed in 1:1 ratios.

\section{Experimentation}

\subsection{Fabrication}

\begin{figure}[!ht]
    \centering
    \includegraphics[width=0.8\linewidth]{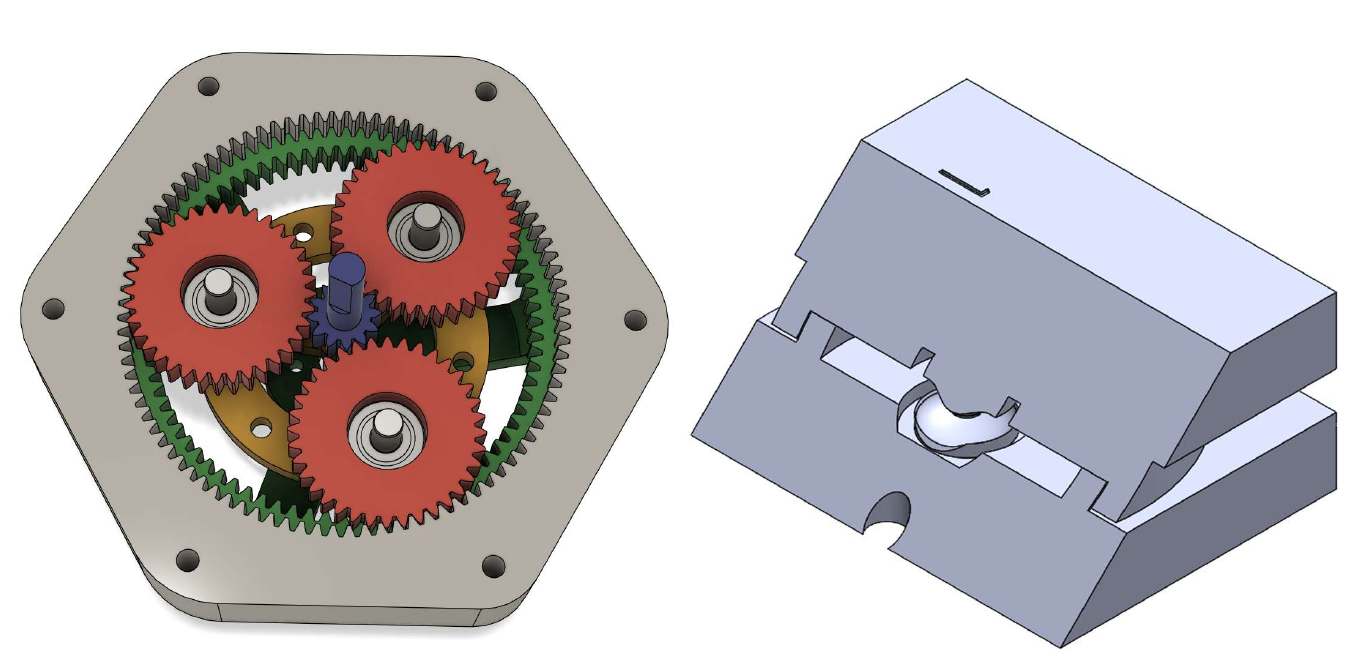}
    \caption{(left) CAD for 3D-printed Wolfrom gear. Parts include planet gears (red), sun gear (blue), stage 1 ring (grey), stage 2 ring (green), and two carriers (one shown, gold). (right) CAD for 3D-printed valve mold in `section view' to show negative space for domed valve and hole for injection molding.}
    \label{fig:Gear-CAD}
\end{figure}

\subsection*{Gears}

The 3D-printed (Bambu X1C) components of the Wolfrom gearbox include the planetary carrier, sun gear, two planet gears, and two ring gears (see Fig.~\ref{fig:Gear-CAD}). We print the full gear on a single build plate, streamlining the workflow. Because the stage 1 ring gear was reused across multiple design iterations, only the unique components needed to be reprinted, which significantly reduced print time and material consumption. We sliced all parts using a consistent 15\% infill with the Gyroid pattern to maintain mechanical consistency. 

Material selection, ABS [Overture] or PLA [Matter Hackers], was defined in the design parameters for each iteration. 
 After printing, we post-process parts to remove supports and clean contact surfaces. 
We press-fit steel dowel pins [McMaster] through the ball bearings [Yamaso], which we then press-fit into the bores of the planet gears. We mount these gear-bearing-pin assemblies onto the stage 2 planetary carrier and mesh them with the rest of the gearbox. 
We fasten the complete gearbox onto the 3D-printed fixture, which secures it coaxially with the Dynamixel MX-28 motor.

\subsection*{Valves}


The fabrication of each valve directly follows \cite{laake2022} and involves three main steps: mold creation (FormLabs3 - Tough1500), silicone injection-molding, and laser cutting. Molds are designed using `equation editor' in Solidworks, parameterized  for thicknesses and dome height. The valve molds consist of one male and one female part, as dictated by the dome protrusion and the dome cavity, respectively. 

One of four materials is used to create the valves - Dragon Skin 10, Dragon Skin 20, Ecoflex 00-30, and Ecoflex 00-50. We use equal parts A and B by weight. We add a thinning agent (NOVOCS) to all Dragon Skin in order to make the mixture less viscous and allow degassing. Before and after casting, the solutions are placed in a vacuum chamber to degas. The molds are injected with a 10~mL syringe. Before injecting, the valve molds are placed in a FormCure for 60 minutes at 70° C, washed with soap and water, and dusted with baby powder to ensure that the molds are dry and the valves cure properly. Valves are set to cure for minimum of four hours. 

The completed valve is placed in a vessel [PLA] similar to the injection mold. The vessel exposes the dome of the valve through a circular cutout ($radius=3~mm$). The valve is then laser-cut [Universal Laser Systems]. The power and speed of the cut are determined by parameters of the valve: the thickness of the valve, the material, and the height of the dome. Each power and speed combination is intended to fully cleave the material while minimizing kerf width. 

\subsection{Experimentation}

\begin{figure}[!ht]
    \centering
    \includegraphics[width=0.6\linewidth]{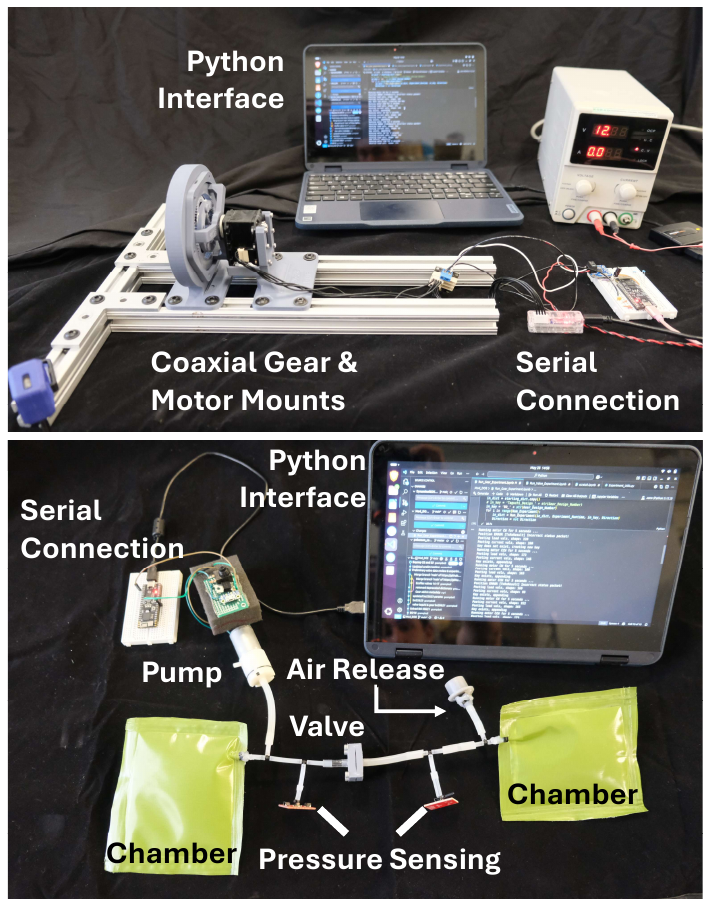}
    \caption{Experimental test setup for (top) gear spin and (bottom) valve pressurization.}
    \label{fig:testing}
\end{figure}
\subsection*{Gears}
We mount the gears to a fixed gear mount and coaxially to a DC motor (Dynamixel MX-28). 12~V DC is applied through a current sensor (INA260), monitored via a microcontroller interface (ESP32) and reported serially to the Python interface. The Python interface communicates with the motor through serial communication (U2D2 connector) to drive the motor and record torque load. The motor functions in `wheel' mode, with a goal speed of 11~RPM for five seconds. Six trials are performed for each gear, three with clockwise rotation and three with counterclockwise rotation.

For each gear, two types of data are measured during testing: torque output as ``DXT Torque \%'' over time (ms), and current draw (mA), measured via the INA260 power sensor. These values are captured during six trials—three clockwise and three counterclockwise—under identical operating conditions. The torque readings represent a percentage of the maximum rated torque measurable by the built-in sensor of the Dynamixel MX-28. For performance evaluation, the highest torque value from the best five of the six trials is selected and used to compute the reward score. 

\subsection*{Valves}

Valves are mounted in a 3D-printed holder chamber, which is positioned between two air pressure sensors (MPRLS0025), each connected to an ESP32 microcontroller to transmit pressure data. Air pressure is supplied by a  \hbox{4.5 V} DC air pump (ZR370-02PM) on the entry side of the valve. TPU-Coated nylon pressure chambers provide pneumatic capacitance on either side of the valve and compressed open-cell foam (McMaster) acts as a pneumatic resistor on the exit side of the valve. We activate the pump for 50 seconds and record pressure at entry to and exit from the valve.
For each valve, we monitor and store pressure data on each side of the valve. When pressure increases on the output side of the valve, the corresponding pressure on the inlet side is noted as $\Delta p_{crack}$. The average difference between pressures on the inlet and outlet side for all data-points after $\Delta p_{crack}$ is taken as the steady-state pressure differential $\Delta p_{ss}$. These values are verified manually for each valve by looking at the time-dependent pressure data. If a valve does not crack during experimentation, the $\Delta p_{crack}$ and $\Delta p_{ss}$ are both recorded as 100~kPa. $\Delta p$ values are scaled between 0 and 1 by dividing by 100~kPa before they are input into the learning client.

\subsection{Active Learning}

To carry out active learning (AL) \cite{ren2021}, we use the Ax platform with modeling from BoTorch. We initialize active learning with a set of n=10 and n=5 tests chosen randomly from the Taguchi trials for gears and valves respectively. This is meant to proxy for a space-filling method without requiring additional testing beyond what we had already performed. The number of initialization trials is chosen as equal to the number of design parameters in each case.

\subsection*{Gears}
For the single task, single objective gear optimization, we specify a `AdditiveMapSaasSingleTaskGP' model, which requires minimal hyperparameter tuning \cite{eriksson2021}. We specify a `warp' input transform class, SingleTaskGP model class, and Matern-5/2 kernel. Mean-squared-error is chosen as the model evaluation criterion. The acquisition function is detailed as a `qLogNoisyExpectedImprovement'.

The active learning procedure is then carried out by selecting the gear $M\in\mathbb{R}^{10}$ which best balances exploration and exploitation, determined using the Expected Improvement (EI) acquisition function. We carry out active learning in the parallel setting, where at each iteration $q=2$ gears are obtained simultaneously. We continue this procedure iteratively, until enough gears are acquired to match the Taguchi trials.


We use a singular score function for the gear optimization that takes into account both the gear ratio (higher is better) and the torque required to backdrive (lower is better). The score to maximize is:

\begin{equation} \label{eq: gear_reward}
    \text{R}_y = \frac{\text{Gear Ratio}}{\text{max}_{\substack{M_i \ne \max_j M_j}}^n M_i}
\end{equation}

where $R_y$ is the reward for gear design $y$, $M_i = max( \tau (t))$ is the maximum motor torque over trial~$i$ as a percentage of the motor's stall torque, and $n$ is the total number of trials for $y$. The denominator is simply the maximum torque percentage, ignoring the worst trial. A higher reward is better.

\subsection*{Valves}
For the single task multi-objective valve optimization, we use Ax's default BoTorch model class 'SingleTaskGP' 
and default acquisition function `qLogNoisyExpectedHypervolumeImprovement'. 

The active learning procedure is then carried out by selecting the valve $M\in\mathbb{R}^{5}$ which best balances exploration and exploitation, determined using the acquisition function. We carry out active learning in the parallel setting, where at each iteration $q=3$ valves are obtained before fabrication and testing. We continue this procedure iteratively, until enough valves are acquired to match the Taguchi trials.

Errors to minimize each objective are designated as:

\begin{equation} \label{eq: valve_cost}
    \text{L}_{y} = \sqrt{(p_{\text{target}} - p_{\text{actual}})^2}
\end{equation}

Where $L_{y}$ is the loss for valve $y$ and $p$ is the pressure. The equation is the same for both crack pressure and steady state pressure difference, and a lower loss is better.

\section{Results}

\subsection*{Gears}
\begin{figure} [!htb]
    \centering
    \includegraphics[width=0.95\linewidth]{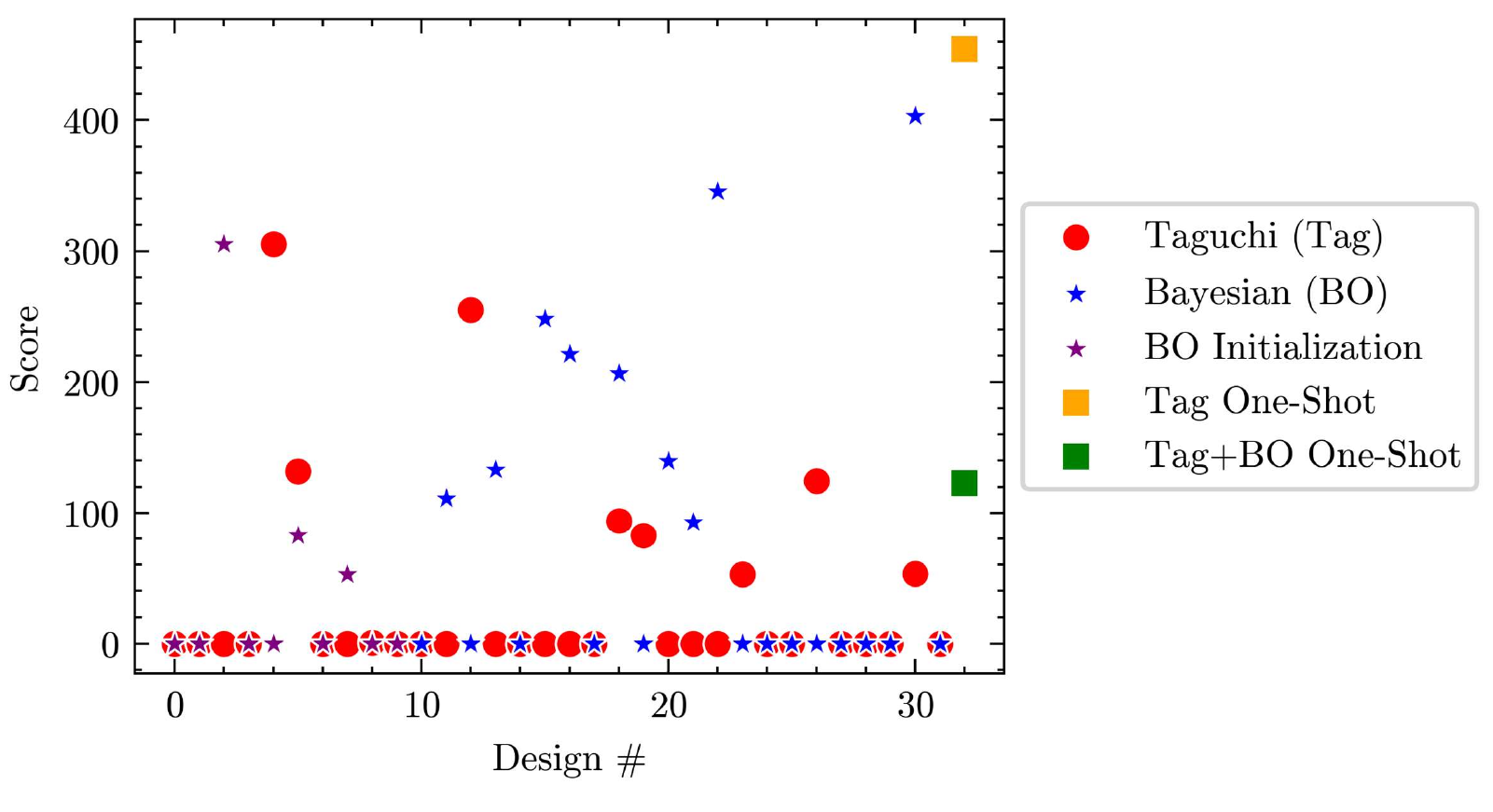}
    \caption{Reward function score (Eqn. \ref{eq: gear_reward}) for gear design iterations among the two design strategies.}
    \label{fig: Score_Graph}
\end{figure}

The Taguchi trials yielded a maximum score of 305 on the fifth design iteration. The seventh iteration of Bayesian Optimization (BO) outperformed the Taguchi methods with a score of 346. The best overall BO design was on the eleventh iteration, achieving a score of 403. A single `one-shot' iteration of BO performed with a model that trained on all Taguchi data outperformed all other methods with a score of 454. A similar `one-shot' iteration of BO with a model that trained on all Taguchi and BO trials did significantly worse with a score of 123.

The best scoring, `one-shot', gearbox uses a gear ratio of 63.6 and backdrives without ever needing more than 0.35~$Nm$ of torque. The next best, BO, gearbox uses a gear ratio of 56.9 and reaches a maximum backdrive torque of 0.35~$Nm$. The other BO gear of note uses a gear ratio of 107.7 and reaches a maximum backdrive torque of 0.78~$Nm$. The parameters for each of these three gears are as follows:


\begin{table}[ht!]
\centering
\vspace{-1em}
\begin{tabular}{ccccccc}
\toprule
Material & Lubricant & $z_{\text{sh}}$ & $z_{r2}$ & $x_s$ & $x_{r2}$ & Gear Ratio \\
\midrule
PLA & LI & 6 & 76 & -0.5872 & -0.0764 & 63.6 \\
PLA & None & 6 & 80 & -0.80 & 0.544 & 56.9 \\
PLA & LI & 6 & 76 & 0.7513 & 0.1686 & 107.7 \\
\bottomrule
\end{tabular}

\vspace{0.5em}

\begin{tabular}{ccccc}
\toprule
$Cl~[mm]$ & $g_{\text{thickness\_h}}~[mm]$ & $p1_{\text{offset}}$ & $p2_{\text{offset}}$ & Max Torque~[Nm] \\
\midrule
.033 & 6 & 2 & 2 & 0.35 \\
3 & 3 & 2 & -1 & 0.35 \\
2.71 & 3 & -1 & 2 & 0.78 \\

\bottomrule
\end{tabular}
\vspace{-1em}
\end{table}

\subsection*{Valves}

The Taguchi trials yielded a four-valve Pareto front with crack pressure errors as low as 0.4~kPa and steady state pressure errors as low as 0.5~kPa. Only one BO trial offered sufficient performance to enter this front.

\begin{figure} [!htb]
    \centering
    \includegraphics[width=\linewidth]{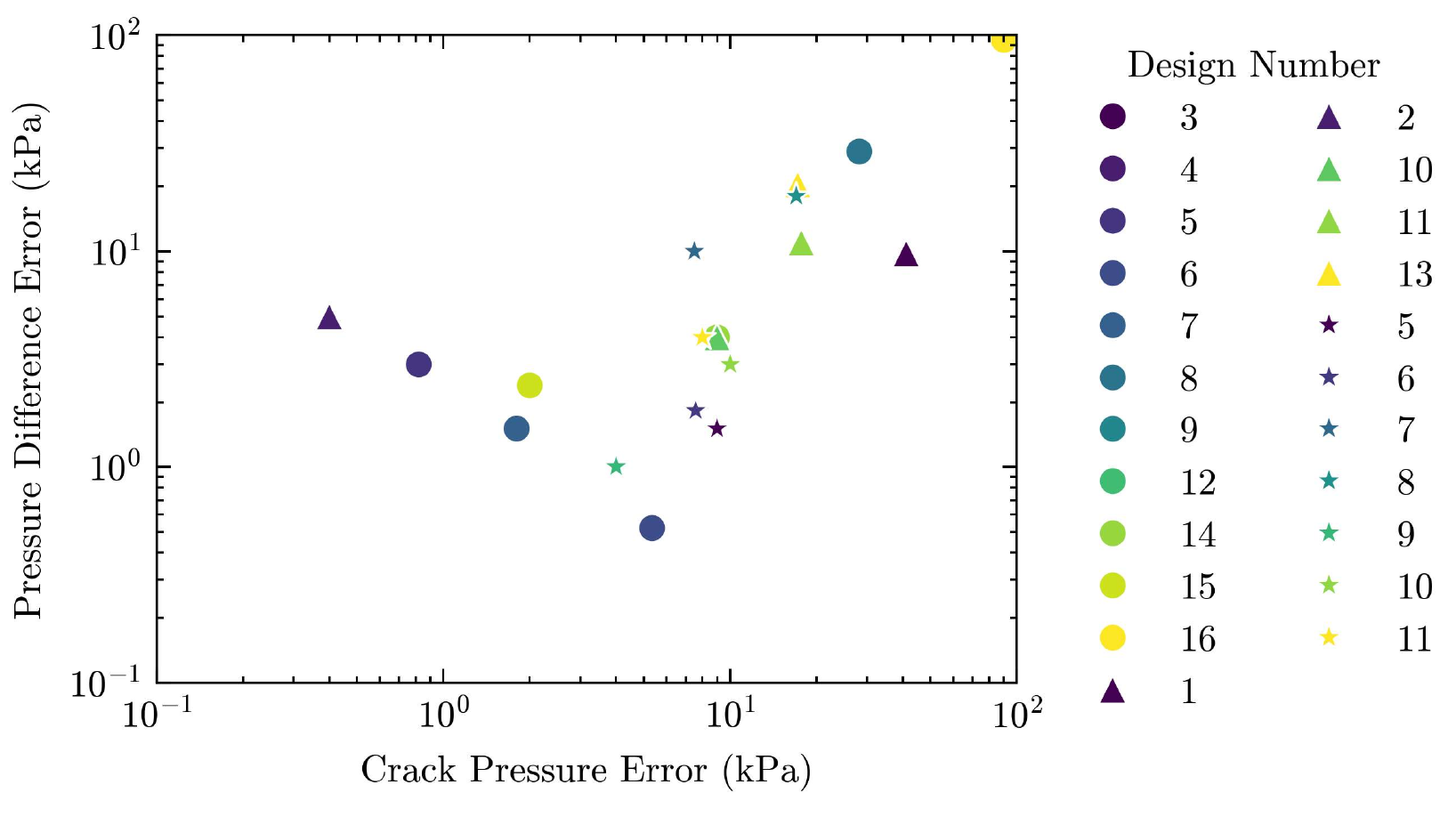}
    \caption{Valve pressure error values for Taguchi (circle and triangle) designs and Bayesian Optimization (triangle initialization, star generated) designs. Logarithmic scale accentuates the Pareto front.}
    \label{fig: Valve_Score_Graph}
\end{figure}

The parameters for each of the valves in the combined Pareto front are as follows: 

\begin{table}[ht!]
\centering
\vspace{-1em}
\begin{tabular}{cccccc}
\toprule
Design & Material & $t$ [mm] & $h_{dome}$ [mm] & $l_{cut}$ [mm] & cut \\
\midrule
2 & DS20  & 0.20 & 1.0 & 0.75 & X \\
5 & E0030 & 0.75 & 1.0 & 1.00 & triple \\
6 & E0030 & 0.20 & 2.2 & 1.50 & slit \\
7 & E0030 & 0.50 & 1.5 & 2.00 & X \\
BO 9 &	DS20 & 0.2& 1.16 &	0.75 &	X \\

\bottomrule
\end{tabular}

\vspace{0.5em}

\begin{tabular}{cccc}
\toprule
$\Delta p_{crack}$ [kPa] & $\Delta p_{ss}$ [kPa] & Crack Error [kPa] & SS Error [kPa] \\
\midrule
10.4 & 10.0 & 0.4 & 5.0 \\
9.2 & 8.0 & 0.8 & 3.0 \\
4.7 & 4.5 & 5.3 & 0.5 \\
8.2 & 6.5 & 1.8 & 1.5 \\
\bottomrule
\end{tabular}
\vspace{-1em}
\end{table}

\subsection{Modeling}

\begin{figure} [!htb]
    \centering
    \includegraphics[width=0.55\linewidth]{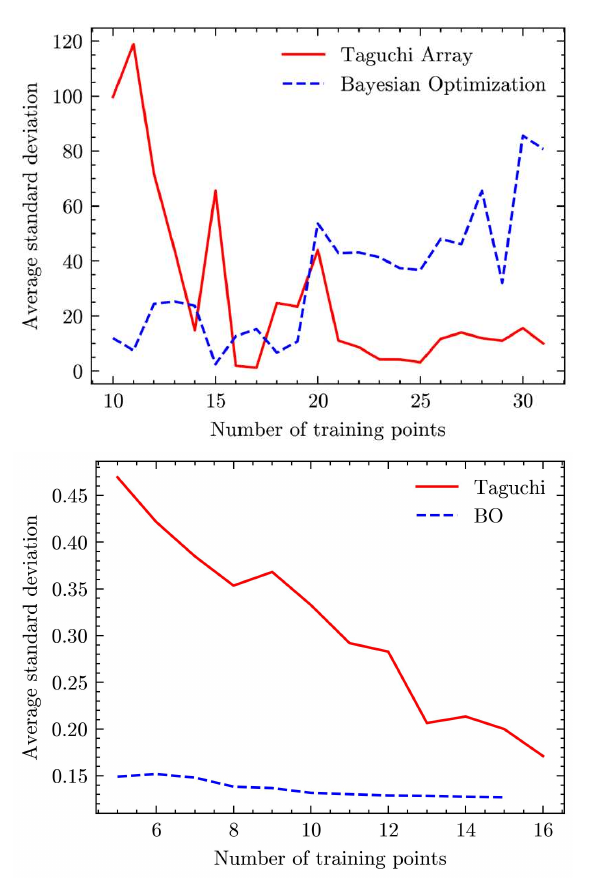}
    \caption{Model uncertainty, quantified by standard deviation, versus number of iterations included in the training set for a Gaussian Process Regression (GPR) model trained on BO (blue) or Taguchi (red) data. (top) Uncertainty for gear predictions. (bottom) Uncertainty for valve crack pressure predictions.}
    \label{fig: Uncertainty}
\end{figure}

The surrogate model uncertainty 
is determined sequentially with the addition of each trial for both Taguchi methods and BO methods beyond the 5 or 10 trial `initialization'. Standard deviation is averaged across 10\_000 sets of design parameters from a Latin hypercube \cite{helton2003} spanning the design space as detailed in Table~\ref{tb:parameters}. We used the trained model to predict the outcome score or crack pressure at each set of parameters, and the average standard deviation from these predictions is used as our measure of model uncertainty.


\subsection{Discussion}

\subsection*{Gears}
Looking at the uncertainty metrics in Fig. \ref{fig: Uncertainty}, we are likely functioning on too-few data points for successful training of the GPR. Our standard deviations are still on the rise for the BO trials. The Ax platform acknowledges that the model is not sufficiently trained upon request for best output with the warning: ``Metric [score] was unable to be reliably fit." The initially low standard deviations after the BO initialization (n=10) imply an overconfidence with insufficient training data. Standard deviations for the Taguchi set are finally starting to reduce near the completion of the Taguchi trials (n>25).  

Despite the high model uncertainty, 
the expected improvement (EI) acquisition function is still successful in outperforming the space-filling Taguchi method. The highest score comes from a mixture of the two methods, with the one-shot BO using a model trained exclusively on Taguchi data.

Looking across all trials, results qualitatively indicated that PLA with a lithium grease lubricant performed best overall, along with a mix of low $z_{r2}$ with high $z_s$. Relatively low clearance and high $p_{2~offset}$ also seemed to contribute to positive results. Quantitatively, Ax provides second-order parameter sensitivity analysis for each surrogate model. The metrics provided by the surrogate model trained on Taguchi data (the model that returned the highest performing design)  predict that $z_{r2}$ is the best predictor of score, followed closely by  material. These predictions are substantially different from those of the model trained during BO trials, which predicts a linear combination of $z_s$ and $p_{1~offset}$ as the best predictor of score, followed by $p_{1~offset}$.

\subsection*{Valves}
Looking at the uncertainty metrics in Fig. \ref{fig: Uncertainty}, the Taguchi trials provide excellent model training data for reducing uncertainty. Again, the initially low standard deviations after the BO initialization (n=5) imply an overconfidence, which remains throughout all executed tests. 

In this case study, the model overconfidence and limited initialization data lead to over-investigation of a local minimum, with the model suggesting designs similar to Taguchi design 2: using DS20 material and `X' cut geometry. This is likely because these were the materials used in the only member of the Taguchi Pareto front included in initialization. 

Looking across all trials, results qualitatively indicate that the optimal valve will likely use either DS20 or E0030 material. The lowest combined error (by both average and RMSE) across both crack and steady state pressure came from Taguchi design 7, which used an `X' cut geometry and a mid-range thickness of 0.5~mm. The metrics provided by a surrogate model trained on all Taguchi data predict that valve thickness has the largest effect on both $\Delta p_{crack}$ and $\Delta p_{ss}$. 

\subsection{Conclusions \& Future Work}

In this work we use active learning for design optimization across two case studies: co-design of design and manufacturing parameters of 3D-printed  gearboxes for large gear ratios and backdrivability, and multi-objective optimization for pressure response of a silicone check-valve. We show that the use of Bayesian Optimization improves score functions for desired outputs relative to a Taguchi baseline approach in only the single objective case, and evaluate metrics of surrogate model 
uncertainty as it is trained on the two test sets. Despite noting improvements in one study, we predict further model training is required to reach optimal results in both cases.

Based on these two case studies, it seems that BO is not yet in the `plug-and-play' regime to match that of Taguchi arrays. Based off design scores alone, Taguchi data seems more valuable than data points chosen by Bayesian Optimization (Expected Improvement (EI)) for model training. The relative performance of the gear one-shot BO following the completion of the gear Taguchi trials implies that sufficiently spanning training data is valuable for improving overall BO performance. This is also seen in the standard deviation metric for both cases, where training on Taguchi data allows for a more reliable decrease in model standard deviation. The valve results further emphasize this issue, as the minimal initialization set led to overconfidence of the model and an over-emphasis on exploitation over exploration. Further work is needed to explore the correct point at which to transition from a model-improving acquisition function (i.e., minimizing total uncertainty) to greedier optimization function (i.e., EI). 

The outputs of our machine learning approach to mechanical design provide helpful sensitivity analysis for understanding important aspects of the model. However, the volatility of this analysis (i.e., as we alter training data across trials) indicates that we may still be far from removing epistemic uncertainty from our models. 

Further work will be required to investigate the efficacy of alternate surrogate and acquisition functions in this type of sequential experimentation. Due to the nature of our physical design experiments, it is expensive to run sequential testing for multiple model strategies. Therefore, open access to the data from individual case studies such as this one will be instrumental in enabling a review across many design problems.
 
We were not successful in using BO to identify precise valve parameters that match the desired pressure differentials for our use case, and struggled to match the performance achieved by Taguchi trials. It is possible that the increased complexity of the soft pneumatic system response to changes in parameters was beyond the scope of the surrogate model being used. Future work may need to investigate alternative model options for this use case.

We were successful in identifying multiple options for gears with sufficiently high gear ratio and low back-drive torque. While a number of parameters seem to be optimized, perfecting these results and scaling the gears for individual mechanical systems is left to future works. This can also include expanding the parameter ranges for future BO, which would no longer need to match the baseline ranges.

\section*{Data Availability}

The majority of the work above was coded in Python, including in Jupyter Notebooks. This code, and a .pkl file containing a dictionary of all the test data used herein, will be made available on Github: \url{https://github.com/gmcampbell/Mod_DOE}.


\chapter{SUPPLEMENTARY INFORMATION}

In this section I try to include some of the practical skills, tricks, or lessons, that have helped me throughout my PhD. The target audience for this work are individuals who are relatively new to these topics. I hope you find something that is useful in your own work.
\section{Lab tools of a soft robotic experimentalist}

\subsection{Rapid prototyping of a pressure chamber for membrane testing}

\begin{figure}[!ht]
      \centering
    \includegraphics[width=0.9\linewidth]{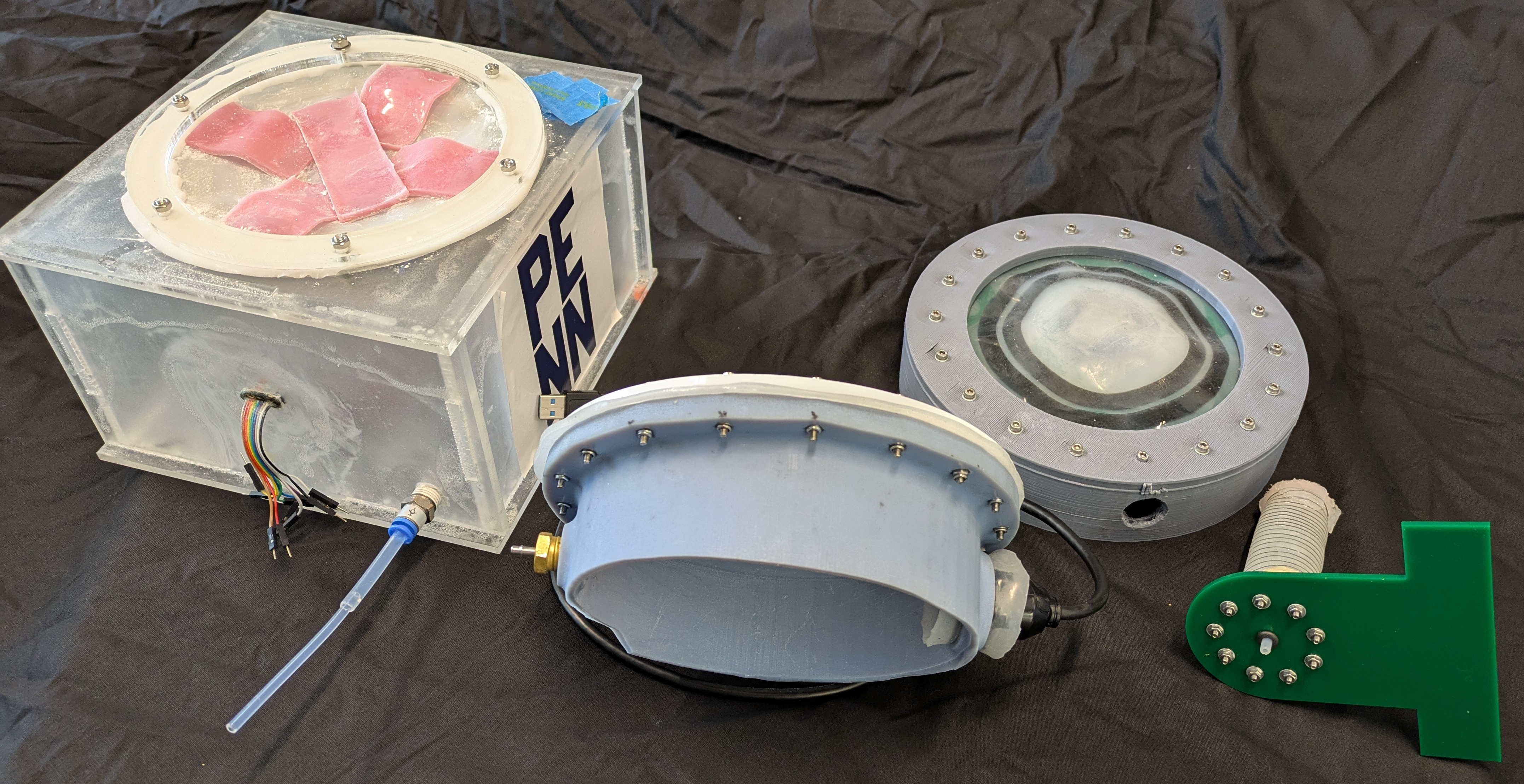}
      \caption{Various pressure-vessel designs. From left to right: laser-cut rectangular chamber (arcylic), burst 3D-printed cylindrical chamber (VeroBlue), 3D-printed and laser-cut mount and port (PLA and acrylic), laser-cut mount and port (acrylic).}
      \label{P_Chambers}
\end{figure}

Air is cheap and abundant and it doesn't make a mess. However, due to the second law of thermodynamics, trapping air is a fundamental problem for soft pneumatic actuators. Therefore a pressure chamber to effectively trap air becomes a fundamental resource. Leaks at transitions from soft structures to hard structures and ports for embedded electronics are particularly difficult to resolve in these pressure chambers. As Jessica Yin, PhD, and I developed standard soft robotics resources within our lab space, we spent significant time and effort overcoming this issue. You can see a number of the designs in Fig. \ref{P_Chambers}, with the oldest on the left and newest on the right. I include some details and references regarding pressure chambers below for posterity's sake. \footnote{Note that much of my research is done at gauge pressures below 20~kPa - I assume these problems only become worse for higher internal pressures.}

I have employed three main strategies for rapid prototyping of chambers that interface between soft pneumatic actuators (specifically thin-film or thin-walled elastomers) and rigid pressure chambers. Across designs, I use a top-ring that clamps down to the chamber (or bottom-plate). To create compression against the actuator with this top-ring, I use simple nuts and bolts evenly spaced around the ring. I greatly increased the number of fasteners across iterations, though the number required can be estimated with equation 8-34 from \citep{shigley_mechanical_2015}. These fasteners penetrate the thin-film at an inextensible location (where the elastomer has been reinforced with a fabric) to prevent ripping.

\subsubsection{Rectangular chambers}
A rectangular chamber is incredibly intuitive and can be created quickly with from laser-cut acrylic and acrylic cement. These chambers are used effectively in a number of relevant works, including \citep{forte2022, stanley2015}. Our main difficulty with these chambers were leaks at seams between laser-cut pieces. We reinforced these seams and any ports with extensive liquid silicone (Rust-Oleum Leak Seal). I would recommend this type of chamber only if you both do not have access to a 3D-printer and need to include pressure sensing within your pressure chamber.

\subsubsection{Cylindrical chambers}
Cylindrical chambers are notable in that they have significantly less seems than their rectangular counterparts. We were inspired towards this footprint by \citep{alspach2019}, and used an iteration in both \citep{campbell2022, campbell2025}. We printed in high resolution in VeroBlue using an Objet 3D-printer. This was particularly nice because we could include a cutout for the nuts and an O-ring channel in the print, which both made attaching the top-ring easier. Similarly, ports could be sized precisely and were leak-proof after being secured with a gasket (no need for liquid silicone). Depending on the wall-thickness you use, these can be prone to cracking. Ours cracked at about 50~kPa.

The down-side to the 3D-printed method is the time and expense. I consider this worthwhile only if you need to mount sensing hardware beneath the soft membrane (i.e., cameras or pressure sensors). If this isn't required, consider non-chambers below.

\subsubsection{Non-chambers (ports)}
Ports are generally the most straight-forward means of getting air behind a thin-walled device. If you can design the device such that it wraps around the port (for instance, a classic latex balloon), then all you'll need is a hose-clamp to connect the two. I have found this particularly effective for fabric designs (i.e., workflows such as \citep{shveda2022} - see the supplement). If you instead want a specific boundary for the soft vessel, this can be enforced with a rigid laser-cut or 3D-printed piece that includes a hole for the port and through-holes for fasteners.

I have used these for both large (150~mm diameter) membranes and small (20~mm diameter) iPAMs \citep{hawkes2016}, as seen in Fig. \ref{P_Chambers}. I found the simplest and most reliable port to be an off-the-shelf grommet press-fit into the lasercut piece with a rigid tube then press-fit into the ID of the grommet. While the designs shown here both feature laser-cut acrylic, I have also made a similar valve-clamp from only 3D-printed PLA (Bambu XC1 printer). This was convenience for implementing an O-ring channel and the port directly (no grommet).

\subsection{Silicone curing for extensible actuators}
Most of the actuators I have fabricated throughout the work in this dissertation \citep{campbell2022, campbell2025} has been gravity-molded in molds made from two pieces of laser-cut acrylic bonded together with acrylic cement. In particular, I have added strain limiters in the same way as \citep{pikul2017}. This process is described in detail in Appendix \ref{Mem_Procedure}. Smooth-On silicones are quite straight-forward to mix, though degassing chambers can be a moderate expense in bringing this process to your lab. I have rarely used mold release for gravity molding, and I do not think it necessary (for Dragonskin or Ecoflex) unless you are dealing with detailed parts.

I have also done significant injection molding, specifically in recreating the work of \citep{hawkes2016} and \citep{laake2022}. I have found success with 3D-printed PLA molds for larger (order of 100~mm) pieces, but use FormLabs printed molds for smaller (order of 10~mm) parts. 

\subsection{Heat-sealing and low-strain materials}
I have not delved too deeply into low-strain materials, though they clearly have potential in soft robotics \citep{hawkes2017, shveda2022, choi2020}. Choosing the correct material was very important. In particular, TPU-coated nylon ripstop (Seattle Fabrics) was excellent for my purposes. I found impulse heat-sealing to lead to stronger seals, but the convenience of patterned seals with a heat-press was great for larger or layered actuators. Finally, double hose-clamps (zip-ties) for soft-hard interfaces and hot-glue around seems made for simple porting. I point you to the supplement of \citep{shveda2022} for more helpful details.

\section{Electroadhesive Theory}\label{EA_Theory}

Many works that model electroadhesive (EA) frictional holding force \citep{Chen2017,imamura2017,diller2018,Germann2012,Hinchet2019} use the following equation to quantify said force. 

\begin{equation} \label{Parallel_Plate_eqn}
f_N=\frac{\epsilon _0\cdot \epsilon _r\cdot A}{2}\cdot \left(\frac{V}{d}\right)^2
\end{equation}

where V is the potential difference between electrodes, $d$ is the distance between the pad and the substrate, $\epsilon_0$ is the permittivity of free space (8.854 * $10^{-12}$ F/m), $\epsilon_r$ is the relative permittivity of the EA pad dielectric (also written as dielectric constant, $k$), and $A$ is the surface area of the electrode. 

In this section, I include a first-principles derivation of this popular friction holding force as well as an alternative derivation that I believe to better reflect the plates I have used throughout my work, taking into account an air gap between plates. The derivation of this equation was a portion of my qualifying assessment, and has bolstered my understanding of the underlying physics behind EA clutches. Though many of the citations in this section are from many years ago even as I write this, I saw Eqn. \ref{Parallel_Plate_eqn} appear as recently as a 2025 conference paper review, so I do still think this is a relevant distinction.

\subsection{References to EA clutch friction}

Eqn. \ref{Parallel_Plate_eqn} relies on an \textit{ideal capacitor model} when calculating the normal force between EA clutch pads. By this, I mean that there is perfect contact at the interfaces between electrodes and dielectrics and, far more presumptively, that perfect contact also exists between dielectric plates.

Chen et al. \citep{Chen2017} discuss the importance of air gaps, noting that \enquote{air gaps exist because of surface roughness and stiffness in the plates that prevent contact.} They continue that “in a limiting case, it can be assumed that the electroadhesive is compliant and comes in close contact with the surface without any air gaps..." Chen et al. assume this limiting, compliant case, and reduce their \enquote{apparent contact area,} the overlapping area between plates, to a \enquote{true contact area,} which they determine based on a visual analysis of the EA pad in contact with glass. Therefore, the assumption seems to become that throughout this true contact area, the limiting case applies and there are no air gaps between the EA and the substrate. It is worth noting that Chen et al.'s work compares experimental results to this area-adjusted equation with great success, but also that the dielectric constant of Perylene (their chosen dielectric) is 2.17-3.15 \citep{SCSParylene}, which is low compared to that of dielectrics used in other works.

Both Diller et al. \citep{diller2018} and Hinchet et al. \citep{Hinchet2019} cite Chen et al.'s \citep{Chen2017} work when discussing their equations for EA force. Notably, neither of them take into account the ratio of true versus apparent contact area that was discussed by Chen et al. Hinchet et al. also note that Eqn. \ref{Parallel_Plate_eqn} is for ideal parallel plate capacitances without ever discussing why their clutch fits (or doesn't fit) that description \citep{Hinchet2019}. Hinchet et al. cite Karagozler et al. \citep{karagozler2007} and Persson \citep{persson2018} as well, without noting that both works in fact support the series-capacitor model and not the parallel plate model, displaying equations that differ from Eqn. \ref{Parallel_Plate_eqn} by a factor of $\epsilon _r$ (see Eqn. \ref{2Di_Electric_Field_Eqn}).

There are other works that also use this imperfect capacitor form of the friction equation, and even more complex models (including fracture). For those interested, I would recommend continuing with  work from the lab of Kevin Turner, PhD, starting with Levine et al.'s work from 2022 \citep{levine2022}.



\subsection{Fundamental Formulas and Difference Between Models}

The physics of the EA clutch is summarized in an electrostatics problem. The charge that builds up on each electrode creates an electric field, which applies an associated stress attracting the opposite charge created on the opposing plate. If we assume that the plates are large compared to the distance between them and ignore edge conditions, the electric field can be solved using Gauss's Law and electrostatic boundary conditions, and the associated stress can be solved using the Maxwell Stress Tensor. We further assume that the electric field is unidirectional and that no magnetic fields exist, which allows for the reduction of the stress tensor to the first half of the Lorentz force law.

As a first step, the strength of the electric displacement field can be calculated using a version of Gauss's law:


\begin{equation} \label{Gauss_Law}
\oint_S {D_n \cdot dA = Q_{inside}}
\end{equation}
where the left-hand side is the integral of $D_n$, the electric displacement field ($D_n = \epsilon _r \cdot \epsilon _0 \cdot E_n$, the electric field strength ($E_n$) multiplied by the total permittivity of the interior material), dot-product with $dA$, a small piece of area, over the surface area of a closed surface. $Q_{inside}$ is the net charge contained  within the closed surface.

Thinking of the clutch as a capacitor allows for the relation of potential difference (V) to charge (Q) by means of capacitance (C): $Q = C \cdot V$.

The more common form of Gauss's Law ($\oint_S {E_n \cdot dA = \frac{Q_{inside}}{\epsilon _0}}$) assumes that there are no dielectric materials other than air surrounding the charges, and therefore that the electric field ($E_n$) can be calculated directly. This can be easily altered if there is a single dielectric present, but is less helpful in the case of multiple dielectrics (in this case, Luxprint and air).

Once the electric field is solved from the electric displacement field ($D = \epsilon \cdot E$), it can be plugged into the Maxwell Stress Tensor to solve for the associated stress. Because we assume both that the electric field is only in one direction, which we will call z, and that there are no magnetic fields, there is only one non-zero term in the Maxwell Stress Tensor, which is reduced as follows:

\begin{equation} \label{Lorentz_Force_Law}
\begin{aligned}
\sigma _{ij} = \epsilon (E_iE_j-\frac{1}{2} \delta _{ij} E^2) + \frac{1}{\mu _0} (B_i B_j - \frac{1}{2} \delta _{ij} B^2)\\
\sigma _{zz} = \frac{\epsilon}{2} \cdot E_z^2
\end{aligned}
\end{equation}
where $\sigma _{ij}$ is the term in the i$^{th}$ row and j$^{th}$ column of the Maxwell Stress Tensor, $\epsilon$ is the permittivity of the material ($\epsilon = \epsilon _0 \cdot \epsilon _r$), $E$ is the electric field strength (subscripts denoting portion of field in a specific direction) and $B$ is the magnetic field (assumed to be zero). $\sigma _{zz}$ is the stress in the direction normal to the plates, and $E_z$ is the electric field in the same direction.

The product of this normal stress can be taken with the area of a clutch plate to solve for the normal force between the plates. Finally, if shear force is required (like in my own clutch work), the product of the normal force and the coefficient of friction at the sliding interface must be taken.

When both air and a dielectric are included instead of only one of the two, the electric displacement field does not change, seeing as it relies only on the charge enclosed and the surface drawn. The electric field, on the other hand, changes by the relative permittivity ($\epsilon _r$) at each dielectric boundary, increasing as it moves from Luxprint to air. The permittivity within the material will also change, by the same factor, decreasing as it moves from Luxprint to air. By Eqn.~\ref{Lorentz_Force_Law}, the stress and associated force will change by $\frac{\epsilon _r^2}{\epsilon _r}$ : the relative permittivity $\epsilon _r$.

\subsection{Electrostatic Derivation of Parallel-Plate Equation}

\begin{figure}[!ht]
    \centering
    \includegraphics[width=0.25\textwidth]{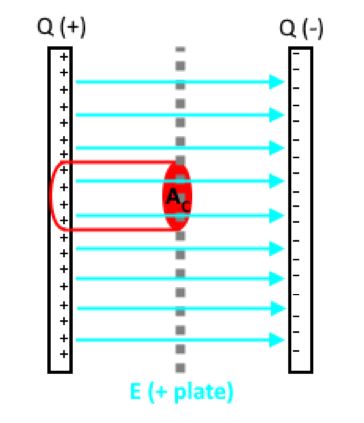}
    \caption{Diagram of surface chosen for use in Gauss's Law calculation of electric field between electrodes with charge Q}
    \label{fig:Gauss_Law}
\end{figure}

Solving Eqn. \ref{Gauss_Law} for the field created by the positive plate over a cylindrical surface, as seen in Fig. \ref{fig:Gauss_Law}, reduces as follows:

\begin{equation} \label{Reduced_Gauss_Law}
\begin{aligned}
    D_n^+ \cdot 2 \cdot A_C = Q_{+} \cdot \frac{A_C}{A_{plate}}\\
    D_n^+ = \frac{Q_{+}}{2 \cdot A_{plate}}
\end{aligned}
\end{equation}
where $A_C$ is the area of the circle at the end of the cylindrical surface (as marked in Fig. \ref{fig:Gauss_Law}), $A_{plate}$ is the area of an electrode, and the + super or subscripts represent that the charge and displacement field are associated only with the positive plate.

Therefore, the electric displacement field due to the positive plate is constant throughout the capacitor, which also means that the electric field is constant within the capacitor, and is defined by the following:

\begin{equation} \label{Electric_Field_Eqn}
E_n^+ = \frac{D_n^+}{\epsilon} = \frac{Q_{+}}{2 \cdot A_{plate} \cdot \epsilon _r \cdot \epsilon _0}
\end{equation}

Charge (Q) is then calculated based on the potential difference (V) and the capacitance (C), where capacitance is solved from clutch parameters ($d$ being the distance between plates):
 \begin{equation} \label{C_eqn}
 \begin{aligned}
    Q = C \cdot V = \frac{\epsilon_r\cdot\epsilon_0\cdot A_{plate}}{d} \cdot V
 \end{aligned}
 \end{equation}
 
Combining Eqn. \ref{Electric_Field_Eqn} and Eqn. \ref{C_eqn}, we get the reduced equation for electric field ($E_n$) based on potential difference (V):

\begin{equation} \label{Updated_Electric_Field_Eqn}
\begin{aligned}
E_n^+ = \frac{\frac{\epsilon_r\cdot\epsilon_0\cdot A_{plate}}{d} \cdot V}{2 \cdot A_{plate} \cdot \epsilon _r \cdot \epsilon _0}  = \frac{V}{2 \cdot d}\\
\end{aligned}
\end{equation}

By symmetry, the negative plate will create the same electric flux and corresponding electric field within this clutch. Therefore, the total electric field ($E_n$) at any point between the plates is equal to $\frac{V}{d}$. Plugging this total electric field into the Maxwell Stress Tensor, we can solve for the stress and ultimately the normal force:

\begin{equation} \label{Electric_Field_Force}
\begin{aligned}
\sigma _{zz} = \frac{\epsilon}{2} \cdot E_z^2 = \frac{\epsilon _0 \cdot \epsilon _r}{2} \cdot \left(\frac{V}{d}\right)^2 \\
F_N= \sigma _{zz} \cdot A_{plate} = \frac{\epsilon _0\cdot \epsilon _r\cdot A_{plate}}{2}\cdot \left(\frac{V}{d}\right)^2
\end{aligned}
\end{equation}

Eqn. \ref{Electric_Field_Force} exactly matches Eqn. \ref{Parallel_Plate_eqn}, informing my claim that this equation relies on the ideal capacitor model. 

\subsection{Electrostatic Derivation Series-Capacitor Equation}

When there is more than one dielectric between the parallel plates, the parallel-plate derivation changes only slightly so long as the air gap is assumed to be small. Because the relative permittivity ($\epsilon _r$) varies between the plates, specifically at the air-dielectric interfaces, Eqn. \ref{Electric_Field_Eqn} holds, but is no longer a constant in the direction normal to the plates. At the interface between the dielectric and the air, the electric displacement field ($D_N$) is constant, but there is an immediate drop in the electric field ($E_N$) by a factor of $\epsilon _r$ because permittivity of the material at the air gap is approximately $\epsilon _0$, instead of the $\epsilon _0 \cdot \epsilon _r$ that it is within the dielectric.

\begin{equation} \label{2Di_Electric_Field_Eqn}
\begin{aligned}
D_2^+ = D_1^+\\
\epsilon _0 \cdot E_2 = \epsilon _r \cdot \epsilon _0 \cdot E_1\\
E_2^+ = \epsilon _r \cdot E_1^+\\
E_2^+ = \frac{\epsilon _r \cdot V}{2 \cdot d}
\end{aligned}
\end{equation}
where $E_2$ is the electric field within the air gap, and $E_1$ is the electric field within the dielectric (which is equivalent to that solved in Eqn. \ref{Updated_Electric_Field_Eqn} so long as the air gap is small). 

Again, the negatively charged plate will contribute an equal electric field, making the total field at any point within the air gap $E_z = \frac{\epsilon _r \cdot V}{d}$. We can move forward as was done in Eqn. \ref{Electric_Field_Force} to solve for the updated resultant force:
\begin{equation} \label{Correct_EA_Clutch}
\begin{aligned}
\sigma _{zz} = \frac{\epsilon}{2} \cdot E_z^2  = \frac{\epsilon _0}{2} \cdot \left(\frac{\epsilon _r \cdot V}{d}\right)^2 \\
F_N= \sigma _{zz} \cdot A_{plate} = \frac{\epsilon _0\cdot A_{plate}}{2}\cdot \left(\frac{\epsilon _r\cdot V}{d}\right)^2
\end{aligned}
\end{equation}


The importance of taking into consideration the air gap between materials is especially relevant at relatively low voltages or with stiffer electrodes. I found Eqn.~\ref{Correct_EA_Clutch} to be substantially more accurate in my experimentation. 

\subsection{Where Assumptions Break Down}

This model has limitations, especially near the edges of the clutch. Because of the infinite-plate assumption, we could safely assume that the electric displacement field and electric field were only in the direction normal to the plates. However, near the edges of the plates, this assumption fails, and the direction of the electric displacement field changes. This would lead to a smaller associated normal force but would also contribute to small forces in tangential directions at these edges. 

This model also assumes that the air gap is small and of a consistent width. For very rough surfaces, there would be an uneven distribution of electric field and a varying distance between clutch plates, leading to inconsistent levels of adhesion. The presence of deeper asperities could also theoretically lead to interlocking and associated shear forces that do not directly relate to calculated normal force and measured coefficient of friction. These would need to be considered separately, and again I point the reader to \citep{levine2022} for fracture considerations.

\subsection{EA Fabrication} Note I only go into the theory of EA, as opposed to the type fabrication details I include for pressure chambers. Stuart Diller, PhD, made a great video about clutch fabrication \citep{diller2018}. As I write this, it is available on the YouTube channel of Steve Collins, PhD, under the title 'Diller 2018 JIMSS Video Fabrication'.

\section{Automated Testing Rig}

Testing for Chapters \ref{ch: Membrane_Design_Section} and \ref{ch: BO_Taguchi} is performed on customized test systems. Details for one of the test systems in Chapter \ref{ch: Membrane_Design_Section} are as follows. 

The test system height is controlled in parallel to the internal pressure of the actuator. All components are controlled by a microcontroller (ESP32 Pico Kit), which receives serial instruction from a high-level controller running in Python while simultaneously reporting sensor states via the same serial channel. The relevant sensors are internal pressure (Qwiic MicroPressure Sensor; Sparkfun), volumetric flow (Renesas FS2012-1100-NG; Digikey), time of flight (VL53L0X; Adafruit), and load cell (\textit{unknown} - HX711 breakout; Sparkfun).



\begin{figure}[ht!]
      \centering
      \includegraphics[width=.8\linewidth]{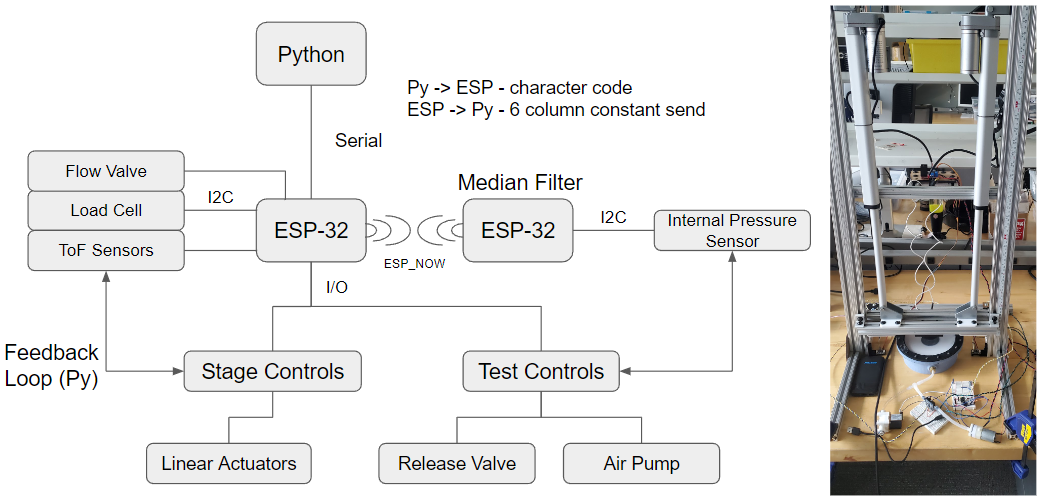}
      \caption{left: testing system electromechanical setup, right: image of testing system}
      \label{AE Tester}
\end{figure}

The communication and control for the system is described in Fig. \ref{AE Tester}. Power is supplied by a AC-DC transformer at 12~V with a 5~V buck converter for motors at the respective voltages. A separate 5~V battery powers the sensors in an effort to separate them from electrical noise.

\chapter{MEMBRANE FABRICATION AND AUTOMATED TESTING PROCEDURE} \label{Mem_Procedure}

The following is an abbreviated version of the instructions that I created for my undergraduate student worker, Jason Matthew, when he performed membrane fabrication and testing for work shared in Chapter \ref{ch: Membrane_Design_Section}. 
We hope these details are useful for someone trying to recreate our membranes or testing.

\section{Membrane Fabrication}

\subsection{Membrane Stabilizer}

Membrane stabilizer (\textit{Soft n’ Shear stabilizer}) is used to apply local strain limitation in a membrane. Stabilizer is shaped by cutting the stock fabric on a laser-cutter.

\textbf{Materials:} Soft n’ shear stabilizer, scissors, masking tape, baggies for cut stabilizer, backing sheet (usually ABS but acrylic/MDF is fine too), \texttt{.dwg} files, laser cutter

\textbf{Output:} Fabric stabilizer sized for individual membranes

\paragraph{Step 0: Setup}
\begin{itemize}
  \item Turn on laser cutter, fume fan, and computer.
  \item Open one of the membrane design \texttt{.dwg} files in DraftSight. Adjust any ring radii as necessary.
  \item Load the backing sheet into the laser cutter.
\end{itemize}

\paragraph{Step 1: Portion \& mount stabilizer}
\begin{itemize}
  \item Cut a length of $\sim$400mm of Soft n’ Shear and tape down the edges.
  \item Adjust tape to relieve wrinkles.
  \item Calibrate the z height using the calibration tool.
\end{itemize}

\paragraph{Step 2: Print, cut, and remove}
\begin{itemize}
  \item Print the \texttt{.dwg} file in DraftSight at 1:1 size ratio (mm)
  \item Use laser cutter control software to duplicate view ($x=2$).
  \item Position prints, check alignment with the stabilizer.
  \item Set laser (red): 25 - 100 - 500 (Power\%, Speed\%, PPI).
  \item Cut, monitoring fabric to make sure it isn't blown away. Pause if needed to remove detached pieces that block the laser.
  \item After cutting, bag rings appropriately.
\end{itemize}

Repeat steps 0-2 as necessary for the current batch of membranes. Note that any rings under ~20mm in width (outer radius less inner radius) need to be double-stacked so they do not snap. That means you’ll need to cut twice as many rings as you’ll make membranes.

\subsection{Ring Preparation} \label{Ring_Prep}

Thin rings of membrane stabilizer need to be double-stacked or they are prone to stress failure during inflation. We can prepare these stacks prior to membrane fabrication.

\textbf{Materials:} Cut stabilizer rings, membrane mold, acrylic ring (optional), silicone Parts A \& B, mixing cup, stir rod, gloves

\textbf{Output:} Doubled stabilizer rings

\paragraph{Step 0: Prepare}
\begin{itemize}
  \item Wear gloves, clean the station.
  \item Make sure that drying racks are within calibration (flat)
    \subitem Adjust leg height as necessary

\end{itemize}

\paragraph{Step 1: Mix \& degas Silicone}
\begin{enumerate}
  \item Shake Parts A and B.
  \item Weigh and mix equal masses.
  \item For Dragonskin: add 10\% NOVOCS, stir again.
  \item Degas for ~8 minutes.
\end{enumerate}

\paragraph{Step 2: Align the rings}
\begin{itemize}
  \item Stack rings in mold, align with acrylic sizing circle or etchings if available.
  \item Deposit $\sim$1g of silicone at 4 points using a stir rod.
  \item Allow to cure $\sim$1 hour.
\end{itemize}

\subsection{Membrane Fabrication}
Once you have sized the stabilizer for your membranes and created double-layer rings, you have the requisite materials to fabricate the membranes themselves. It is critical that these membranes be flat (equal thickness throughout) and that rings be concentric. Failure to meet either of these criteria will result in invalid test results.

\textbf{Materials:} Doubled rings, mold, silicone, gloves, stir tools

\textbf{Output:} Ready-to-test membranes

\paragraph{Steps 0 and 1:} Identical to Section \ref{Ring_Prep}

\paragraph{Step 2: Mold and stabilizer prep}
\begin{itemize}
  \item Clean mold, place a small circle of VHB tape in the center.
  \item Add contact stiffener (central piece) and rings concentrically.
\end{itemize}

\paragraph{Step 3: Silicone pouring \& degassing}

\begin{itemize}
  \item Tare mold on scale.
  \item Pour silicone (30g/mm thickness, slightly more for Dragonskin due to NOVOCS) into mold. Use stir rod to even out as necessary.
  \item Degas poured mold (2 minutes).
  \item Add edge piece of stabilizer (OD = mold OD) to set membrane radius and secure screws.
  \item Cure for 2 hours on level drying rack, set 24 hours before use.
\end{itemize}

\paragraph{Step 4: Mass \& labels}
\begin{itemize}
  \item After curing, label membranes with mass, date, material, rack, and ring dimensions.
\end{itemize}

\section{Membrane Testing}

Scripts: \url{https://github.com/gmcampbell/SPA_Design} (Data Acquisition)

Use python scripts to control testing and data collection with our custom testing rig. Note that Run\_Experiment.py contains sections for both experiment-specific and computer-specific inputs. You’ll need to update AT LEAST the filepath for saving data and likely the port numbers for the microcontroller and camera. If you haven’t used python extensively in the past, you may need to install packages with pip.

\subsection*{WARNING}
This testing system uses significant electrical power and equally powerful motors. It is an experimental system and does not have all the safety systems of a commercial system. Never put your hands beneath the moving parts of the system. Be aware of the red, emergency stop button prior to using the test system.

\subsection{Calibration} \label{Calibration}

\textbf{Materials:} Calibration device, test rig

\paragraph{Step 0: Setup} 
\begin{itemize}
  \item Power order: USB (microcontroller) $\rightarrow$ External (sensors) $\rightarrow$ Wall (actuators)
  \item Check COM ports and camera settings.
  \item Level the load-cell with external level (i.e., Phone app)
\end{itemize}

\paragraph{Step 1: Calibration Script}

This script is not included in the public git, but essentially just levels the load-cell and confirms time-of-flight sensors correctly measure contact height relative to the membrane.

\begin{itemize}
  \item Run \texttt{Test\_Rig\_Calibration.py}
  \item Measure manually using calibration tools.
  \item Paste output into \texttt{Serial\_Comms.py}'s affine transform.
\end{itemize}

\subsection{Mount Membrane}

\textbf{Materials:} Pressure chamber, O-ring, drill, membrane

\paragraph{Step 1: Remove Old Membrane}
\begin{itemize}
  \item Fit the M2.5 haxagonal bit into the drill and use it to remove the 18 pressure-chamber screws. After unfastening a screw, next remove it’s partner screw that is opposite across the ring.
  \item Remove the ring-membrane pair and place both 'upside down' (screws up) on the table. Pull the membrane away while leaving the ring on the table.
\end{itemize}

\paragraph{Step 2: Mount New Membrane}
\begin{itemize}
  \item With the top ring upside-down (screws facing up), poke the exposed screws through the holes in the ‘edge piece’ of stabilizer one at a time, securing the membrane concentric to the top ring.
  \item Add an Ecoflex gasket if membrane thickness is 1~mm or if using a stiffer (DragonSkin) membrane.
  \item Add O-ring to the pressure-chamber's embedded channel. Check internal electronic connections.
  \item Drill setting: forward, speed 1, torque 2.
  \item Align and secure membrane with all 18 screws, pinching top-ring to pressure chamber as you apply each screw. After each screw, next secure it’s partner screw that is opposite across the ring.
  \item Apply baby powder on testing surface to reduce friction.
\end{itemize}

\subsection{Perform Test}

\textbf{Materials:} Pressure chamber, testing rig

\paragraph{Step 0: Setup} See Section \ref{Calibration}

\paragraph{Step 1: Run \texttt{Run\_Experiment.py}}
\begin{itemize}
  \item Update \texttt{Sample\_String} with: material, thickness, radius, etc.
  \item Run the test, aligning contact patches during first inflation to ensure conformal contact.
  \item If a pop-event occurs, manually depress a limit switch (push the lever until it clicks). Allow one more test to begin and depress a limit switch again before canceling remaining tests by pressing ctrl+c in the terminal.
  \subitem Delete post-pop data.
  \item Mark date on membrane.
\end{itemize}

\subsection{Data Upload \& Active Learning}

\textbf{Code:} \url{https://github.com/gmcampbell/SPA_Design}

\textbf{Main Script:} \texttt{select\_next\_experiment.ipynb}

\textbf{Output:} Updated \texttt{.pkl} file and new design parameters.

This script is well documented (credit to Leonardo Ferreira Guilhoto) and it should be somewhat straightforward to upload a data file (.pkl) and receive future test parameters.
\end{append}

\begin{bibliof}
\bibliography{bibliography}
\end{bibliof}
\end{document}